\documentclass[Afour,sageapa,times]{sagej}

 \usepackage{moreverb}
 \usepackage{url}
 \usepackage[utf8]{inputenc}
 \usepackage{dirtytalk}
 \usepackage{amsmath}
 \usepackage{amssymb}
 \usepackage{amsfonts}
 \usepackage{algorithm}
 \usepackage{algpseudocode}
 \usepackage{graphicx}
 \usepackage{mslapa}
 \usepackage{adjustbox}
 \usepackage{csquotes}
 \usepackage{longtable}
 \usepackage{textcomp}
 \usepackage{booktabs}
 \usepackage{xtab}
 \usepackage{afterpage}
 \usepackage{lipsum}
 \usepackage{xcolor}
 \usepackage{array}
 \usepackage{color}
 \usepackage{multirow}
 \usepackage{soul}
 \usepackage{xfrac}
 \usepackage{caption}
 \usepackage{subcaption}
 \usepackage{scalerel}
 \usepackage{tikz}
 \usetikzlibrary{svg.path}
 \usetikzlibrary{matrix,positioning,quotes} \usepackage[colorlinks,bookmarksopen,bookmarksnumbered,citecolor=red,urlcolor=red]{hyperref}

\newcommand\BibTeX{{\rmfamily B\kern-.05em \textsc{i\kern-.025em b}\kern-.08em
T\kern-.1667em\lower.7ex\hbox{E}\kern-.125emX}}

\newcolumntype{L}[1]{>{\raggedright\let\newline\\\arraybackslash\hspace{0pt}}m{#1}}
\newcolumntype{C}[1]{>{\centering\let\newline\\\arraybackslash\hspace{0pt}}m{#1}}
\newcolumntype{R}[1]{>{\raggedleft\let\newline\\\arraybackslash\hspace{0pt}}m{#1}}

\newtheorem{Definition}{Definition}
\definecolor{googleblue}{rgb}{0.259,0.522,0.957}
\definecolor{googlegreen}{rgb}{0.060,0.620,0.350}
\definecolor{googlered}{rgb}{0.859,0.267,0.220}
\definecolor{bleudefrance}{rgb}{0.19,0.55,0.91}
\definecolor{gulfblue}{RGB}{182,217,235}
\definecolor{gulforange}{RGB}{245,102,0}
\definecolor{tricol-orange}{RGB}{241,163,64}
\definecolor{tricol-white}{RGB}{247,247,247}
\definecolor{tricol-purple}{RGB}{153,142,195}

\def\volumeyear{2022}

\usepackage{orcidlink}

\def\BibTeX{{\rm B\kern-.05em{\sc i\kern-.025em b}\kern-.08em
    T\kern-.1667em\lower.7ex\hbox{E}\kern-.125emX}}

\begin{document}

\runninghead{Hepworth et al.}

\title{Swarm Analytics: Designing Information Markers to Characterise Swarm Systems in Shepherding Contexts}

\author{Adam J. Hepworth\orcidlink{0000-0003-0567-1965}\affilnum{1}, Aya Hussein\orcidlink{0000-0003-3204-0712}\affilnum{1}, Darryn J. Reid\affilnum{1,2} and Hussein A. Abbass\orcidlink{0000-0002-8837-0748}\affilnum{1}}

\affiliation{\affilnum{1}School of Engineering and Information Technology, University of New South Wales, Canberra, Australia\\
\affilnum{2}Defence Science and Technology Group, Adelaide, Australia}

\corrauth{Adam J. Hepworth, School of Engineering and Information Technology, University of New South Wales, Canberra, Australia}

\email{a.j.hepworth@unsw.edu.au}

\begin{abstract}
    Contemporary swarm indicators are often used in isolation, focused on extracting information at the individual or collective levels. Consequently, these are seldom integrated to infer a top-level operating picture of the swarm, its members, and its overall collective dynamics. The primary contribution of this paper is to organise a suite of indicators about swarms into an ontologically-arranged collection of information markers to characterise the swarm from the perspective of an external observer\textemdash, a recognition agent. Our contribution shows the foundations for a new area of research that we title \emph{swarm analytics}, whose primary concern is with the design and organisation of collections of swarm markers to understand, detect, recognise, track, and learn a particular insight about a swarm system. We present our designed framework of information markers that offer a new avenue for swarm research, especially for heterogeneous and cognitive swarms that may require more advanced capabilities to detect agencies and categorise agent influences and responses.
\end{abstract}


\maketitle

\section{Introduction}

Artificial agents sit between two theoretical extremes; reactive and cognitive agents~\shortcite{Ferber:1999:MSI:520715}. Reactive agents have direct mappings from their sensorial information to their actuators. Cognitive agents embed an architecture that sits between the inputs and outputs, performing deliberate planning and \emph{thinking}. In practice, agents are often designed to be sitting between these two extremes, with some aspects of their behaviours sitting more on the reactive side while others are on the cognitive side. The decision on the architecture is influenced by many aspects, including the availability of fast models to act as shortcuts between the inputs and outputs and the complexity of the operating environment~\shortcite{9256255}. Research remains ongoing to clearly define boundaries for architecture classification, further compounded by complexities related to artificial agents~\shortcite{McGivern2020:Cognition}.

Swarm systems consist of artificial or biological agents whose joint action displays order and coordination in time and space. A classic example of a swarm is bird flocking, fish schooling and sheep herding~\shortcite{Reynolds1987Boids}. Nearly all of the literature on swarm systems rely on reactive agents. The simplicity of these agents comes with advantages in real-world situations, including light computations, speed and simplicity in the logic used inside each agent for transparency of individual behaviours. Despite this simplicity, the swarm as a whole displays complex self-organised behaviours. The non-linear dynamics that aggregate the behaviour of individuals into the behaviour of the whole can hardly be reversed; leading to a few challenges described below:

\begin{enumerate}
\item How to guide and control the swarm without impacting intra-swarm dynamics?
\item How to explain the swarm's performance to an external human observer?
\item How to make the individuals smarter without increasing the complexity of the internal logic of a swarm member?
\end{enumerate}

One solution for the first challenge lies in \emph{shepherding}~\shortcite{Long2020:Comprehensive}, a bio-inspired swarm guidance method that mimics how sheepdogs guide a swarm of sheep. The concept of shepherding for swarm guidance has been applied in many applications including agriculture~\shortcite{Strombom:2014}, crowd control~\shortcite{10.1007/978-3-642-34381-0_48,MouldCounterTerrorActivityRecognition}, and uninhabited vehicle (UxV) navigation~\shortcite{Abbass2020:UxV}, and has proved viable in limited  communication settings~\shortcite{9504706}. The two remaining challenges also exist in shepherding research. Long et al.,~\citeyear{Long2020:Comprehensive} note that there is a scarcity of tools to analyse the interactions between the sheep, between the shepherds and between the sheep and shepherds, going on to discuss the need to understand influence vectors amongst agents, where analytical tools such as social network analysis may be viable~\shortcite{Long2020:Comprehensive}. Identifying the critical pieces of information which discriminate particular states or infer specific strategies is difficult without domain knowledge, requiring complex transformations of signal data. Designing the space of information and features to focus on can often be complex, requiring substantial domain contextualisation with features crafted bottom-up at the instance level.

The second challenge motivated a line of research on activity recognition of human-swarm interaction~\shortcite{Hepworth2021:ARS}, as well as designing ontologies to represent the space of concepts lying between humans and the swarm~\shortcite{abbass2020a,Baxter:HST-GO,Onto4MAT:2022}, contributing to a holistic theory to inform how humans and swarm should interact~\shortcite{Hasbach2021:HSI}. The third challenge motivated the design of contextual indicators that could be extracted from the sensorial information to guide the swarm. These indicators could inform the three challenges. A preliminary attempt is presented in~\shortcite{Hepworth2020:Footprints}.

This paper attempts to answer the question: \emph{what indicators can we purely design from the positional information of the swarm to inform a dashboard on the collective behaviour of the swarm?} Answering this question contributes to all three challenges above. The indicators could inform the swarm's guidance, explain swarm performance to an external observer, and create smarter individuals within a swarm. Positional information are the only pieces of information required by almost all reactive agents in the swarm literature. By relying only on positional information, we do not overload the swarm with further requirements, such as additional sensors. Swarm information markers are complementary to research first proposed by Matari\'c~\citeyear{Mataric1995:AdaptiveGroup}, who explored \enquote{common properties across various domains of multiagent interaction for the purpose of classifying group behaviour}~(pg.52), introducing the idea of a \emph{basis behavior} to describe agent interactions at the spatial level.

The proposed swarm markers offer three extra advantages. The first is through the lens of the swarm agents, enabling activity recognition of other agents and the collective~\shortcite{Baxter:HST-GO}. The second is through the lens of an external observer who can classify behaviours and infer intents~\shortcite{Hepworth2021:ARS}. The third and shared between the first two advantages is a requirement to enhance an agent\textquoteright s situational awareness~\shortcite{Abbass2020:UxV} to develop individual and collective understanding. By using a suite of markers, an agent could polarise its attention to particular aspects in the environment by using a subset of the markers.

The remainder of this paper is organised as follows. In the following section, we present a review of contemporary swarm modelling approaches that highlight the methods and techniques to analyse swarm systems, focusing on measures as indicators with discriminatory power. We then structure the problem space and provide supporting definitions before introducing information and swarm markers. Following this, we discuss our proposed situation recognition system of swarm markers and highlight critical challenges. Next, our Experiment Design and Analysis sections present a systematic experiment to evaluate the swarm markers. Finally, we conclude the paper with a discussion on open research questions for future investigation.

\section{Background Materials}

This section is structured into two sub-sections. In the first sub-section, we present a high-level summary of indicators to analyse a swarm system covering geometric, spatial, information-theoretic, time series, physics and graph-based indicators. We then use three lenses to look at the literature. An individual-agent lens focuses on individuals and their traits; an influence lens focuses on the role of an agent in a group, including leadership and followership; and an emergence lens, where the focus is on the global observable dynamic of the swarm as a whole. Our literature review has identified over 40 methods, techniques and measures.

\subsection{Swarm Indicators}

The literature on indicators of swarm behaviour is multi-disciplinary, with some indicators focusing on extracting information on individuals in the swarm, while others focus on the swarm's interaction level and aggregate level as a whole. In addition, some indicators rely on information theoretic foundations, while others utilise theories in physics, time-series analysis and graph theory.

Indicators that focus on characteristics of individuals in a swarm tend to analyse information on an agent\textquoteright s level, such as angular velocity~\shortcite{Hepworth2021:ARS}, speed, and acceleration. Some indicators are borrowed from the biological literature, such as the Overall Dynamic Body Acceleration (ODBA), an integrated measure of body motion in the three spatial dimensions~\shortcite{ODBA:2011}. These individual-based indicators usually act as raw indicators that get used in more complex ones, such as information-theoretic indicators~\shortcite{crosatoSwarmIntell}.

Agent interaction indicators are concerned with capturing the dynamics among agents, the relationships between swarm agents and the collective, and between swarm agents and external agents, such as a control agent. This includes measures such as distance to a global or local swarm centre of mass or level of alignments between an individual and its neighbours; neighbours here can be the closest $k$, or the number of agents within a  sensing range. These indicators are used in swarm control methods, including the seminal works of Reyolds~\citeyear{Reynolds1987Boids} and Str\" {o}mbom et al.,~\citeyear{Strombom:2014}. Others relied on observations from biological field trials~\shortcite{Yaxley2020SkyShepherding} to derive more systemic indicators that capture high-level interaction such as predation risk and situation awareness~\shortcite{Hepworth2020:Footprints}. Predation risk is designed to illuminate a swarm\textquoteright s proximity to a predator relative to the configuration of the swarm, whereas situation awareness captures the amount of obstruction between an agent and a predator.

A broad selection of information-theoretic measures are used to analyse swarm systems, often to qualitatively describe swarm dynamics~(pg.115)~\shortcite{Bossomaier2016}. Information-theoretic analyses often seek to quantify the information transfer in a swarm, demonstrating the flow of information through time. Transfer Entropy and its derivations are widely adopted~\shortcite{Bossomaier2016}, often selected because of the intuitiveness of its interpretation and the established body of research use~\shortcite{Miller2014,10.1371/journal.pone.0040084,crosatoSwarmIntell,pilkiewicz:2020,Porfiri2018}. Transfer Entropy is a model-free, non-parametric approach that measures the directed information flow from a source to a target process~\shortcite{Bossomaier2016}. Derivations of TE often seek to answer specific questions on the swarm, be it looking at the aggregate as with Global Transfer Entropy~ (average collective Transfer Entropy)~\shortcite{Bossomaier2016}, or individual level~\shortcite{crosatoSwarmIntell,Bossomaier2016}. A complementary measure to that of Transfer Entropy is Information Storage, capturing the amount of historical information relevant to predicting the future state of a process~\shortcite{10.1371/journal.pone.0040084}. Other entropic formulations are also employed, for example ranging from classic Shannon Entropy to investigate emergent behaviour~\shortcite{Hamann2012:Emergence}, cross-entropy to evaluate swarm robustness~\shortcite{e22060597} or causation entropy to identify causal relationships~\shortcite{7809221}.

Time series analysis techniques are often used to develop a higher-order understanding of what the swarm and its agents are doing. For example, Dynamic Time Warping (DTW) is used to infer agent leadership traits in a collective~\shortcite{mFLICA:2021}. Spectral analysis is highlighted as a technique to evaluate collective behaviour in crowds, for instance, applying spectral-based techniques to determine motion dynamics by measuring flow-field information~\shortcite{1698861}. Finally, complexity measures are employed to investigate causality, such as the compression-complexity causality~\shortcite{Kathpalia:2021}, based on the effort-to-compress measure~\shortcite{ETC:2013}.

Physics-based approaches are distinct from other methods in that they treat the swarm as a continuous collective, in contrast to techniques discussed that consider the swarm as an aggregate of individual agents. Haeri et al.,~\citeyear{10.1115/1.4046580} employ a thermodynamics approach to assess collective behaviour, using the context of fluid flow to define macroscopic swarm states~\shortcite{10.1115/1.4046580}. Such approaches are aimed to enable more accessible state information representation and classification of emergent behaviours, especially for unknown swarms~\shortcite{10.1115/1.4046580}. Mavridis et al.,~\citeyear{SwarmInteractions:2021} investigate coordinated movements of swarms, proposing a scheme to infer the laws for inter-agent coordination by observing the swarm density evolution over time~\shortcite{SwarmInteractions:2021}.

Graph-theoretic approaches provide a connectivity lens to analyse agents, swarms and their dynamics and infer the influence between agents in a swarm. Shang \& Bouffanais~\citeyear{Shang2014:SwarmTopo} analyse biological swarms using network and graph theoretic approaches, noting that the predominant approach to swarm model development has been in \enquote{generating consensus behaviors, often in the form of group alignment or polarization}~(pg.5)~\shortcite{Shang2014:SwarmTopo}. Res\'{e}ndiz-Benhumea et al.,~\citeyear{10.1162/isal_a_00229} study a swarm robotic system inspired by biological systems. The approach integrates social network analysis with agent-based modelling to investigate swarm influence and emergent dynamics, suggesting that social network analysis can lead to a better understanding of the emergent properties in swarms~\shortcite{10.1162/isal_a_00229}.

\subsection{Categories of Swarm Analysis}

\begin{figure}[ht]
    \centering
    \begin{adjustbox}{width=0.5\textwidth}
        \begin{tikzpicture}
        	\begin{scope}[fill opacity = .5]
                \draw[fill=tricol-orange, draw = black] (-1.5,1) circle (3);
                \draw[fill=tricol-purple, draw = black] (1.5,1) circle (3);
                \draw[fill=tricol-white, draw = black] (0,-1.5) circle (3);
            \end{scope}
            \node at (-3,5)[align=center,text width=4cm]{\textit{Leadership, Coordination and Influence}};
            \node at (3,5)[align=center,text width=4cm]{\textit{Dynamics and Emergent Behaviour}};
            \node at (0,-5)[align=center,text width=4cm]{\textit{Agents and Individual Characterisation}};

                \node at (-2,1.5)[align=left,text width=3cm]{$A$};

                \node at (2,1.5)[align=right,text width=2cm]{$B$};

                \node at (0,-3)[align=center,align=center,text width=5cm]{$C$};

                \node at (0,2)[align=center,text width=2cm]{$D$};

                \node at (-1,-1)[align=left,text width=2cm]{$E$};

                \node at (1.75,-1)[align=center,text width=2cm]{$F$};

                \node at (0,0)[align=center,text width=2cm]{$G$};
        \end{tikzpicture}
    \end{adjustbox}
    \caption{Synthesis of swarm and related intelligent agent literature, depicting the prevalence of approaches across the three focal lenses identified. The literature identified in each segment is presented in Table~\ref{table:VennDiagram}.}
    \label{fig:VennDiagram}
\end{figure}
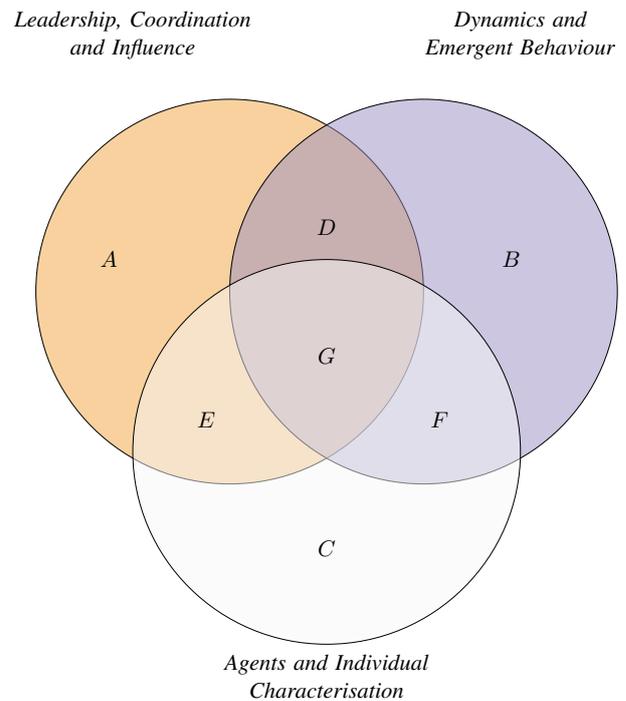

In this section, we group the indicators into three categories, presenting information on the swarm from a particular lens. Our lenses cover the three groups of information required to characterise a swarm: individual traits, the role of an individual in a group, and group dynamics. The three lenses, when combined, offer an overall picture of the swarm. We represent the methods and measures contained in Table~\ref{table:VennDiagram} as a Venn Diagram of categorisations in Figure~\ref{fig:VennDiagram}. This figure describes the use of source literature in one or more categories of analysis for swarms, highlighting the distribution present. An extension of Figure~\ref{fig:VennDiagram} and Table~\ref{table:VennDiagram} is given in the appendix at Table~\ref{table:AnalysisLiterature}. We summarise each lens below.

\begin{table*}[ht]
    \centering
    \begin{tabular}{L{6em} L{40em}}
        \toprule
        \textbf{Segment} & \textbf{Literature} \\
        \midrule
        \rule{0pt}{4ex} $A$  & Mocanu et al.,\citeyear{6973878}, Surasinghe \& Bollt~\citeyear{e22040396}, Nagaraj et al.,~\citeyear{ETC:2013}, Kathpalia~\citeyear{Kathpalia:2021}, Spinello C \& M~\citeyear{spinello:2019} , Mavridis et al.,~\citeyear{SwarmInteractions:2021}, Lord et al.,~\citeyear{7809221}, Pilkiewicz et al.,~\citeyear{pilkiewicz:2020}, Res\'{e}ndiz-Benhumea et al.,~\citeyear{10.1162/isal_a_00229}, Wang et al.,~\citeyear{10.1371/journal.pone.0040084}, Bossomaier et al.,~\citeyear{Bossomaier2016}, Papaspyros et al.,\citeyear{10.1371/journal.pone.0220559} \\
        \rule{0pt}{4ex} $B$  & \rule{0pt}{4ex}Wu et al.,~\citeyear{Wu2011:Emergence}, Reynolds~\citeyear{Reynolds1987Boids}, Jankovic~\citeyear{Jankovic2018:CFD}, Puckett et al.,~\citeyear{PhysRevLett.114.258103}, Haeri et al.,\citeyear{10.1115/1.4046580},~Hamann et al.,~\citeyear{Hamann2012:Emergence}, Gleiss et al.,~\citeyear{ODBA:2011}, Martín López et al.,~\citeyear{10.1111/2041-210X.13751}, Andrade et al.,\citeyear{1698861}, Bossomaier et al.,~\citeyear{Bossomaier2016}, Cofta et al.,~\citeyear{e22060597}, Wang et al.,~\citeyear{Wang2011MeasuringIS}, Baldi \& Frasca~\citeyear{BALDI2019935} , Brown \& Goodrich~\citeyear{10.5555/2615731.2615798}, Traboulsi \& Barbeau~\citeyear{9070871} \\
        \rule{0pt}{4ex} $C$  & \rule{0pt}{4ex}Gleiss et al.,~\citeyear{ODBA:2011}, Martín López et al.,~\citeyear{10.1111/2041-210X.13751}, Hepworth~\citeyear{Hepworth2021:ARS}, Andrade et al.,\citeyear{1698861}, Schaerf et al.,~\citeyear{Schaerf:2021}, Hepworth et al.,~\citeyear{Hepworth2020:Footprints}, Str\" {o}mbom et al.,~\citeyear{Strombom:2014}, Valentini et al.,~\citeyear{Valentini:2019b}, Chakraborty et al.,~\citeyear{Chakraborty2020}, Abbass and Hunjet~\citeyear{Abbass2020:UxV} \\
        \rule{0pt}{4ex} $D$ & \rule{0pt}{4ex}Amornbunchornvej~\citeyear{mFLICA:2021}, Mateo et al.,~\citeyear{Mateo:2017}, Bossomaier et al.,~\citeyear{Bossomaier2016}, Lord et al.,~\citeyear{7809221}, Shang \& Bouffanais~\citeyear{Shang2014:SwarmTopo} \\
        \rule{0pt}{4ex} $E$ & \rule{0pt}{4ex}Crosato et al.,~\citeyear{crosatoSwarmIntell} \\
        \rule{0pt}{4ex} $F$ & \rule{0pt}{4ex}Hepworth et al.,~\citeyear{Hepworth2020:Footprints}, Wang et al.,~\citeyear{10.1371/journal.pone.0040084}, Crosato et al.,~\citeyear{crosatoSwarmIntell}, Bossomaier et al.,~\citeyear{Bossomaier2016}, Li et al.,~\citeyear{Li2004:CollabSwarm} \\
        \rule{0pt}{4ex} $G$ & \rule{0pt}{4ex}Amornbunchornvej~\citeyear{mFLICA:2021},Wang et al.,~\citeyear{10.1371/journal.pone.0040084},~Bossomaier et al.,~\citeyear{Bossomaier2016}, Porfiri~\citeyear{Porfiri2018}, Spinello C \& M~\citeyear{spinello:2019}, Mert Karakaya et al.,~\citeyear{karakaya:2020}, Valentini et al.,~\citeyear{Valentini:2019a}, Butail et al.,~\citeyear{PhysRevE.93.042411} \\
        \bottomrule
    \end{tabular}
    \caption{Synthesis of swarm and related intelligent agent literature, depicting the prevalence of approaches across the three focal lenses identified: leadership, coordination and influence; dynamics and emergent behaviour; and agents and individual characterisation. Segments sets are listed in reference to Figure~\ref{fig:VennDiagram}, with an exhaustive summary presented in Table~\ref{table:AnalysisLiterature}.}
    \label{table:VennDiagram}
\end{table*}

\subsubsection{Agents and individual characterisation.}

This category includes research focused on swarm parameterisations and investigations of agent decision models, abilities and traits. Swarms containing homogeneous agents are most prevalent in the literature; for example, the seminal formulation of Reynolds~(1987) relies on homogeneous agents. Recently, heterogeneous swarm formulations have gained more attention as complex swarm behaviours can be generated from simple heterogeneous behaviours~\cite{Kengyel:2015} to develop new agent types or re-parameterise existing agents in a swarm. Research into swarm heterogeneity is consistent with literature from the biological shepherding domain. Williams~\citeyear{Williams2007:working} characterises different individual abilities and traits of a herding agent (swarm control agent\textemdash a sheepdog), noting that these are the markers to identify how well-trained a herding agent is. Classifying distinct behaviours within heterogeneous swarms has been explored, such as by Hepworth et al.,~\citeyear{Hepworth2020:Footprints} who employed an Information Theoretic approach to distinguish between swarm agent types, based on the underlying model introduced by Str\" {o}mbom et al.,~\citeyear{Strombom:2014}. The approach was to parameterise sensing and interaction weights amongst agents, identifying the impact on a swarm. Szwaykowska et al.,~\citeyear{7063970} analyse agents with heterogeneous capabilities, where agent decision capabilities are homogeneous, but the interaction dynamic weights are not. Kengyel et al.,~\citeyear{Kengyel:2015} analyse four behaviour types in a biological swarm, identifying that complex behaviour can be generated from simple heterogeneous behaviours.

\subsubsection{Leadership, coordination and influence.}

This category includes studies that seek to uncover leadership and followership roles within swarms, understand coordination mechanisms in both biological and simulated swarms, and determine causal interactions of influence. Understanding influence responses may help design biologically-inspired agents to serve more complex swarm applications. For example, Yaxley et al.,~\citeyear{Yaxley2021:SS} discuss the roles of leaders, followers and uncooperative followers in biological shepherding and Duikman~\citeyear{Diukman:2012} characterise the underlying organisational leadership and followership structures of a swarm. Butail et al.,~\citeyear{PhysRevE.93.042411} and Basak~\citeyear{Basak2021:leaderFollower} employ information-theoretic approaches with biological agents, with Butail et al. successfully inferring leadership in zebrafish pairs using trajectory data. Porfiri~\citeyear{Porfiri2018} suggests that Information Theory offers a robust framework for the objective analysis of cause-effect relationships using raw data (e.g., behavioural observations or individual trajectory tracks). This is supported by a range of experimental studies with similar analysis approaches, for instance~\shortcite{crosatoSwarmIntell,Hepworth2020:Footprints,Bossomaier2016,karakaya:2020}.

\subsubsection{Swarm dynamics and emergent behaviour.}

This category includes work that seeks to uncover rules for individual and collective movement, analysing emergent properties of the swarm from seemingly simple interactions (for instance, see Reynolds~\citeyear{Reynolds1987Boids}). Learning swarm behaviours is vital to understanding how individual agents cooperate to achieve a global, swarm-level behaviour~\shortcite{8444217}. A common approach to the behaviour recognition problem is to observe features of the swarm through time (sensor-based recognition), for instance, characterising underlying swarm interactions~\shortcite{doi:10.2514/1.G004115,8444217}, quantifying the strength and asymmetry of interaction dynamics~\shortcite{Hepworth2020:Footprints}, or investigating how information propagates~\shortcite{10.1371/journal.pone.0040084,sipahi:2020}. Model-based approaches identify particular typical and a-typical swarm behaviours~\shortcite{brown2014a}. Information Theory is used by Liu et al.,~\citeyear{liu2018a} to detect emergence over time, identifying intervention opportunities to influence the swarms resulting state. Wang et al.,~\citeyear{Wang2011MeasuringIS} investigate the propagation of information through swarms with an Information Theoretic framework, demonstrating that such measures can be applied to non-trivial models to reveal the dynamics that cannot otherwise be visually detected.

\section{Methodology Conceptualisation}

The primary scope of this work is the existence of an agent external to the swarm with interest in understanding what the swarm does. In particular, we call a swarm agent as a \emph{sheep} and the swarm controller as a \emph{sheepdog}~\shortcite{baumann2016learning}. We assume that the sheepdog\textquoteright s interest is to understand the swarm and its influence on the swarm. We denote the swarm controller agent (sheepdog) as $\beta$, and swarm agents (sheep), given by $\Pi = \{\pi_1, \pi_2, \dots, \pi_N\}$. Both $\beta$ and $\pi$ sense raw data from their independent sensors, process this data to transform it into information, decide what to do with it, and generate an action. Each $\pi$ is reactive to other agents, employing a combination of attraction and repulsion actions (cohesion, separation, and alignment~\shortcite{Reynolds1987Boids}) to actuate. The agent $\beta$ is also reactive, employing a combination of collect and drive actions to actuate~\shortcite{Strombom:2014}, guiding $\Pi$ towards a designated goal location. The actions of each agent may manifest as an influence on another agent, transmitted through the environment as a type of information\textemdash, a force vector. When $\beta$ positions itself, the influence vector is a portion of the total information propagated $\beta\rightarrow\pi$. The resulting action of $\pi$ is, in part, a response to an action of $\beta$.

\begin{figure}[ht]
    \centering
    \includegraphics[width=0.5\textwidth]{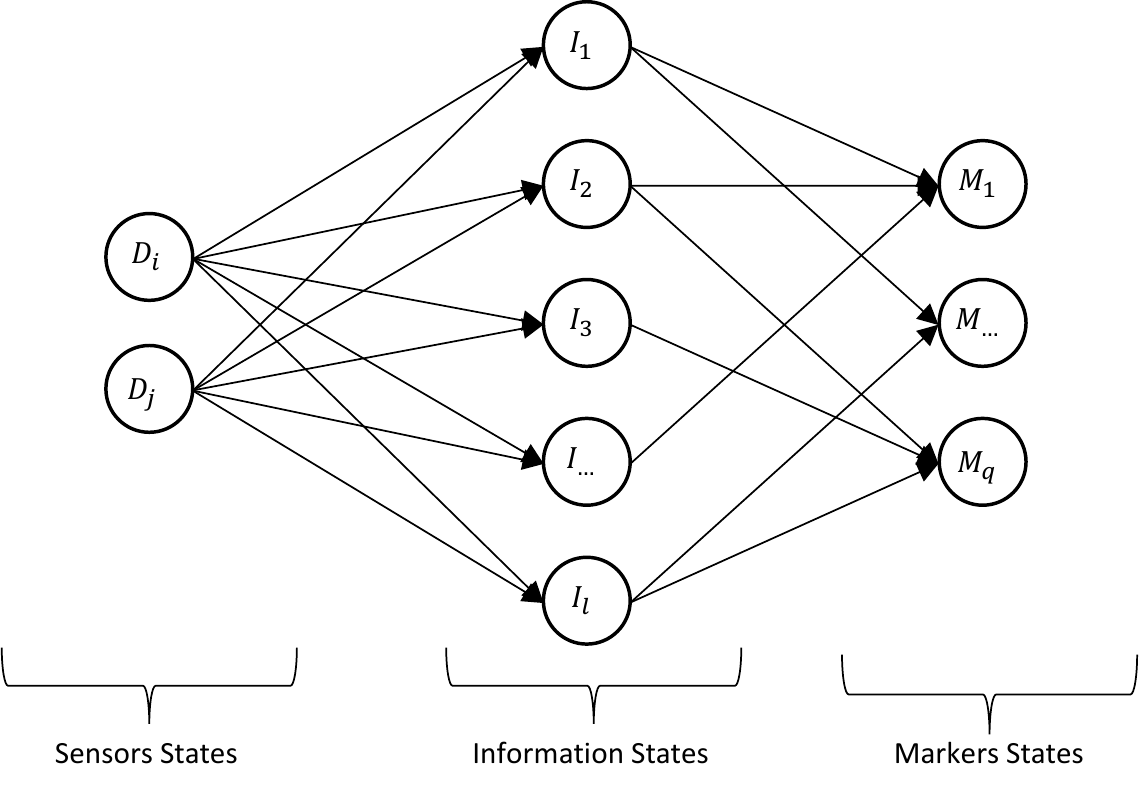}
    \caption{The flow of information from $D\rightarrow\mathcal{I}\rightarrow\mathcal{M}$, where $D$ are the raw positional data ($P^t$), $\mathcal{I}$ are operations on the raw data, generating information features, and markers are transformations of information that act on the intermediate representation, $\mathcal{I}$.}
    \label{fig:StateFlowDiagram}
\end{figure}

The context of this paper is whether or not we can design indicators; we call them \emph{information markers}, to detect state information on the swarm from their positional data. In effect, information markers transform functions with pre-defined meanings in a domain. They enable the recognition of situations and contexts by detecting information in the three categories presented in the previous section: information about a single agent, information about the role and behaviour of an agent relative to others, and information on the global dynamics of the swarm. We use positional data as a single information type, which could be obtained in a real-world situation using one of many sensors, including vision-based sensors, LIDAR, or even a remote sensing system. Using a single sensor source in context recognition is prevalent in both simulation and real-world studies (for example, see Table~3 in Pernek \& Ferscha~\citeyear{Pernek2017}). Nevertheless, a single information type, such as the position of an agent, is used to calculate multiple pieces of information, such as the speed and acceleration of an agent, the centre of masses for groups of agents, and the speed and acceleration of groups. The state flow from raw data to information to information markers is depicted in Figure~\ref{fig:StateFlowDiagram}.

Figure~\ref{fig:StateFlowDiagram} depicts the conceptual flow of data to information to information markers. For example, positional information are considered the data in this particular instance. One can then transform these into velocity and acceleration vectors and aggregate these on a group level. These are the classic information used by Boids and Shepherding for an agent to act. We call these information ''states''. Information markers take these information states and generate situation states as a higher-order aggregate state of information. We call these ''concepts'', which then form an ontology. These ontologies inform recognition systems to distinguish particular situations in a swarm. For example, a sheepdog must collect or drive the sheep in a classic shepherding problem in the most specific setting.

These two situations are characterised by particular information markers on whether or not an astray sheep exists and whether the sheep are grouped. An external observer may be interested in other information, such as whether or not the level of energy in the sheep is diminishing or whether or not the sheep\textquoteright s actions are coordinated. Information markers offer objective measures of state information with discriminatory power to reach these conclusions. Information markers offer benefits such as illuminating what is occurring, providing historical understanding, warning of potential dynamics change, identifying individual and collective risk factors, and uncovering causal factors of influence. More concisely, information markers aim to leverage historic positional information to illuminate what is occurring and what is expected to occur regarding risk factors or potential dynamics change.

\subsection{Formal Definitions}

We provide formal definitions that illustrate information flow in the recognition system. We define an external observer ($\kappa$) as an agent that is not allowed to actuate or produce an influence vector in the system but can receive information from the system and with interest in understanding what the system is doing. This assumption of passivism of $\kappa$ ensures that $\kappa$ can understand the system without needing to be proactive about probing it. We define $\mathfrak{s}^t_{\pi_i}$ to be the state vector for agent $\pi_i$ at time $t$ and $\mathfrak{s}^t_\Pi$ to be the state vector of the swarm, $\Pi$, at time $t$.

We use a classic definition of data and information, where \textbf{data}, $\mathcal{D}$, \textquote{\shortcite[p.~71]{Davis2000}~{consists of representations of events, people, resources, or conditions. The representations can be in various forms, such as numbers, codes, text, graphs, or pictures}}. \textbf{Information}, $\mathcal{I}$, {\shortcite[p.~71]{Davis2000}}~{is a result of processing data. It provides the recipient with some understanding, insight, conclusion, decision, confirmation, or recommendation}, that is $\mathcal{I}\leftarrow~F(\mathcal{D})$, with $F$ a vector function transforming $\mathcal{D}$ into $\mathcal{I}$. An agent transforms data into information that it can use to generate actions. While these actions are outputs by agents, they are also the sensed data by agents. Hence, we can generalise the behaviour of an agent to be a set of information. If $\mathcal{I}$ is the superset of information an agent possess, then a behaviour $\Sigma_k$, is a subset of this set transformed into actions, $\Sigma_k=\{g(\mathcal{I}_1),g(\mathcal{I}_2),\dots,g(\mathcal{I}_n)\}$. The set of all $d$ behaviours in a system is denoted $\Sigma=\{\Sigma_1, \dots, \Sigma_d\}$.

\begin{Definition}\label{def:behaviour}
        \textbf{Behaviour} is a label associated with a set of information, $\Sigma_k = \{ g(\mathcal{I}_1), g(\mathcal{I}_2), \dots, g(\mathcal  {I}_n) \}$, describing the actions displayed by an agent, $k = 1 \dots, d$.
\end{Definition}

Behaviours in our formulation are ontological concepts/labels, associating some contextual meaning to particular pieces of information, $\Sigma\subseteq\mathcal{I}$. As a form of information, behaviours may be considered observable messages between agents.

An \emph{Information Maker} is a subset of information that could reveal aspects of an agent\textquoteright s behaviour; thus, an information marker ($\mathcal{M}$) possesses some semantics to recognise the situation an agent is facing and the corresponding actions it generates to handle these situations.

\begin{Definition}\label{def:infoMarker}
    An \textbf{Information Marker}, is a set of transformed information $\mathcal{M}_l = \{ f(\mathcal{I}_1), f(\mathcal{I}_2), \dots, f(\mathcal{I}_m) \}$, correlated with a subset of the information in $\Sigma$. Information markers are information that can reveal the presence or absence of a particular behaviour.
\end{Definition}

We assume that as more marker states are obtained towards the complete set of $\mathcal{M}_l$ required to identify a particular behaviour $\Sigma_k$ perfectly, that opportunity exists to anticipate (predict) response states and behaviours. Prior to the complete set of $\mathcal{I}~\in~M$, each new $\mathcal{M}_l$ identified reduces the \emph{search space} for the plausible $\Sigma$ which may be observed. The sequence of markers identified contributes to an explanation of \emph{why} a particular $\Sigma$ has been identified, offering an opportunity to detect the early presence of an influence event and post-event to assist with an explanation of why a particular decision was made, or behaviour was completed. In scenarios requiring increased system transparency, $\mathcal{M}$ may fulfil the requirements to evaluate a system and report to the user~\shortcite{Hepworth2021:HST3}, providing a quantitative way to measure tenets of transparency.

Before arriving at our final definition in this section, two concepts require further discussion: contexts and situations. First, we will define a context to be the effective superset of information in a problem space; with that, we mean if the problem space consists of a system and the environment it is operating within, then a context is all information in the system and its environment that are needed to operate the system and the environment, including different constraints and goals. By effective, we mean that information not used by the system or the environment is excluded.

\begin{Definition}\label{def:context}
    A \textbf{context}, $\mathbb{C}$, is the effective superset of information, $\mathcal{I}$, required by a system and its environment to operate autonomously.
\end{Definition}

A context may contain sub-contexts. An agent may not have access to information to know the actual context; instead, we define $\mathbb{C}^o$ as the observable context by an agent. Situations representing information subsets do not change for a period of observation and get repeated in a context. Situations have a timeless property. Situations \enquote{are ultimately founded on objects, their properties, relations and the occurrences they participate in}~{\shortcite[p.~32]{8423994}}, where the subset of information they represent is unique for that circumstance and time.

\begin{Definition}\label{def:situation}
    A \textbf{situation}, $s$, is an invariant subset of $\mathcal{I}$ over a period of time, $t$, given as $\mathcal{I}_s^t$~\shortcite{8658079}.
\end{Definition}

In classic shepherding, herding is a context. Within this example context, a sheepdog recognises and acts on two situations. The first situation is when the sheep are clustered, the sheepdog needs to drive the sheep towards the goal. The second situation is when an astray sheep is away from the flock, wherein the sheepdog needs to collect that sheep towards the flock. It is important to note that a situation is associated with a system boundary; that is, the situation from a sheepdog perspective is invariant information in those held by the dog, while a situation from an external observer would be an invariant subset in the information held by the external observer. Relating this description to Definition~\ref{def:context}, herding as the context contains the unique information required to instantiate a particular situation, recognise an element, or bound an environment.

A type of information marker in our problem is a \emph{swarm marker}, which has value in inferring context towards a situation on swarms. Swarm markers are used to make decisions about the swarm by an external observer; for instance, they could be used to
    \begin{itemize}
        \item Understand what the swarm is doing and the manifestation of influences in the swarm.
        \item Detect a category of agent types as traits, such as a weighting system for decision making.
        \item Focus the attention of an observing agent on some aspect of the swarm's behaviour.
        \item Overcome some of the observing agent's internal bias on how to look at the swarm.
    \end{itemize}

\begin{Definition}\label{def:swarmMarker}
    A \textbf{Swarm Marker} is an information marker, $\mathcal{M}^S$, about a swarm rather than the individuals in the swarm.
\end{Definition}

\begin{figure}[ht]
    \centering
    \includegraphics[width=0.5\textwidth]{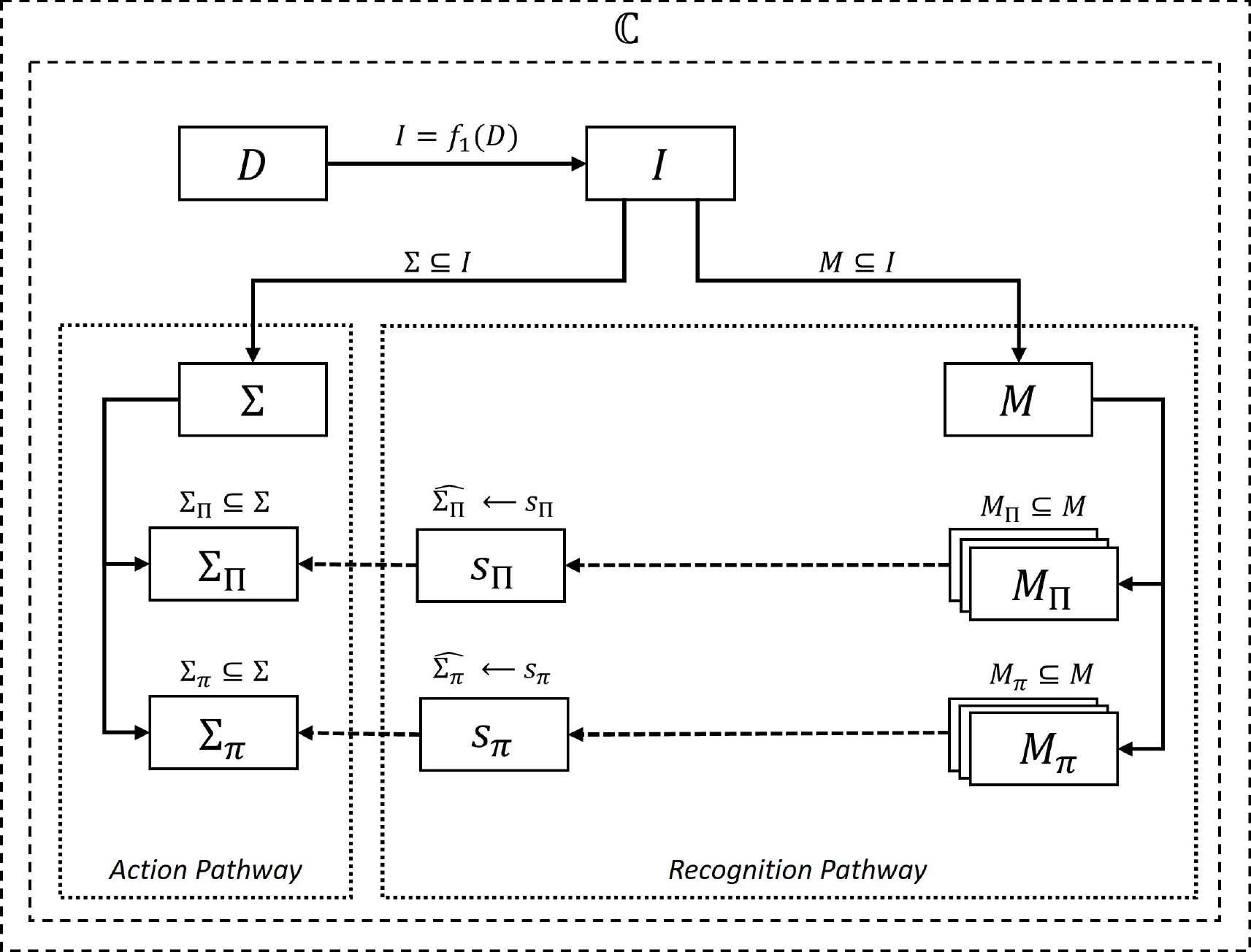}
    \caption{The system boundary of key concepts which describe the flow of computations from data ($D$) to information ($I$) to the correlation required between that information associated with behaviours ($\Sigma$) and those associated with information markers ($\mathcal{M}$). The overall figure represents all information forming the context ($\mathbb{C}$). Two pathways exist in the system: an action pathway, where an agent uses the information available to it to act, thus, generating behaviours, and a recognition pathway, where an agent uses the information available to it to create markers correlated with the behaviours it observes.}
    \label{fig:sysDefMap}
\end{figure}

Figure~\ref{fig:sysDefMap} summarises the relationships between the definitions discussed, highlighting the nuance between behaviours and markers as both transformations of information subsets. Information markers are \textit{indicators} which provide evidence for specific contexts, guiding the transformation of information. To define a formal relationship between $\Sigma$ and $\mathcal{M}$, we must first consider the perspective of each subset of information. Our first is the \textit{action} pathway, focused on the agent(s) under observation. The action agent uses particular aspects of the full information to make decisions about their actions in the system, displayed through behaviours.

The second perspective is that of an agent observing the actions of another, which is the \textit{recognition} pathway. The recognition agent uses a transformation of a subset of the total information available to calculate $\mathcal{M}$ from features of the information focused on a particular agent or swarm behaviour. The marker state vector for each swarm or agent is the \textit{context} which contributes to recognising a \textit{situation}. The situation is the estimated behaviour ($\hat{\Sigma}$) under observation, $s\rightarrow\hat{\Sigma}$. The key idea here is that markers are information transformations that identify the context(s) and situations to describe the observed behaviour.

\begin{figure}[ht]
    \centering
    \includegraphics[width=0.5\textwidth]{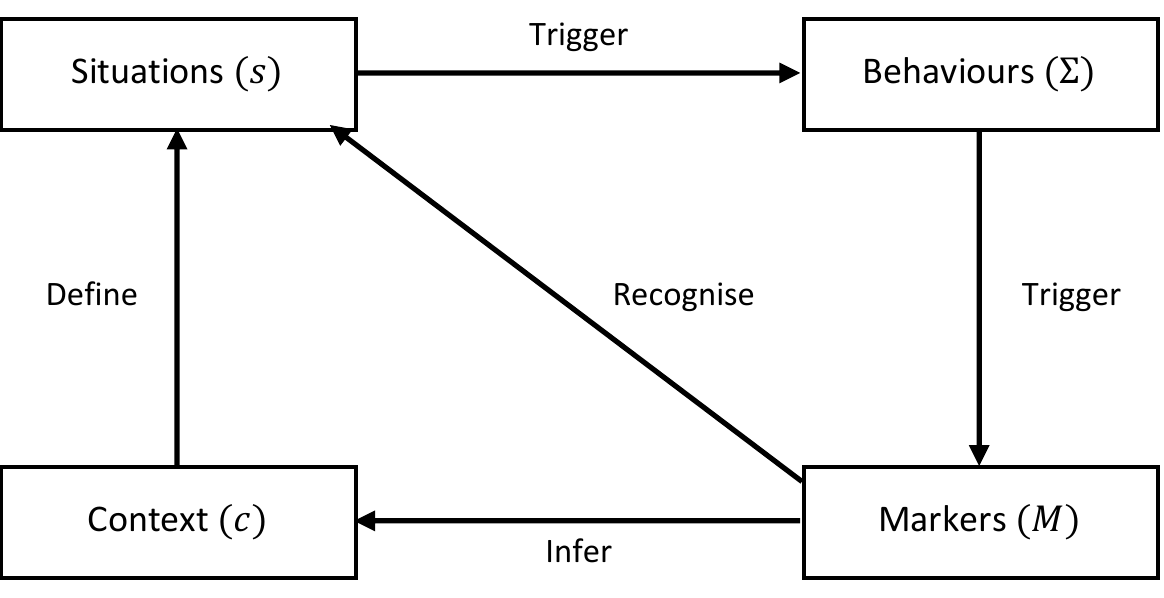}
    \caption{Conceptual relationships between the definitions introduced in the methodology conceptualisation section. This figure highlights the role of markers to \emph{recognise} situations and \emph{infer} contexts, being \emph{triggered} by behaviours of the swarm and its agents in an environment.}
    \label{fig:DefinitionsMapSmall}
\end{figure}

Figure~\ref{fig:DefinitionsMapSmall} depicts the links between each concept, showing how markers are designed to correlate with behaviours for recognising situations. A context contains a set of situations and acts as a set of constraints on situations. Situations trigger particular behavioural responses, necessitating a \emph{need} to act, triggering a particular set of markers. Markers provide a way to recognise a situation and allow context inference. The combination of each available marker generates the entire information situation for an agent, with each marker having a context that may or may not provide unique information \textemdash markers contain redundant information.

\section{Designing Information Markers}

After introducing key definitions in the previous section, we now synthesise our literature review and methodology conceptualisation discussions, connecting methods and measures from the swarm analysis literature and integrating these to describe their use. The primary opportunity for a swarm system is to apply markers as part of the recognition process. Such an approach may enable us to discover a system's causal rules and agent influences, offering potentially more robust strategies to deal with increased sensor noise~\shortcite{Hung2020:noise} and environmental complexities~\shortcite{9256255}. Our first task is to systematically select the appropriate markers that lead us to recognise agent contexts in the system. Our review of literature in the background section identifies five primary fields to analyse swarms across three swarm-focus areas. 

We first discuss the organisation of information markers for recognition, highlighting what constitutes a subset of information and the interdependencies between markers. We then present the markers selected in our study to recognise swarm situations and contexts. Designing an ontology of markers requires an understanding of what needs to be recognised for each category of indicators in the literature review. Table~\ref{table:2x2matrix} depicts a configuration of characteristics and swarm perspectives, synthesising the discussed literature review. The columns contain the agent perspective, being the individual agent and collective swarm levels, with rows containing the traits of either the individual or collective. Traits are categorised as either stationary or non-stationary.

It is essential to clarify that the \emph{value} of the method, measure or technique used to estimate information, be it stationary or non-stationary, can be dynamic. The main point of difference here is in what is being evaluated. Stationary information to be evaluated does not change in the environment, such as an agent's desire to be part of a group, a propensity to separate from a threat or other swarm members, and maximum speed. At the swarm level, stationary traits could include the number of swarm teams in the environment and the speed of the swarm. Non-Stationary information to be evaluated may change over time, such as an agent's relative propensity for leadership or followership, indicating the influence and interaction of an individual. At the swarm level, dynamic information could include collective actions and tactics. Discriminating characteristics at the individual and collective levels provide an opportunity to discover and exploit heterogeneous information in the swarm, providing novel insights.

\begin{table*}[ht]
    \centering
    \resizebox{0.85\textwidth}{!}{%
    \begin{tabular}{@{}lll@{}}
        \toprule
         &
          \multicolumn{1}{c}{\textbf{Individual Agent}} &
          \multicolumn{1}{c}{\textbf{Swarm}} \\ \midrule
        \textbf{Stationary Information Traits} &
          \begin{tabular}[c]{@{}l@{}}Transfer Entropy\\ Synchronicity\\ Situation Awareness\\ Predation Risk \\ Spatial Distance\\ Speed\\ Dynamic Body Acceleration\\ Acceleration\\ Angular Velocity\\ Spatial Measures\\ Correlation Function\\ Heading\end{tabular} &
          \begin{tabular}[c]{@{}l@{}}Transfer Entropy\\ Shannon Entropy\\ Spatial Distance\\ Speed\\ Frequency Analysis\\ Correlation Function\\ Heading\\ Acceleration\\ Dynamic Body Acceleration\\ Dynamic Time Warping\\ Granger Causality\\ Lyapunov Exponent\end{tabular} \\ \midrule
        \textbf{Non-Stationary Information Traits} &
          \begin{tabular}[c]{@{}l@{}}Transfer Entropy\\ Information Flow\\ Dynamic Time Warping\\ Correlation Function\\ Lyapunov Exponent\\ Frequency Analysis\end{tabular} &
          \begin{tabular}[c]{@{}l@{}}Transfer Entropy\\ Information Flow\\ Dynamic Time Warping\\ Correlation Function\\ Lyapunov Exponent\\ Nodal Analysis\\ Frequency Analysis\end{tabular}  \\ \bottomrule
    \end{tabular}
    }
    \caption{Synthesis of techniques identified the literature review (Table~\ref{table:VennDiagram} and Table~\ref{table:AnalysisLiterature}), organised by focus (individual agent or collective swarm) and trait type (stationary or non-stationary). This table provides the basis of rationale for information marker selection in the following design and analysis sections and informs further analysis techniques to uncover particular aspects of the swarm and its agents. Contained in the Appendix is a summary of select mathematical expressions and interpretations.}
    \label{table:2x2matrix}
\end{table*}

\begin{figure*}[ht]
    \centering
    \includegraphics[width=\textwidth]{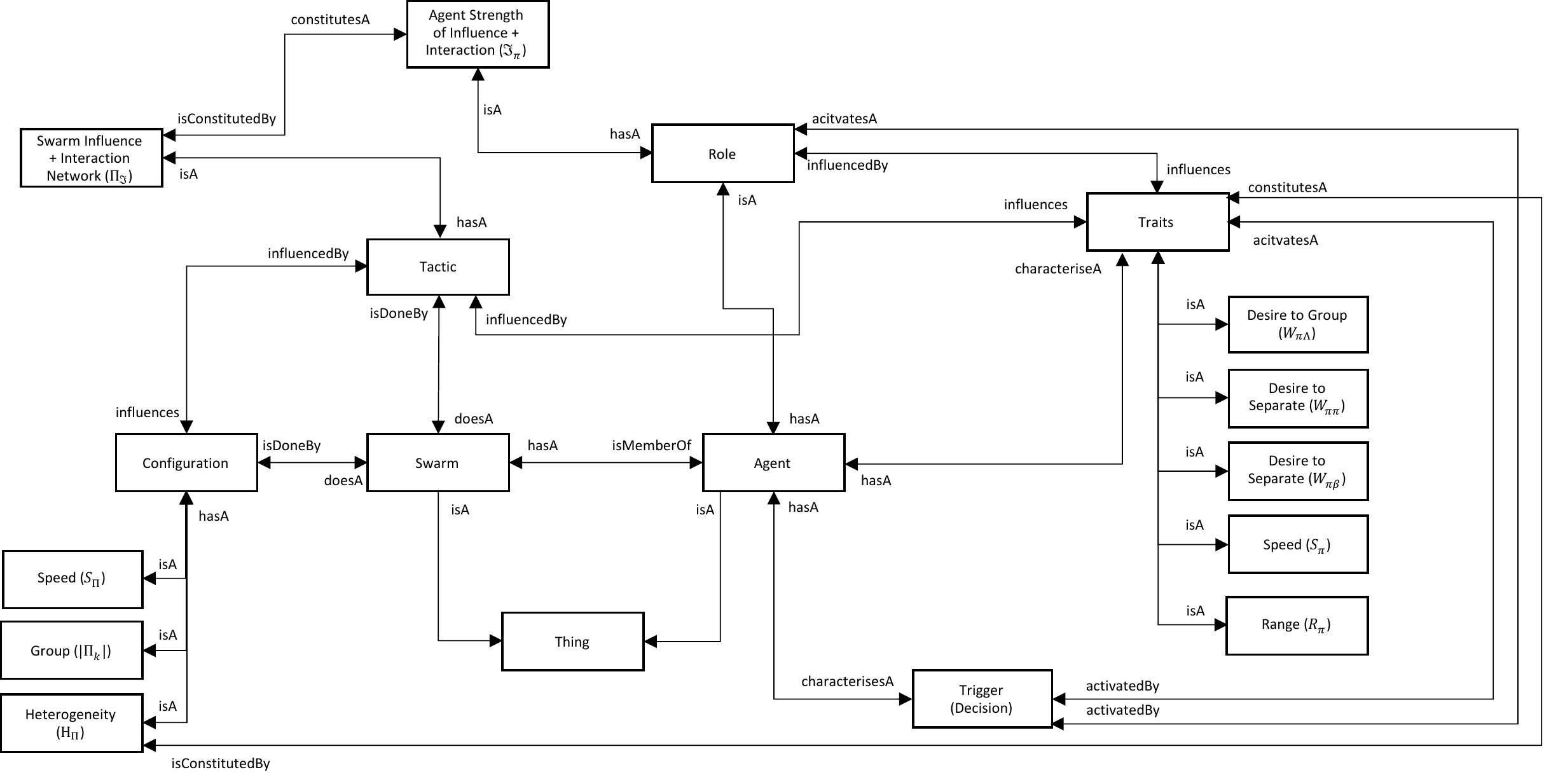}
    \caption{The ontological organisation of markers includes those aspects of the agent or swarm characteristics that are revealed (be these stationary or dynamic information traits), the information that the metric is acting on (for instance, being focused within an agent, on an agent property, or a link between multiple agents), and what information is used by each metric (the input elements of information).}
    \label{fig:MarkerOntology}
\end{figure*}

Following the identification of markers in the literature review and arrangement of markers in Table~\ref{table:2x2matrix}, Figure~\ref{fig:MarkerOntology} outlines an organisation of the information markers with meaning, highlighting the recognition requirements for each category (agent and swarm focus with stationary and non-stationary types). Our ontology systematically organises markers to identify particular aspects of the agent and swarm, for instance, illuminating properties of an agent, the link of interaction for leadership dynamics, and the link between the agent and the swarm. As we move between information categories, the type of information required to calculate and recognise each category changes, as described in Table~\ref{table:2x2matrix}.

The marker ontology is designed from the perspective of a recognition agent, viewing the swarm and its agents. The top-level ontology classes \textit{swarm} and \textit{agent} are composed of sub-classes representing aspects in the system that are desired to be uncovered. The sub-classes that comprise an agent are traits and triggers. Traits are categorised as stationary (innate properties) or non-stationary (functional capability roles), with triggers representing individual decision thresholds based on traits. The sub-classes constituting a swarm are configuration (stationary properties) and tactic (non-stationary properties). For each of our agent and swarm stationary and non-stationary classes, we select a marker set to identify the desired aspects; the marker sets used for this study are as indicated in Table~\ref{table:2x2matrix}. This organisation now enables us to guide the development of analyses discussed in the following section.

    \begin{table*}[ht]
        \centering
        \resizebox{0.85\textwidth}{!}{%
        \begin{tabular}{@{}cccc@{}}
            \toprule
            \textbf{Parameter} & \textbf{High} & \textbf{Medium} & \textbf{Low} \\ \midrule
            $W_{\pi\Lambda}$            & 1.50                                & 1.05~\shortcite{Strombom:2014} & 0.50~\shortcite{Hepworth2020:Footprints} \\
            $W_{\pi\pi}$                & 3.00~\shortcite{Hepworth2020:Footprints} & 2.00~\shortcite{Strombom:2014} & 1.50~\shortcite{9504706} \\
            $W_{\pi\beta}$              & 1.90~\shortcite{9504706}                 & 1.00~\shortcite{Strombom:2014} & 0.50~\shortcite{Himo2022:HeterogeneousResponse} \\
            $\sfrac{s_\pi}{s_\beta}$     & 1.00                                & 0.67~\shortcite{Strombom:2014} & 0.50~\shortcite{9256255} \\
             \bottomrule
        \end{tabular}
        }
        \caption{We formulate $\pi_i$-agents as particle models with parameter variations identified in the literature, providing the basis for homogeneous and heterogeneous agent-profile implementations. Note that the agent decision models are homogeneous, with interaction weights ($W$) and agent speed ($s_\pi, s_\beta$) varied in our study.}
        \label{table:WeightRanges}
    \end{table*}

\section{Experimental Design}

Our experiments investigate how markers represent distinct situations by understanding marker sets' contribution to answering a given question. We are guided by the ontological relations given in Figure~\ref{fig:MarkerOntology}, using the sets defined in Table~\ref{table:2x2matrix}. Our experimental design is based on the particle-based shepherding model introduced by Str\" {o}mbom et al.,~\citeyear{Strombom:2014}. To develop heterogeneous contexts for swarm agents in this model, we parameterise the value of three weights ($W_{\pi\Lambda}$, $W_{\pi\pi}$, $W_{\pi\beta}$) and the speed differential between the swarm agent and control agent ($\sfrac{s_\pi}{s_\beta}$). The use of parameter variations to generate heterogeneous agent types is well established; see, for instance, Lee and Kim~\citeyear{s17122729} and Himo et al.,~\citeyear{Himo2022:HeterogeneousResponse}. These characteristic parameterisations are defined as
\begin{itemize}
    \item $W_{\pi\Lambda}$, the attraction strength for $\pi$ to their local centre of mass $\Lambda$.
    \item $W_{\pi\pi}$, the repulsion strength for a $\pi$ to another $\pi$.
    \item $W_{\pi\beta}$, being the repulsion strength for a $\pi$ to the control agent $\beta$.
    \item $\sfrac{s_\pi}{s_\beta}$, being speed differential between a $\pi$ agent and $\beta$.
\end{itemize}

\subsection{Agent Parametrisation and Swarm Heterogeneity}

We surveyed the available shepherding literature based on the model of Str\" {o}mbom et al., identifying variations from the original model for these weights. Varying specific agent parameters in shepherding models are well established, capturing distinct abilities and traits of agents~\shortcite{Abbass2020:UxV,9256255,Strombom:2014,Hussein:AAMAS22,9504706}. Table~\ref{table:WeightRanges} summarises our agent weight values selected across three levels (high, medium and low), with citations provided where weights are drawn directly from the literature. Where weights exist without citation to a particular study, we select a weight that ensures we have a magnitude-appropriate setting that remains faithful to the descriptions given by Str\" {o}mbom et al.~\citeyear{Strombom:2014}.

After developing our parameterisation levels, we reviewed the available biological shepherding literature to identify essential characteristics, abilities and traits able to be represented with our model. Our agent parameterisations are contained in Table~\ref{table:AgentCharacteristics}. Our first agent ($A1$) we describe linguistically as a \textit{scout}. The scout has a lower propensity to swarm, higher resistance to the swarm control agent's influence, and equal speed with the control agent. Our second agent ($A2$) we label as a \textit{control detractor}, who has a higher propensity to swarm, higher resistance to the swarm control agent~\shortcite{Himo2022:HeterogeneousResponse} and a lower relative speed than that of the swarm control agent. We characterise the third swarm agent ($A3$) as a \textit{swarm detractor}, who possesses a lower propensity to swarm and higher repulsion to other swarm agents. We describe the next agent ($A4$) as a \textit{nomad}, who has a lower propensity to swarm and higher repulsion to the swarm control agent. We label our fifth agent ($A5$) as a \textit{dispersed swarmer}, characterised by a higher repulsion to other swarm agents. The sixth agent ($A6$) we label as \textit{unwilling} is characterised by a lower repulsion to other swarm agents and lower relative speed than the swarm control agent. Our final agent ($A7$) is the classic agent introduced by Str\" {o}mbom et al.,~\citeyear{Strombom:2014}.

    \begin{table}[ht]
        \centering
        \resizebox{0.49\textwidth}{!}{%
        \begin{tabular}{clcccc}
            \toprule
            \textbf{Agent State} & \textbf{Name} & \textbf{$W_{\pi\Lambda}$} & \textbf{$W_{\pi\pi}$} & \textbf{$W_{\pi\beta}$} & \textbf{$\dfrac{s_\pi}{s_\beta}$} \\ \midrule
             \rule{0pt}{4ex}A1 & \rule{0pt}{4ex}Scout                 & \rule{0pt}{4ex}0.50 & \rule{0pt}{4ex}2.00 & \rule{0pt}{4ex}0.50 & \rule{0pt}{4ex}1.00 \\
             \rule{0pt}{4ex}A2 & \rule{0pt}{4ex}Control Detractor     & \rule{0pt}{4ex}1.50 & \rule{0pt}{4ex}2.00 & \rule{0pt}{4ex}0.50 & \rule{0pt}{4ex}0.50 \\
             \rule{0pt}{4ex}A3 & \rule{0pt}{4ex}Swarm Detractor       & \rule{0pt}{4ex}0.50 & \rule{0pt}{4ex}3.00 & \rule{0pt}{4ex}1.00 & \rule{0pt}{4ex}0.67 \\
             \rule{0pt}{4ex}A4 & \rule{0pt}{4ex}Nomad                 & \rule{0pt}{4ex}0.50 & \rule{0pt}{4ex}2.00 & \rule{0pt}{4ex}1.90 & \rule{0pt}{4ex}0.67 \\
             \rule{0pt}{4ex}A5 & \rule{0pt}{4ex}Dispersed (Protector) & \rule{0pt}{4ex}1.05 & \rule{0pt}{4ex}3.00 & \rule{0pt}{4ex}1.00 & \rule{0pt}{4ex}0.67 \\
             \rule{0pt}{4ex}A6 & \rule{0pt}{4ex}Unwilling             & \rule{0pt}{4ex}1.05 & \rule{0pt}{4ex}1.50 & \rule{0pt}{4ex}1.00 & \rule{0pt}{4ex}0.50 \\
             \rule{0pt}{4ex}A7 & \rule{0pt}{4ex}Classic               & \rule{0pt}{4ex}1.05 & \rule{0pt}{4ex}2.00 & \rule{0pt}{4ex}1.00 & \rule{0pt}{4ex}0.67 \\
             \bottomrule
        \end{tabular}
        }
        \caption{Summary of agent state vector parameterisations used for swarm agent parameterisations in this study. Seven $\pi_i$ profile types are presented, developed from available studies and empirical field trials.}
        \label{table:AgentCharacteristics}
    \end{table}

We reviewed biological shepherding literature after developing the agent parameterisations and available field experiment studies to design both homogeneous and heterogeneous swarms for marker experimentation. We developed a homogeneous swarm with each agent type and established four heterogeneous swarm distributions between two and four agent types per swarm. Table~\ref{table:SwarmScenarios} describes the distribution of each agent type within the 11 swarms developed.

    \begin{table*}[ht]
        \centering
        \resizebox{\textwidth}{!}{%
        \begin{tabular}{clccccccc}
            \toprule
            \textbf{Scenario State} & \textbf{Name} & \textbf{A1} & \textbf{A2} & \textbf{A3} & \textbf{A4} & \textbf{A5} & \textbf{A6} & \textbf{A7}  \\ \midrule
             $S1$  & Find and Guide~\shortcite{Hepworth2020:Footprints}    & 0.20 &      &      &      &      &      & 0.80 \\
             $S2$  & Disrupted~\shortcite{Yaxley2021:SS}                   &      & 0.20 & 0.20 &      &      & 0.20 & 0.40 \\
             $S3$  & Separated~\shortcite{Nowak2008SheepBehaviour}         &      &      &      & 0.80 &      &      & 0.20 \\
             $S4$  & Dispersed Search                                      & 0.20 &      &      &      & 0.20 &      & 0.60 \\
             $S5$  & Str\" {o}mbom et al.,~\citeyear{Strombom:2014}        &      &      &      &      &      &      & 1.00 \\
             $S6$  & Homogeneous A1                                        & 1.00 &      &      &      &      &      &      \\
             $S7$  & Homogeneous A2                                        &      & 1.00 &      &      &      &      &      \\
             $S8$  & Homogeneous A3                                        &      &      & 1.00 &      &      &      &      \\
             $S9$  & Homogeneous A4                                        &      &      &      & 1.00 &      &      &      \\
             $S10$ & Homogeneous A5                                        &      &      &      &      & 1.00 &      &      \\
             $S11$ & Homogeneous A6                                        &      &      &      &      &      & 1.00 &      \\
             \bottomrule
        \end{tabular}
        }
        \caption{Distributions of agent types (presented in Table~\ref{table:AgentCharacteristics}) constitute the swarm scenarios in this study, represented as proportions of each agent type. Scenarios $S5$ to $S11$ are homogeneous swarm types, with Scenarios $S1$ to $S4$ representative of natural-system heterogeneous swarm types.}
        \label{table:SwarmScenarios}
    \end{table*}

    \begin{figure*}[ht]
     \centering
     \begin{subfigure}{0.49\textwidth}
         \centering
         \includegraphics[width=\textwidth]{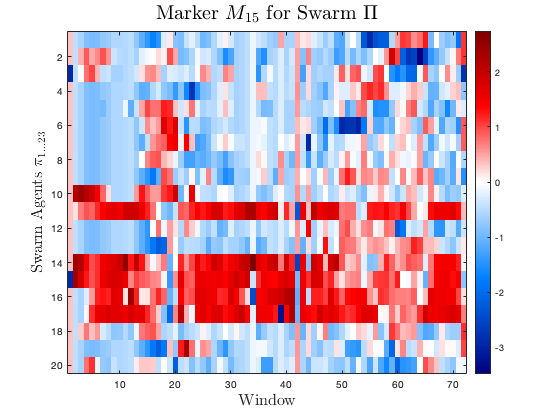}
         \caption{Mean of the Situation Awareness Marker ($M_{15}$) for each agent in $\Pi$ over the simulation observation period.}
         \label{fig:MarkerFootprints1}
     \end{subfigure}
     \hfill
     \begin{subfigure}{0.49\textwidth}
         \centering
         \includegraphics[width=\textwidth]{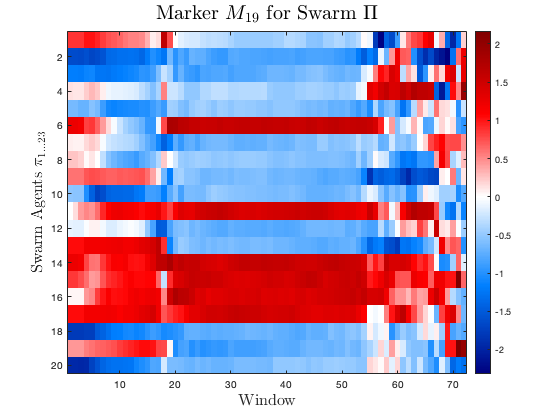}
         \caption{Maximum of the Distance Marker ($M_{19}$) for each agent in $\Pi$ over the simulation observation period.}
         \label{fig:MarkerFootprints2}
     \end{subfigure}
     \hfill
     \begin{subfigure}{0.49\textwidth}
         \centering
         \includegraphics[width=\textwidth]{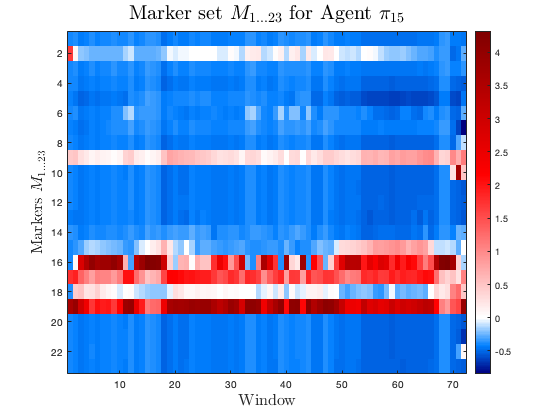}
         \caption{Marker subset $M_{1,\dots ,23}$ for agent $\pi_{15}$ over the simulation observation period, depicted as sequence of state vectors.}
         \label{fig:MarkerFootprints3}
     \end{subfigure}
     \hfill
     \begin{subfigure}{0.49\textwidth}
         \centering
         \includegraphics[width=\textwidth]{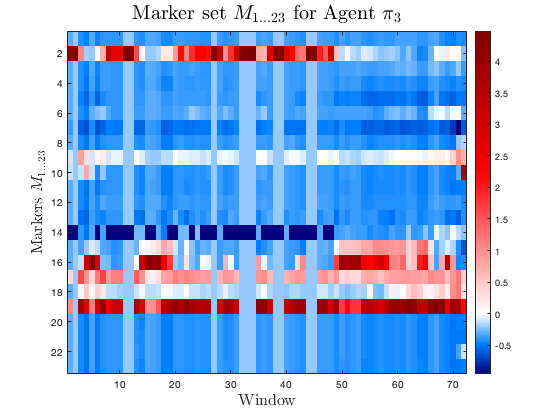}
         \caption{Marker subset $M_{1,\dots ,23}$ for agent $\pi_{3}$ over the simulation observation period, depicted as sequence of state vectors.}
         \label{fig:MarkerFootprints4}
     \end{subfigure}
        \caption{Visualisation of scenario $S_1$ (Table~\ref{table:MarkerLiteratures}), depicting a single marker output across all the agents in the swarm (sub-figure $a$ and $b$), as well as a subset of markers ($\mathcal{M}_{1,\dots ,23}$) for a single agent type in the swarm (sub-figure $c$ and $d$). Note that figures are depicted as normalised values, given as a vector-wise z-score for each window (column), with mean $0$ and standard deviation $1$.}
        \label{fig:MarkerFootprints}
\end{figure*}

\subsection{Generating Simulation Data}

In our simulation environment, we implement the 11 swarm scenarios in Table~\ref{table:SwarmScenarios}. These were implemented as described in Str\" {o}mbom et al.,~\citeyear{Strombom:2014}, with agent parameterisations and swarm agent distributions described per the previous section. We used a swarm size of $N=20$ agents for our simulations with a single $\beta$. Information markers were calculated over marker windows consisting of 20 to 100 observations, with 25-75\% overlap between each window~\shortcite{KLEANTHOUS2022442}. This resulted in 165 variations across the 11 scenarios. Data were shuffled randomly to break time series associations and split into a training set (80\%) and a test set (20\%). We verified consistency in the representation of each agent type across both data sets.

Figure~\ref{fig:MarkerFootprints} depicts exemplar outputs for a single scenario ($S_1$), providing two perspectives. Figure~\ref{fig:MarkerFootprints1} and Figure~\ref{fig:MarkerFootprints2} show the state value of $\mathcal{M}_{15}$ and $\mathcal{M}_{19}$ for all agents, $\Pi$ in scenario $S_1$. With these sub-figures, we conduct the Kruskal-Wallis to demonstrate the individual power of a marker to highlight particular aspects of each swarm agent. In Figure~\ref{fig:MarkerFootprints1} we reject the null hypothesis that each agent has the same profile distribution for $M_{15}$ ($H(19)=501.07, p<0.001$); for Figure~\ref{fig:MarkerFootprints2} we also reject the null hypothesis that each agent has the same profile distribution for $M_{19}$ ($H(19)=781.92, p<0.001$). Expanding this analysis to the 23-marker sub-group, only two markers failed to reject the null hypothesis at the $p=0.05$ level, with the remaining 21 markers rejecting it. 

We observe that markers discriminate a particular aspect of each agent in the swarm, such as $\pi_{14\dots 18}$ having disparate state values to other members in the swarm. Common to both figures are key system changes in the simulation. At $t\approx 18$, we observe a change in marker states, aligning to a control state change where $\beta$ is actively shepherding $\Pi$. At $t\approx 55$, we observe another marker state change, corresponding to the simulation stage where $\beta$ is at the final phase of simulated shepherding. Figure~\ref{fig:MarkerFootprints3} and Figure~\ref{fig:MarkerFootprints4} depict a complete marker state (as described in Table~\ref{table:MarkerLiteratures}) consisting of 23 markers for each agent in observation window (per Table~\ref{table:2x2matrix}). These figures show the individual \emph{footprints} of $\pi_{15}$ and $\pi_3$ through the scenario, quantifying the unique state of an agent over time. We also compare the marker state values for each agent, using the Kruskal-Wallis to test if markers provide redundant information on an agent. In Figure~\ref{fig:MarkerFootprints3} we reject the null hypothesis that each marker has the same distribution for $\pi_{15}$ ($H(22)=1121.16, p<0.001$); for Figure~\ref{fig:MarkerFootprints4} we also reject the null hypothesis that each marker has the same distribution for $\pi_{3}$ ($H(22)=973.89, p<0.001$). Again, we expand this analysis to all 20 agents of the swarm; we reject the null hypothesis for all agent cases at the $p=0.05$ level. 

    \begin{table}[ht]
        \centering
        \begin{tabular}[width=0.475\textwidth]{@{}lll@{}}
            \toprule
            \textbf{Marker}     & \textbf{Method, Technique or Measure} & \textbf{Variation}        \\ \midrule
             M1                 & Speed                                 & Segment                   \\
             M2                 & Distance                              & Segment Rate              \\
             M3                 & Speed                                 & Mean                      \\
             M4                 & Speed                                 & Var                       \\
             M5                 & Heading                               & Mean                      \\
             M6                 & Heading                               & Var                       \\
             M7                 & Situation Awareness                   & Mean                      \\
             M8                 & Situation Awareness                   & Var                       \\
             M9                 & Predation Risk                        & Mean                      \\
             M10                & Predation Risk                        & Var                       \\
             M11                & Dynamic Body Acceleration             & Mean                      \\
             M12                & Dynamic Body Acceleration             & Var                       \\
             M13                & Dynamic Body Acceleration             & Cumulative                \\
             M14                & Rate Of Change (Angular)              & Velocity                  \\
             M15                & Cross Correlation                     & Mean                      \\
             M16                & Cross Correlation                     & Var                       \\
             M17                & Distance                              & Mean                      \\
             M18                & Distance                              & Var                        \\
             M19                & Distance                              & Max                        \\
             M20                & Distance                              & Min                       \\
             M21                & Synchronicity                         & Mean                      \\
             M22                & Synchronicity                         & Var                       \\
             M23                & Transfer Entropy                      & Net                       \\
             M24                & Dynamic Time Warping                  & Mean                      \\
             M25                & Dynamic Time Warping                  & Var                       \\
             M26                & Active Information Storage            & Mean                      \\
             M27                & Transfer Entropy                      & Total                     \\
             M28                & Effort to Compress                    & $<$value$>$               \\
             M29                & Transfer Entropy                      & Internal Net              \\
             M30                & Transfer Entropy                      & External Net              \\
             M31                & Transfer Entropy                      & Agg. Infl.                \\
             M32                & Transfer Entropy                      & Net Source                \\
             M33                & Information Flow                      & Mean                      \\
             M34                & Information Flow                      & Var                       \\
             M35                & Information Flow                      & Mean                      \\
             M36                & Information Flow                      & Var                       \\
             M37                & Lyapunov Exponent                     & Mean                      \\
             M38                & Lyapunov Exponent                     & Var                       \\
             M39                & Noise-to-Signal                       & Mean                      \\
             M40                & Noise-to-Signal                       & Var                       \\
             M41                & Power Spectral Density                & Entropy                   \\
             M42                & Shannon Entropy                       & $<$value$>$               \\
             \bottomrule
        \end{tabular}
        \caption{Summary of the 42 markers selected for use in this study to classify agent types. Each marker is selected to generate a distinct perspective on the swarm agents, designed with high discriminatory power. Markers selected are derived from MTMs in Table~\ref{table:AnalysisLiterature} and organised into information marker states per Table~\ref{table:2x2matrix}.}
        \label{table:MarkerLiteratures}
    \end{table}

\section{Analysis}

\begin{figure*}[ht]
    \centering
    \includegraphics[width=\textwidth]{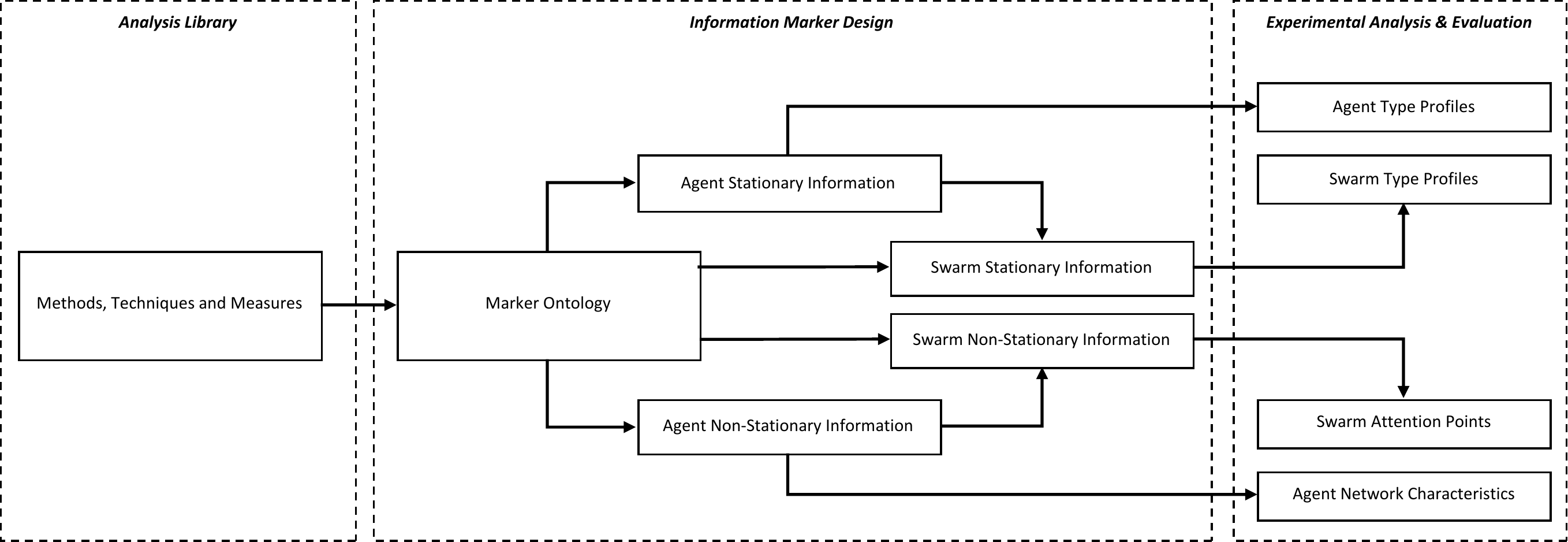}
    \caption{Conceptual methodological process for experimental design, depicting the information marker organisation and state flow, from the collection of methods, techniques and measures in Table~\ref{table:AnalysisLiterature}, conceptual organisation of the ontology in Figure~\ref{fig:MarkerOntology}, and analysis detailed in the following sections.}
    \label{fig:MethodProcess}
\end{figure*}

Our analysis aims to demonstrate the application of the information marker method through examples. The first is to illustrate the marker overlap by investigating the pairwise correlation between all markers, which seeks to quantify the information similarity between markers and identify existing groupings. The second analysis reports key findings from the marker set in Table~\ref{table:2x2matrix}. These focus on swarm and agent profiles (classification). Our final analysis aims to detect a change in an agent's interaction role as the swarm's tactic. The analyses are ontologically guided to answer particular questions about what is desired to be understood on the swarm. There are endless possibilities for analysis to inform on new aspects of the swarm, limited only by the imagination of an analyst and desired aspects sought to be understood. Figure~\ref{fig:MethodProcess} outlines the conceptual flow from data to information to markers and the analysis types presented in this section.

\subsection{Exploration of Marker State Vector}

Our initial task is identifying the information dependencies between information markers and their contribution to classifications. We include 23 of the 42 information markers from nine methods and measures in Table~\ref{table:AnalysisLiterature}, ensuring coverage across the three focal lenses identified (per Table~\ref{table:2x2matrix}). Table~\ref{table:MarkerLiteratures} contains our summary of markers for this study. Next, we formulate the learning of distinct agent types from markers as a classification problem. This task is consistent with the depiction given in Figure~\ref{fig:sysDefMap}, illuminating our intent to uncover the behaviour of swarm agents through the detection of context.

We conducted a high-level assessment of five different classification models, selecting the decision tree as our classification model for its established use in recognition tasks~\shortcite{HAR2022:CPS}, interpretability of its model output, limited input data preparation, fast training and prediction costs. Model hyperparameters were optimised with a Bayesian scheme, with 10-fold cross-validation used throughout the training. Model training resulted in a classifier accuracy (validation) of 81.5\% for 23 markers (M1-M23). Our analysis focus here is to understand the contribution of each marker to the overall classification of each agent and swarm state, with markers used for this experimental recognition task. The evaluations were conducted on the marker inputs to determine the classification impact of missing markers. Our motivation is to understand the level of confidence in settings of swarm control for recognitions provided to a swarm control agent.

Our regime to explore the marker state vectors consists of two evaluation stages. The first stage (E1, Table~\ref{table:AgentProfileE1}) is a feature ablation to study the system performance by varying different features on the dataset~\shortcite{Sheikholeslami1349978}. We retained our model with a leave-one-out policy, systematically removing each marker to assess the impact of that marker on the overall classification. We then computed the Mutual Information (MI) between markers to measure information uniqueness. Our goal here is to select the minimum set of markers that provided $>95$\% of the cumulative MI uniqueness from the complete set ($\mathcal{M}_{MI(95)}$), where $\mathcal{M}_{MI(95)} = \{M1, M3, M4, M5, M6, M11, M12, M13, M14, M15, \\M16, M22, M23\}$ is the minimum marker set containing $>95\%$ of the mutual information variance. We then removed the \textit{Centre of Influence}~\shortcite{Hepworth2020:Footprints}, where $\mathcal{M}_{\text{COI}} = \{M7, M8, M9, M10, M21, M22\}$. Our results for this stage of the ablation study are contained in Table~\ref{table:AgentProfileE1}, which indicate that the ablation of an individual marker has a predominantly negligible impact on the performance of a pre-trained classifier for swarm agent behaviour. The notable exception to this is the ten markers not included within the $\mathcal{M}_{MI(95)}$ group, resulting in a substantial decrease in accuracy.

    \begin{table}[ht]
    \centering
        \begin{tabular}[width=0.475\textwidth]{@{}lcc@{}}
            \toprule
            \textbf{Marker Set} & \textbf{Accuracy} & \textbf{\% Change} \\ \midrule
            $\mathcal{M}$                            & 83.0 & \textemdash \\
            \midrule
            $\mathcal{M}$~\textendash~\{M1\}         & 80.1 & -3.5  \\
            $\mathcal{M}$~\textendash~\{M2\}         & 78.6 & -5.3  \\
            $\mathcal{M}$~\textendash~\{M3\}         & 79.6 & -4.1  \\
            $\mathcal{M}$~\textendash~\{M4\}         & 81.2 & -2.2  \\
            $\mathcal{M}$~\textendash~\{M5\}         & 83.2 & +0.2  \\
            $\mathcal{M}$~\textendash~\{M6\}         & 81.1 & -2.9  \\
            $\mathcal{M}$~\textendash~\{M7\}         & 83.0 &  0.0  \\
            $\mathcal{M}$~\textendash~\{M8\}         & 82.9 & -0.1  \\
            $\mathcal{M}$~\textendash~\{M9\}         & 81.1 & -2.9  \\
            $\mathcal{M}$~\textendash~\{M10\}        & 82.9 & -0.1  \\
            $\mathcal{M}$~\textendash~\{M11\}        & 83.0 &  0.0  \\
            $\mathcal{M}$~\textendash~\{M12\}        & 81.0 & -2.4  \\
            $\mathcal{M}$~\textendash~\{M13\}        & 83.0 &  0.0  \\
            $\mathcal{M}$~\textendash~\{M14\}        & 83.7 & +0.8  \\
            $\mathcal{M}$~\textendash~\{M15\}        & 83.2 & +0.2  \\
            $\mathcal{M}$~\textendash~\{M16\}        & 82.9 & -0.1  \\
            $\mathcal{M}$~\textendash~\{M17\}        & 80.3 & -3.6  \\
            $\mathcal{M}$~\textendash~\{M18\}        & 78.0 & -6.2  \\
            $\mathcal{M}$~\textendash~\{M19\}        & 80.4 & -3.1  \\
            $\mathcal{M}$~\textendash~\{M20\}        & 83.0 &  0.0  \\
            $\mathcal{M}$~\textendash~\{M21\}        & 83.1 & +0.1  \\
            $\mathcal{M}$~\textendash~\{M22\}        & 82.8 & -0.2  \\
            $\mathcal{M}$~\textendash~\{M23\}        & 83.0 &  0.0  \\
            $\mathcal{M}$~\textendash~\{MI 5\%\}     & 54.1 & -34.8 \\
            $\mathcal{M}$~\textendash~\{COI\}        & 81.2 & -2.2  \\
            \bottomrule
        \end{tabular}
        \caption{Detailed analysis results continued from the analysis section, evaluating the discriminatory power of markers for detecting different different profiles (classification) for $M_{1,\dots ,23}$, as given in Table~\ref{table:AnalysisLiterature} and Table~\ref{table:MarkerLiteratures}. This table reports feature ablation results for the systematic removal of individual features, as well as two designed feature groups ($\mathcal{M}_{MI}$ and $\mathcal{M}_{COI}$). Each row indicates the training of a new classifier based on the identified marker set, with results representing the overall accuracy and change in accuracy compared to the complete marker set ($\mathcal{M}=23$). This analysis highlights information inter-dependency between markers, contributing to the overall classification performance.}
        \label{table:AgentProfileE1}
    \end{table}

Our second evaluation stage (E2, Table~\ref{table:AgentProfileE2}) replicates the methodology process of stage one; however, employing only the classification model trained on the subset of 23 markers, with all markers present during the classifications. We conducted the same systematic changes per the policies outlined for stage one. We modify this through the systematic transformation of each marker input. To achieve this, every observation of the relevant marker(s) under evaluation was set as the mean of that marker~\shortcite{Emmanuel2021:MissingData}. Our purpose for this evaluation is to investigate impacts on a marker when dealing with changes in the underlying data. This is important during swarm control for tactic and strategy selection, where sensor inputs may impact the control agent decisions. This is particularly important for settings where data acquisition cannot be guaranteed and imputation must occur dynamically. Our evaluation uses the F1 score, observing an increased variance in classifier performance. E2 highlights the potential for non-linear interactions between markers, mainly observed for both the $\mathcal{M}_{MI(95)}$ and Centre of Influence marker groups, as well as for some individual markers (see, for example, M18).

    \begin{table}[ht]
    \centering
        \begin{tabular}[width=0.475\textwidth]{@{}lcc@{}}
            \toprule
            \textbf{Marker Set} & \textbf{F1 Score} & \textbf{\% Change} \\ \midrule
            $\mathcal{M}$                            & 81.0 & \textemdash \\
            \midrule
            $\mathcal{M}$~\textendash~\{M1\}         & 70.3 & -13.2 \\
            $\mathcal{M}$~\textendash~\{M2\}         & 66.5 & -17.9 \\
            $\mathcal{M}$~\textendash~\{M3\}         & 69.6 & -14.1 \\
            $\mathcal{M}$~\textendash~\{M4\}         & 71.3 & -12.0 \\
            $\mathcal{M}$~\textendash~\{M5\}         & 77.8 &  -3.9 \\
            $\mathcal{M}$~\textendash~\{M6\}         & 72.7 & -10.3 \\
            $\mathcal{M}$~\textendash~\{M7\}         & 75.2 &  -7.2 \\
            $\mathcal{M}$~\textendash~\{M8\}         & 79.9 &  -1.4 \\
            $\mathcal{M}$~\textendash~\{M9\}         & 72.6 & -10.4 \\
            $\mathcal{M}$~\textendash~\{M10\}        & 78.5 &  -3.1 \\
            $\mathcal{M}$~\textendash~\{M11\}        & 80.7 &  -0.4 \\
            $\mathcal{M}$~\textendash~\{M12\}        & 79.9 &  -1.4 \\
            $\mathcal{M}$~\textendash~\{M13\}        & 81.0 &   0.0 \\
            $\mathcal{M}$~\textendash~\{M14\}        & 79.7 &  -1.6 \\
            $\mathcal{M}$~\textendash~\{M15\}        & 79.1 &  -2.4 \\
            $\mathcal{M}$~\textendash~\{M16\}        & 79.4 &  -2.0 \\
            $\mathcal{M}$~\textendash~\{M17\}        & 55.8 & -31.3 \\
            $\mathcal{M}$~\textendash~\{M18\}        & 47.7 & -41.1 \\
            $\mathcal{M}$~\textendash~\{M19\}        & 54.3 & -33.0 \\
            $\mathcal{M}$~\textendash~\{M20\}        & 81.0 &   0.0 \\
            $\mathcal{M}$~\textendash~\{M21\}        & 81.1 &  +0.1 \\
            $\mathcal{M}$~\textendash~\{M22\}        & 80.1 &  +0.1 \\
            $\mathcal{M}$~\textendash~\{M23\}        & 81.0 &   0.0 \\
            $\mathcal{M}$~\textendash~\{MI 5\%\}     & 29.5 & -63.6 \\
            $\mathcal{M}$~\textendash~\{COI\}        & 64.8 & -20.0 \\
            \bottomrule
        \end{tabular}
        \caption{This table reports the impact of marker transformation evaluations, representing modulated inputs. At each row, the identified marker or sub-set of markers was set to their mean value across all observations. This analysis assesses the $F1$-score impact of a transformed marker for the setting where a classifier is trained with all markers present ($M_{1,\dots ,23}$). We note that classic feature selection methods, such as excluding the bottom-$k$\% of features based on Mutual Information, significantly impact classifier performance, with the bottom $5$\% of features resulting in a 63.6\% classification performance decline.}
        \label{table:AgentProfileE2}
    \end{table}

\subsection{Agent and Swarm Profiles}

For our agent and swarm profile classifications, we use the same underlying methodology discussed in the previous section; however, we now include the full set of 42 markers identified in Table~\ref{table:MarkerLiteratures}. The decision tree model classifier is optimised with a Bayesian scheme, with 10-fold Cross Validation and a train/test data split of 80/20 applied across all generated data. This analysis aims to classify agent and swarm types based on the marker state vector generated and understand the impact of observation window size and observation window overlap on classifier performance. Observation window variation is established to improve accuracy, latency, and the associated cost of processing~\shortcite{JaenVargas2022:WindowSizeAR}. We initially train a classifier to discover agent profile types (Table~\ref{table:AgentCharacteristics}) for each observation window size $\{20, 40, 60 ,80, 100\}$ and observation window overlap, $\{0.25, 0.50, 0.75\}$. Our results are in Table~\ref{table:AgentTestAccuracy}, reporting validation test model accuracy across 7 classes ($A_{1,\dots ,7}$). We observe maximum classifier performance for detecting agent profiles for a window size of 20 and window overlap of 0.75 (75\%), $[20, 0.75]$.

    \begin{table}[ht]
            \centering  
            \begin{tabular}[width=0.475\textwidth]{@{}ccccccc@{}}
                \toprule
                \multicolumn{1}{l}{} & \textbf{Type} & \textbf{20} & \textbf{40} & \textbf{60} & \textbf{80} & \textbf{100} \\ \midrule
                \multirow{2}{*}{\textbf{75\%}} & V & 87.9  & 83.8 & 85.5 & 85.7 & 85.5 \\
                                               & T & 88.7  & 84.4 & 87.2 & 86.7 & 86.0 \\
                \multirow{2}{*}{\textbf{50\%}} & V & 83.4  & 83.1 & 83.6 & 84.4 & 86.4 \\
                                               & T & 83.0  & 85.9 & 83.8 & 86.5 & 86.9 \\
                \multirow{2}{*}{\textbf{25\%}} & V & 79.4  & 80.2 & 82.1 & 79.5 & 80.9 \\
                                               & T & 79.7  & 80.0 & 83.1 & 79.5 & 80.2 \\
                \bottomrule
            \end{tabular}
        \caption{Summary agent profile classification performance results, reported as validation and test model accuracy for 7 classes ($\pi_1,\dots,\pi_{7}$). Our selected model is a decision tree classifier optimised with a Bayesian scheme (opt. max $n=30$).Columns with numbered headers report the window observation size; rows with percentages report the window overlap. Type refers to validation or test data results. No feature pre-processing of marker state vectors is conducted; each classifier uses $\mathcal{M}=42$ marker state vectors for the classification task. 10-fold Cross Validation with a train/test split of 80/20 is applied for all model data, with temporal dependencies between data broken with observation permutation prior to training. This table reports the classifier's performance for 15 distinct hyper-parameter settings across all scenarios (Table~\ref{table:SwarmScenarios}). The hyper-parameters varied are window size (number of observations per marker state vector calculation) and sliding window overlap (percentage), with maximum classifier performance achieved with window size $w=20$ and window overlap $o=75\%$.}
        \label{table:AgentTestAccuracy}
    \end{table}

\begin{figure}[ht]
        \centering
        \includegraphics[width=0.5\textwidth]{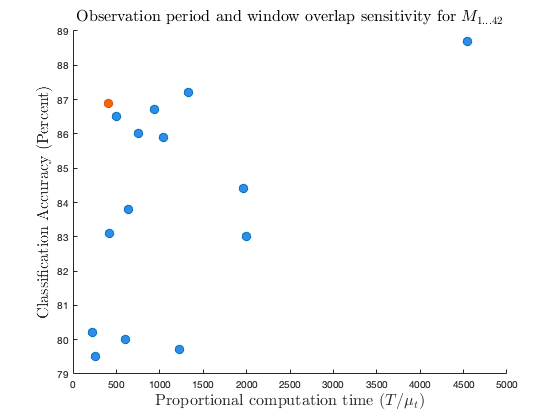}
        \caption{Depiction of agent profile classifications contained in Table~\ref{table:AgentTestAccuracy} and Table~\ref{table:AgentComputeTime} data, depicting the trade-off between classification accuracy and compute time. Classification accuracy is given from test data (20\% withheld). Proportional computation time is calculated as the average total marker computation time per simulation scenario, divided by the average mean marker computation time for a single observation period. As classification accuracy increases, we observe an increase in the total number of computations conducted per scenario, characterised by decreased observation periods and increased total computation time. A window size gives the best-identified trade-off between classification accuracy and proportional computation time $100$ and window overlap $0.5~(50\%)$, identified by the orange marker (accuracy $= 86.9$\%, mean computation time $= 2.85$ seconds).}
        \label{fig:AgentClassSensitivity}
    \end{figure}

We evaluate classifier performance in recognition settings by considering the computational cost, reported as mean compute-time ($\mu_t$) and total compute time ($T$) for each observation window size and observation window overlap (Table~\ref{table:AgentComputeTime}). In this setting, the hyperparameters that maximise classification accuracy, $[20, 0.75]$, are associated with the highest computational cost, with marginal classification performance increases over less computationally intensive hyperparameter variations. Through this lens, it is prudent to determine the optimally cost-efficient hyperparameters, particularly for online implementation settings. Figure~\ref{fig:AgentClassSensitivity} depicts the coupling observed between classification accuracy and the proportional computation time, defined as the average total marker computation time per simulation scenario, divided by the average mean marker computation time for a single observation period, $\sfrac{T}{\mu_t}$. The best-identified trade-off between classification accuracy and proportional computation time for agent type profile classification is given by a window size $100$ and window overlap $0.5~(50\%)$, given as $[100, 0.5]$, identified by the orange marker (accuracy $= 86.9$\%, mean computation time per window $= 2.85$ seconds). We compare the mean compute time for each setting described in Table~\ref{table:AgentComputeTime} and reject the null hypothesis that compute times are equivalent ($F_{14,16832}=13888.19, p<0.001$). 

    \begin{table}[ht]
        \centering
        \resizebox{0.49\textwidth}{!}{%
        \begin{tabular}[width=0.475\textwidth]{ccccccc}
            \toprule
            \multicolumn{1}{l}{} & \textbf{Type} & \textbf{20} & \textbf{40} & \textbf{60} & \textbf{80} & \textbf{100} \\ \midrule
            \multirow{2}{*}{\textbf{75\%}} & $\mu_t$  & 1.1    & 1.6    & 2.0    & 2.5    & 2.9    \\
                                           & $T$      & 444.5  & 281.7  & 240.8  & 215.4  & 199.2  \\
            \multirow{2}{*}{\textbf{50\%}} & $\mu_t$  & 1.1    & 1.6    & 2.0    & 2.4    & 2.9    \\
                                           & $T$      & 193.7  & 147.4  & 114.3  & 115.6  & 105.5  \\
            \multirow{2}{*}{\textbf{25\%}} & $\mu_t$  & 1.1    & 1.6    & 2.0    & 2.6    & 2.9    \\
                                           & $T$      & 120.1  & 87.2   & 77.7   & 61.0   & 59.5   \\
            \bottomrule
        \end{tabular}
        }
        \caption{Summary of classification compute-time data (reported in seconds), for the hyper-parameter settings described in Table~\ref{table:AgentTestAccuracy}. Columns with numbered headers report the window observation size; rows with percentages report the window overlap. Type refers to mean time per calculation or mean cumulative computation time per simulation. The mean represents the average compute time for a marker state vector under the specified conditions (window observation size and overlap percentage) across all scenarios, with the total time representing the average cumulative marker computation time across all scenarios. We observe that the optimal hyper-parameter settings ($w=20, o=75$) from Table~\ref{table:AgentTestAccuracy} are associated with the greatest computational cost, with only marginal performance increases above other results with substantially lower computational costs. Figure~\ref{fig:AgentClassSensitivity} and Figure~\ref{fig:SwarmSensitivity} highlight the trade-off between information gain and computational cost.}
        \label{table:AgentComputeTime}
    \end{table}

    \begin{figure*}[ht]
        \begin{subfigure}{0.5\textwidth}
            \centering
            \includegraphics[width=\textwidth]{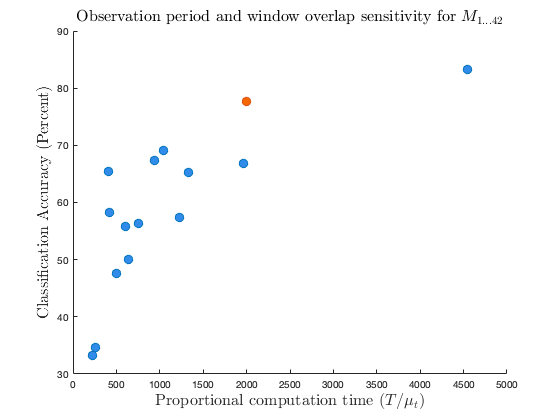}
            \caption{Summary of Table~\ref{table:SwarmTestAccuracy11class} and Table~\ref{table:AgentComputeTime} data, depicting the trade-off between classification accuracy and compute time (11-class). The best-identified point is given by window size $20$, window overlap $0.75~(75\%)$ and mean computation time $1.07$ seconds.}
            \label{fig:Swarm11ClassSensitivity}
        \end{subfigure}
        \hfill
        \begin{subfigure}{0.5\textwidth}
            \centering
            \includegraphics[width=\textwidth]{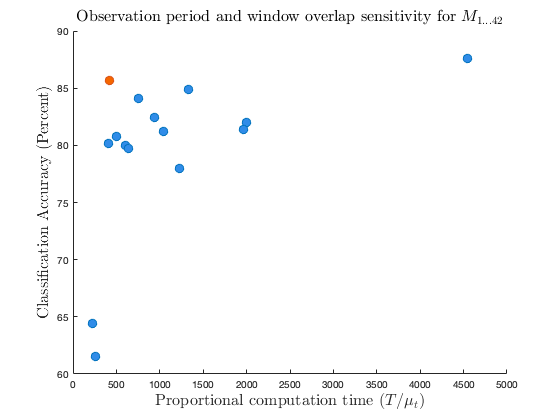}
            \caption{Summary of Table~\ref{table:SwarmTestAccuracy2class} and Table~\ref{table:AgentComputeTime} data, depicting the trade-off between classification accuracy and compute time (2-class). The best-identified point is given by window size $60$, window overlap $0.75~(75\%)$ and mean computation time $2.03$ seconds.}
            \label{fig:Swarm2ClassSensitivity}
        \end{subfigure}
        \caption{Classification accuracy is given from test data (20\% withheld). Proportional computation time is as described in Figure~\ref{fig:AgentClassSensitivity}. As classification accuracy increases, we observe an increase in the total number of computations conducted per scenario, characterised by decreased observation periods and increased total computation time. The orange marker identifies the best-identified trade-off between classification accuracy and proportional computation time in each sub-figure. We observe a non-linear increase in computation time for a given classification accuracy, with the notable outlier being for window size $20$ and window overlap $0.75~(75\%)$.}
        \label{fig:SwarmSensitivity}
    \end{figure*}

At the swarm level, our objective is to classify the swarm profile in two ways. The first is to identify the scenario as observed  (11 classes, $S_{1,\dots, 11}$, contained in Table~\ref{table:SwarmTestAccuracy11class}) and the second is to identify if the swarm is homogeneous or heterogeneous (2 classes, contained in Table~\ref{table:SwarmTestAccuracy2class}). We observe optimal classifier performance for detecting 11-class swarm profiles with a window size of 20 and window overlap of 0.75 (75\%), $[20, 0.75]$, depicted in Figure~\ref{fig:Swarm11ClassSensitivity}. In contrast, we observe optimal classifier performance for the 2-class swarm profile with a window size of 60 and overlap of 0.25 (25\%), $[60, 0.75]$, depicted in Figure~\ref{fig:Swarm2ClassSensitivity}. As with our agent classification, we seek to find the optimal trade-off between classification performance and computational efficiency. For the 11-class setting, the hyperparameters $[20, 0.5]$ are identified as optimal, whereas $[60, 0.25]$ is identified as optimal for the 2-class setting. The figures contained at Figure~\ref{fig:AgentClassSensitivity} depict relationships akin to those observed for the agent type profile classifications in Figure~\ref{fig:AgentClassSensitivity}, with an observed super-linear increase in computation time for a given classification.

\begin{table}[ht]
        \centering
        \begin{tabular}[width=0.475\textwidth]{@{}ccccccc@{}}
            \toprule
            \multicolumn{1}{l}{} & \textbf{Type} & \textbf{20} & \textbf{40} & \textbf{60} & \textbf{80} & \textbf{100} \\ \midrule
            \multirow{2}{*}{\textbf{75\%}} & V & 81.7 & 66.5 & 68.2 & 60.1 & 59.0 \\
                                           & T & 83.3 & 66.8 & 65.3 & 67.4 & 56.3 \\
            \multirow{2}{*}{\textbf{50\%}} & V & 73.0 & 67.3 & 60.0 & 53.9 & 55.2 \\
                                           & T & 77.7 & 69.1 & 50.0 & 47.5 & 65.4 \\
            \multirow{2}{*}{\textbf{25\%}} & V & 59.3 & 51.7 & 49.9 & 48.6 & 43.9 \\
                                           & T & 57.3 & 55.8 & 58.3 & 34.6 & 33.3 \\
            \bottomrule
        \end{tabular}
        \caption{Summary of swarm profile classification results, reported as validation and test model accuracy (11 classes, $S_1,\dots,S_{11}$, defined in Table~\ref{table:SwarmScenarios}). Columns with numbered headers report the window observation size; rows with percentages report the window overlap. Type refers to validation or test data results. Model selection, optimisation and training are as described in Table~\ref{table:AgentTestAccuracy}, with results indicating a reduction in accuracy over agent-type classifications. Additional simulation data were generated to understand the impact of sample size on accuracy across the 11 classes; however, there is a lack of evidence to suggest that additional samples are statistically significant in increasing classifier performance. The hyper-parameters varied are window size (number of observations per marker state vector calculation) and sliding window overlap (percentage), with maximum classifier performance achieved with window size $w=20$ and window overlap $o=75\%$.}
        \label{table:SwarmTestAccuracy11class}
\end{table}

\begin{table}[ht]
        \centering
        \begin{tabular}[width=0.475\textwidth]{@{}ccccccc@{}}
            \toprule
            \multicolumn{1}{l}{} & \textbf{Type} & \textbf{20} & \textbf{40} & \textbf{60} & \textbf{80} & \textbf{100} \\ \midrule
            \multirow{2}{*}{\textbf{75\%}} & V & 88.3 & 82.6 & 85.3 & 77.7 & 79.7 \\
                                           & T & 87.6 & 81.4 & 84.9 & 82.4 & 84.1 \\
            \multirow{2}{*}{\textbf{50\%}} & V & 81.7 & 81.7 & 81.7 & 75.1 & 82.2 \\
                                           & T & 82.0 & 81.2 & 79.7 & 80.8 & 80.2 \\
            \multirow{2}{*}{\textbf{25\%}} & V & 81.5 & 77.1 & 75.7 & 82.1 & 65.0 \\
                                           & T & 78.0 & 80.0 & 85.7 & 61.5 & 64.4 \\
            \bottomrule
        \end{tabular}
        \caption{Summary of swarm profile classification results, reported as validation and test model accuracy for 2 classes, being \emph{homogeneous} ($S_{5,\dots ,11}$) or \emph{heterogeneous} ($S_{1,\dots ,4}$), based on Table~\ref{table:SwarmScenarios}. Columns with numbered headers report the window observation size; rows with percentages report the window overlap. Type refers to validation or test data results. Model selection, optimisation and training are as described in Table~\ref{table:AgentTestAccuracy}, with results indicating a reduction in accuracy over agent-type classifications. We observe an increased classification performance over those reported in Table~\ref{table:SwarmTestAccuracy11class}. The hyper-parameters varied are window size (number of observations per marker state vector calculation) and sliding window overlap (percentage), with maximum classifier performance achieved with window size $w=20$ and window overlap $o=75\%$.}
        \label{table:SwarmTestAccuracy2class}
\end{table}

\begin{figure*}[ht]
        \begin{subfigure}{0.5\textwidth}
            \centering
            \includegraphics[width=\textwidth]{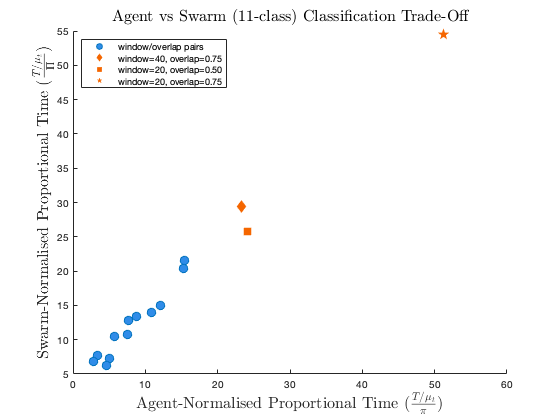}
            \caption{Comparison of agent and swarm classification accuracy for the 11-class swarm setting (swarm type classification target).}
            \label{fig:AgentSwarm11Class}
        \end{subfigure}
        \hfill
        \begin{subfigure}{0.5\textwidth}
            \centering
            \includegraphics[width=\textwidth]{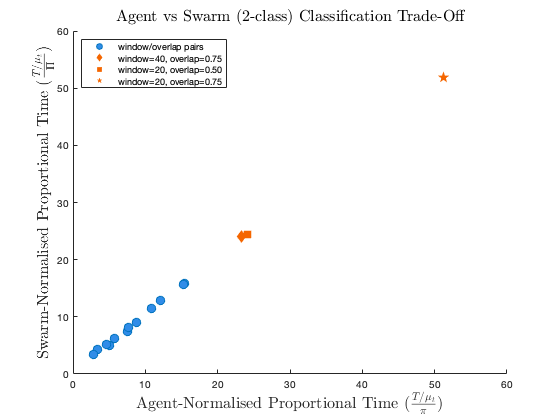}
            \caption{Comparison of agent and swarm classification accuracy for the 2-class swarm setting (swarm type as either homogeneous or heterogeneous).}
            \label{fig:AgentSwarm2Class}
        \end{subfigure}
        \caption{Visualisation of Table~\ref{table:AgentTestAccuracy}, Table~\ref{table:AgentComputeTime}, Table~\ref{table:SwarmTestAccuracy11class} and Table~\ref{table:SwarmTestAccuracy2class}, depicting highly linear relationships between agent and swarm classification accuracy and computation time. This figure aims to identify an optimal marker window size and window overlap percentage for the computation of both agents and swarm markers. Four categories of data are highlighted in this figure. The first is the blue markers that represent non-optimal window/over pairs. The second (orange diamond) and third (orange square) categories represent identified feasible data that feature increased classification accuracy over the first category while not substantially increasing compute time. The fourth category (orange star) represents the optimal classification setting for both agent and swarm settings; however, we see disproportionately higher compute times in both sub-figures.}
        \label{fig:AgentSwarmSensitivity}
    \end{figure*}

In real-world applications, selecting only a single hyperparameter pair (window size and overlap) for both agent and swarm-level classifications is practical, minimising the required computational cost. In Figure~\ref{fig:AgentSwarmSensitivity}, we depict the relationship between agent-time normalised swarm-time normalised classification for both the 11-class (Figure~\ref{fig:AgentSwarm11Class}) and 2-class (Figure~\ref{fig:AgentSwarm2Class}) swarm settings. We observe a linear relationship between each dataset, indicating the feasibility of selecting a single hyperparameter pair for online classification settings. Four categories of data are highlighted in this figure. The first is the blue markers that represent non-optimal window/over pairs. The second (orange diamond) and third (orange square) categories represent identified feasible data that feature increased classification accuracy over the first category while not substantially increasing compute time. The fourth category (orange star) represents the optimal classification setting for both agent and swarm settings; however, we see disproportionately higher compute times in both sub-figures. The best-identified overall observation window periods and window overlaps are either $[20, 0.5]$ or $[40, 0.75]$, with proportional classification accuracy and computation time across all three settings. We suggest these window settings from this sensitivity analysis for agent and swarm-type profile classifications in swarm shepherding settings.

\subsection{Interaction Dynamics}

The objective of our final analysis with information markers is to develop statistics of the network among agents, developing an understanding of \emph{role} and \emph{tactic} concepts from Figure~\ref{fig:MarkerOntology} that focuses on non-stationary information about the agents and swarm, using the markers identified in Table~\ref{table:2x2matrix} at the agent and swarm levels. This analysis addresses an aspect of the challenges introduced earlier in the paper, specifically identifying those critical pieces of information that discriminate particular states or strategies.

The first interaction dynamics analysis focuses on the agent level, where our objective is to identify agent associations and interaction distributions. Algorithm~\ref{algo:association} summarises the following method outlined. For each marker observation period ($M^p$), we calculate and build a sub-state of identified markers, calculating statistics from markers about each agent $(\mathcal{M}_{\pi_i})$. We use the statistics for summarising each agent's state in reference to all other swarm agents. We achieve this by normalising each agent's marker sub-state as a proportion of the total, calculating each marker independently $(\sfrac{M^p}{\vert\vert M^p \vert\vert})$. We obtain an interaction state vector for each agent, with each value in the vector being a summary measure of interactions calculated through each marker. We summarise this vector by calculating the $L1$-Norm $(\vert\vert \mathcal{M}_{\pi_i} \vert\vert_1)$, normalised for that observation period with the swarm.

Our goal is to calculate the association of each agent with other agents across the evolution of a scenario $(\pi_i\rightarrow\pi_j,~i\neq j)$, assuming we know the number of agent types in the swarm; this could be calculated prior, such as using the classifier methods introduced previously. For each marker observation period, we cluster all swarm agents using the k-means algorithm based on the number of agent profiles in the swarm, establishing an undirected, unweighted adjacency matrix for agent connectivity $(A(\pi_i))$. If an agent $\pi_i$ is in the same cluster as another agent $\pi_j$ in that period, then we say that the agents are connected with weight one else; if not in the same cluster, then we say that $\pi_i$ and $\pi_j$ do not share an edge. Our method is somewhat similar to the clustering coefficient discussed by Novelli \& Lizier~\citeyear{10.1162/netn_a_00178} and partially inspired by the early work of Li et al.,~\citeyear{Li2004:CollabSwarm}, who propose a clustering method to estimate swarm diversity and specialisation. Across all observation periods, we generate the graph and calculate centrality statistics based on a $\pi_i$-degree.

\begin{figure}[ht]
        \centering
        \includegraphics[width=0.5\textwidth]{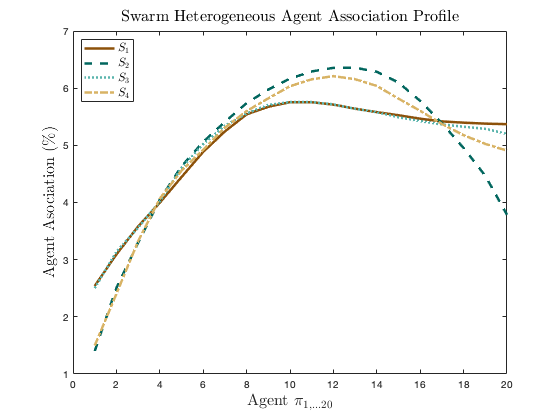}
        \caption{Agent association is calculated from data in Table~\ref{table:M2summaryData}, given as the pairwise propensity of agents to cluster together. We calculate the agent association for a marker observation period by first clustering agents using the $k$-means algorithm; we assume that the number of swarm agent types in the swarm is known or able to be determined, such as described for the agent type profile classifications in Table~\ref{table:AgentProfileE1}, Table~\ref{table:AgentProfileE2} and Table~\ref{table:AgentTestAccuracy}. We build a binary agent adjacency matrix and calculate the centrality of agents using degree importance, normalising each agent's association across the swarm. The resulting mean percentage value of agent association is visualised. We interpret lower association values as an agent with a desire to associate with different agents across a scenario, measuring traits such as \textit{gregariousness}. When considered in conjunction with Figure~\ref{fig:M2ScenarioComparison} and Figure~\ref{fig:M4combined}, we can establish agent role profiles, for instance, suggesting that $\pi_1$ (type $A_1$) in $S_4$ associates with many different agents in the swarm, accounting for a high proportion of total swarm interactions. In contrast, $\pi_{13}$ (type $A_7$) in $S_4$ more frequently associates with the same collection of agents while accounting for a below-average proportion of the total swarm interactions.}
        \label{fig:M2agentAssociation}
    \end{figure}

\begin{figure}[ht]
        \centering
        \includegraphics[width=0.5\textwidth]{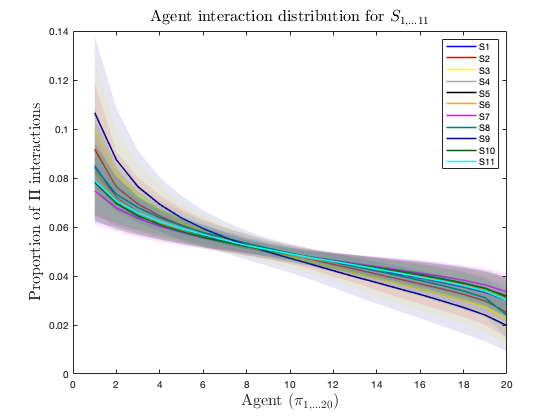}
        \caption{Scenario distribution of agent interactions, visualising data summarised in Table~\ref{table:M2summaryData}. Note that $\pi_i$ are sorted for the largest to smallest proportion of $\Pi$ interactions. We observe a non-linear distribution of agent interactions across each scenario, typically with non-negative skew.}
        \label{fig:M2ScenarioComparison}
    \end{figure}

The use of network analysis to generate statistics on the agent interaction is well established, with many examples proposed in the literature~\shortcite{Rezaei2022:VitalNode,Shang2014:SwarmTopo,6973878,10.1162/isal_a_00229}. We define the scenario agent association score $(\mathcal{A}_{\pi_i})$ as the proportion of total pairwise interactions an agent has across a scenario (propensity of cluster association). Table~\ref{table:M2summaryData} summarises these calculations across all scenarios, subsequently depicted via the mean percentage value of agent association in Figure~\ref{fig:M2ScenarioComparison}, with heterogeneous scenarios depicted in Figure~\ref{fig:M2agentAssociation}. We interpret lower association values as an agent wanting to associate with different agents across a scenario, measuring traits such as \textit{gregariousness}~\shortcite{HAUSCHILDT201615}. We compare an agent's association across the four heterogeneous scenarios; our goal here is to demonstrate that each parameterised agent type possesses a unique association profile. Using the ANOVA test, we conclude that there are differences between each agent type association profile ($F_{19,60}=52.66, p<0.001$).

\begin{algorithm}[ht]
    \caption{Agent Association ($\pi_i\in\Pi$)}\label{algo:association}
    \begin{algorithmic}[1]
        \State Set observation window size and window overlap
        \For{$M^p$}
            \For{$\pi_i\in\Pi$} 
                \State Calculate $\mathcal{M}_{\pi_i}^p$ \Comment{Marker $p$ for agent $\pi_i$.}
            \EndFor
            \State Summarise $\mathcal{M}^p$ \Comment{Marker-wise $\pi_i$-vector.}
            \State Normalise $M^p,~\text{such that}~\sfrac{M^p}{\vert\vert M^p \vert\vert}$
            \State Calculate $\vert\vert\mathcal{M}_{\pi_i}\vert\vert_1$ \Comment{L1-norm.}
            \State Calculate $k$-clusters $\forall~\pi_i\in\Pi$
            \State Build $A(\pi_i)~\forall~\pi_i\in\Pi$
        \EndFor
        \State Calculate $\mathcal{A}_{\pi_i}=\sum{A(\pi_i)}$ \Comment{Cumulative associations.}
        \State \textbf{Return} $\mathcal{A}_{\pi_i}\forall\pi_i\in\Pi$ \Comment{Agent association.}
    \end{algorithmic}
\end{algorithm}

    \begin{table}[ht]
        \centering
        \begin{tabular}[width=0.475\textwidth]{@{}cccc@{}}
            \toprule
            \textbf{Scenario} & \textbf{Max} & \textbf{Min} & \textbf{Range} \\ \midrule
            S1       & 28.76                            & 0.57                             & 19.28          \\
            S2       & 22.94                            & 0.49                             & 13.49          \\
            S3       & 23.71                            & 0.54                             & 14.20          \\
            S4       & 20.27                            & 0.62                             & 11.50          \\
            S5       & 20.64                            & 0.65                             & 11.91          \\
            S6       & 23.12                            & 0.67                             & 14.27          \\
            S7       & 17.93                            & 0.76                             & 9.91           \\
            S8       & 20.76                            & 0.63                             & 11.26          \\
            S9       & 25.62                            & 0.49                             & 16.19          \\
            S10      & 27.01                            & 0.63                             & 18.17          \\
            S11      & 14.37                            & 0.78                             & 6.42           \\
            \bottomrule
        \end{tabular}
        \caption{Summary of agent interactions (non-stationary traits) for each scenario given as percentages, as depicted in Figure~\ref{fig:M2agentAssociation} and Figure~\ref{fig:M2ScenarioComparison}. Agent statistics report the maximum, minimum and mean contributions across all interactions. Markers used for this analysis are as discussed in Table~\ref{table:2x2matrix}. Our objective is to identify agent network characteristics to evaluate individual $\pi_i$ and collective $\Pi$ over time, for instance, to assess the change in agent roles.}
        \label{table:M2summaryData}
    \end{table}

\begin{figure*}[ht]
        \begin{subfigure}{0.5\textwidth}
            \centering
            \includegraphics[width=\textwidth]{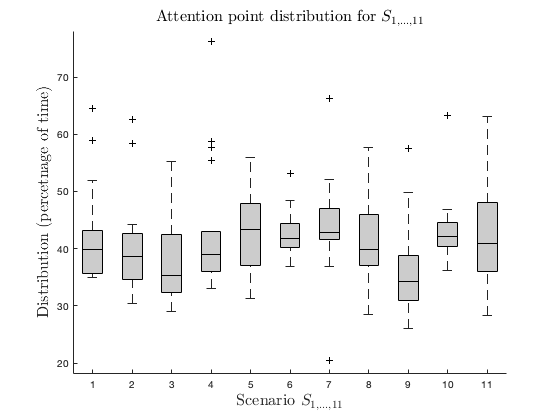}
            \caption{Scenario distribution of attention points, visualising data summarised in Table~\ref{table:M4summaryData} for $\eta = 0.5$. }
            \label{fig:M4AttentionPoint}
        \end{subfigure}
        \hfill
        \begin{subfigure}{0.5\textwidth}
            \centering
            \includegraphics[width=\textwidth]{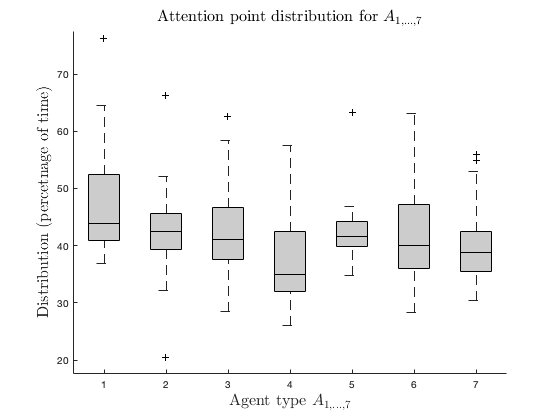}
            \caption{Agent distribution of attention points, visualising data from Table~\ref{table:M4summaryData} organised by agent type for $\eta = 0.5$.}
            \label{fig:M4AgentAttention}
        \end{subfigure}
        \caption{The selection of $\eta$ impacts the distribution of agents as attention points, with $\eta\rightarrow~0$ observing an increased variance and $\eta\rightarrow~100$ observing a decreased variance of attention point distributions. The agent perspective (Figure~\ref{fig:M4AgentAttention}) highlights the potential roles of an agent type within different swarms, quantifying interaction behaviours.}
        \label{fig:M4combined}
    \end{figure*}

When considered in conjunction with Figure~\ref{fig:M2ScenarioComparison} and Figure~\ref{fig:M4AttentionPoint} we can establish agent role profiles, for instance suggesting that $\pi_1$ (type $A_1$) in $S_4$ associates with many different agents in the swarm, accounting for high proportion of total swarm interactions. In contrast, $\pi_{13}$ (type $A_7$) in $S_4$ (Figure~\ref{fig:M2ScenarioComparison}) more frequently associates with the same collection of agents while accounting for a below-average proportion of the total swarm interactions. Relative to other agents in a swarm, we can begin to detect non-stationary swarm roles. This may help to identify agent adaptation and learning over time, particularly for cognitive settings where an agent's \emph{desire} may be stationary; however, their swarm role may not be. This could be of interest for analysing differences between homogeneous and heterogeneous swarms, particularly the configuration of constituent agents. When further considered with a measure of interactions as given in Figure~\ref{fig:M2ScenarioComparison}, we could begin to assign leadership and followership roles in a swarm~\shortcite{garland:2018}.

The second interaction dynamics analysis focuses on the swarm level, where our objective is to identify \emph{attention points} across the swarm, building on our understanding of agent associations. Algorithm~\ref{algo:attention} summarises the following method outlined. This analysis builds from that outlined in Algorithm~\ref{algo:association}, branching after the calculation of the L1-norm $(\vert\vert M_{pi_i} \vert\vert_1)$. We employ a user-defined threshold, $\eta~\in~(0, 1]$, selecting the set ($Q$) of minimum number of $\pi_i$ agents where the cumulative sum of values is greater than or equal to $\eta$, given as $Q = \min_{\pi_i}\text{such that}\sum \vert\vert M_{\pi_i} \vert\vert_1\geq\eta$. An agent who is a member of the set $Q$ is considered an attention point for the given marker observation period. Table~\ref{table:M4summaryData} summarises the distribution of swarm attention points over each scenario for $\eta = 0.5$. In Figure~\ref{fig:M4combined}, we illustrate both the scenario and agent perspective of attention point distributions, particularly highlighting the variance over each agent type across all scenarios (Figure~\ref{fig:M4AgentAttention}). 

We conduct ANOVA to test for statistically significant differences in each sub-figure. For Figure~\ref{fig:M4AttentionPoint}, we fail to reject the null hypothesis ($F_{10,209}=1.72, p>0.05$) and conclude that there is insufficient evidence to detect a difference in the scenario attention point distributions. We expect this outcome as Algorithm~\ref{algo:attention} considers the agent interaction and not the scenario context of the agent, supporting our interpretation that for a constant of $\eta$, each scenario should contain similar distribution properties. For Figure~\ref{fig:M4AgentAttention} we reject the null hypothesis ($F_{6,217}=5.64, p<0.001$) and conclude that there are differences between agent types across all scenarios, returning a result that supports the claims made regarding Figure~\ref{fig:M2agentAssociation} in earlier sections. This is the expected outcome, as each agent type is designed with distinct interaction properties. Selection of the attention point threshold $\eta$ impacts the granularity of insights on the swarm, for instance, where a low threshold may be used to identify individual \emph{centres of influence}~\shortcite{Hepworth2020:Footprints} or a high threshold be used to identify a stable \emph{centre of mass}~\shortcite{Strombom:2014}. Values of $\eta\rightarrow~1$ will increase the number of agents considered as attention points in the swarm, whereas $\eta\rightarrow~0$ will observe fewer agents.

\begin{algorithm}[ht]
    \caption{Swarm Attention Points ($\Pi$)}\label{algo:attention}
    \begin{algorithmic}[1]
        \State Set $\eta$
        \State Sort $\vert\vert M_{\pi_i} \vert\vert$ \Comment{Sort descending from Algorithm~\ref{algo:association}.}
        \For{$i = 1,\dots,\vert\Pi\vert$}
            \If{$\sum_i \vert\vert M_{\pi_i} \vert\vert > \eta$}
                \State Set $Q(\pi_i) = 0$
            \Else
                \State Set $Q(\pi_i) = 1$
            \EndIf
        \EndFor
        \State \textbf{Return} Q \Comment{Vector of swarm attention points.}
    \end{algorithmic}
\end{algorithm}

    \begin{table}[ht]
        \centering
        \begin{tabular}[width=0.475\textwidth]{@{}ccccccc@{}}
            \toprule
            \textbf{Scenario} & \textbf{Mean} & \textbf{Std. Dev.} & \textbf{Range} & \textbf{Max} & \textbf{Min}\\
            \midrule
                S1       & 42.18 & 8.27  & 29.61 & 64.53 & 34.93 \\
                S2       & 40.20 & 8.03  & 32.18 & 62.63 & 30.45 \\
                S3       & 38.21 & 7.97  & 26.28 & 55.26 & 28.98 \\
                S4       & 42.93 & 10.87 & 43.27 & 76.22 & 32.96 \\
                S5       & 43.07 & 7.14  & 24.72 & 55.93 & 31.20 \\
                S6       & 42.57 & 3.93  & 16.24 & 53.10 & 36.86 \\
                S7       & 43.68 & 8.46  & 45.87 & 66.28 & 20.41 \\
                S8       & 41.11 & 7.21  & 29.13 & 57.68 & 28.55 \\
                S9       & 36.36 & 8.29  & 31.37 & 57.46 & 26.08 \\
                S10      & 43.13 & 5.56  & 27.15 & 63.36 & 36.22 \\
                S11      & 42.60 & 9.02  & 34.90 & 63.18 & 28.29 \\
            \bottomrule
        \end{tabular}
        \caption{Summary swarm attention points given as percentages, as depicted in Figure~\ref{fig:M4AttentionPoint}. Agent statistics are reported as mean percentages for each scenario, where $100$\% is the total scenario length. Our objective here is to identify swarm attention points, defined as an agent with traits of focus in the swarm. For each observation period, an agent is considered  to be an attention point if they are a member of the set, $k$.}
        \label{table:M4summaryData}
    \end{table}

\subsection{Findings Summary}

Our objective in this work has been to introduce information markers based on the low-level positional information of swarm agents to understand the individual and collective behaviour of the swarm. We summarise our findings relative to the design outlined in Table~\ref{table:2x2matrix}, highlighting an example range of analysis and recognition insights possible with the information markers framework for decision-making. The classification analysis demonstrates that information markers can discover an agent and swarm profiles and deliver meaning about the swarm. With stationary information traits on an agent, we have shown that markers can discover agent desires and traits as profiles to understand the strength of response and interaction-type similarity. The swarm stationary information traits extend this to characterise the type of swarm with information markers, be it homogeneous or heterogeneous. We have shown that heterogeneity presents additional challenges to identifying information vectors and understanding the role of an agent in a swarm, often not required in the models used for homogeneous swarming. Information markers overcome this challenge by dissecting these elements and uncovering differences between agents, their interactions and the impact on the collective.  

For agent non-stationary traits, pairwise information markers quantify the relationship of interactions between agents and their associations over time, identifying the agent's role and how this changes concerning a swarm's configuration. Understanding the movement complexity and coordination similarity highlights the importance of interactions between each agent type in the environment. Extending this to the swarm level, we enable the assessment of attention points within the swarm as the link between agent and swarm level indicators, where detecting change points in the evolution of tactic execution (interaction types) identifies to an observer specific agents to focus on as potentially crucial in a swarm. The non-stationary markers situate an agent in the swarm and enable us to assess change through time. Figure~\ref{fig:MarkerOntology} captures the relationship between the stationary and non-stationary aspects, where the behavioural responses of an agent do not change (stationary traits), but the nature of the interaction (non-stationary traits) does, informed by the environmental context that an agent is situated.

\section{Conclusion}

We have designed information markers to detect changes in context across a swarm and its agents, framed within the setting of swarm shepherding. Our review of the literature with this particular perspective grounds the context recognition performance of the information markers. Our objectives were to evaluate situation and context recognition performance for markers across both homogeneous and heterogeneous settings, investigating the value of markers as inputs to an agent recognition system. In this work, we have presented information markers as a method to recognise the situations and infer particular contexts of a swarm as a unified approach to analysing swarms. 
The experimental analyses confirmed the recognition power of the information markers in answering questions on both the agent and swarm levels. Notably, the information markers enable identifying agent behavioural characteristics and
interaction tendencies. At the swarm level, the markers identify points of influence critical to induce change in the swarm state. These types of information are essential for an observer to get insights into swarm intricacies and for a control agent to plan for the best courses of action for a given objective.

As presented in this work, we indicate several new directions to expand on information markers for recognition. The first future direction is for marker selection. In this study, we employ an inclusive policy that enumerates the measures and metrics in the swarm literature. While suitable to ensure we uncover all aspects of an agent, this could become computationally expensive for settings with a magnitude increase in the number of markers identified. 

The second new direction considers adaptive swarm agents. Information markers support an external observer in understanding the context of a swarm and its agents. The analyses conducted in this work were limited to swarm agents with static profiles where an agent's traits do not change over time or in response to environmental conditions. Future studies should relax these assumptions to consider agent adaptation through a scenario or learning over multiple scenarios, such as the introduction of environmental obstacles, additional influence vectors or adversarial agents. Empirical studies in biological settings support the inclusion of swarm agent adaptation and learning; evaluating the effectiveness of information markers across these settings offers many rich opportunities to enhance the understanding of decision-making processes in swarms. The third new direction is for use in swarm control, where a control agent is supported with information markers to recognise the swarm scenario and select an appropriate behavioural response. Evaluating the comparative performance of a markers-enabled agent to a non-markers-enabled agent in a range of homogeneous and non-homogeneous settings offers an exciting extension to this work. 

\begin{dci}
    All authors declared that they have no conflicts for the research, authorship, and/or publication of this article.
\end{dci}

\begin{funding}
    The authors received no financial support for the research, authorship, and/or publication of this article. 
\end{funding}

\bibliographystyle{mslapa}
\bibliography{bibliography}

\begin{thebibliography}{}

\bibitem[\protect\citeauthoryear{Abbass \& Hunjet}{Abbass \&
  Hunjet}{2021}{a}]{abbass2020a}
Abbass, H. \& Hunjet, R. (2021a).
\newblock Smart shepherding: Towards transparent artificial intelligence
  enabled human-swarm teams.
\newblock In {\em Shepherding UxVs for Human-Swarm Teaming: An Artificial
  Intelligence Approach to Unmanned X Vehicles} (p.\ 1–28). Springer.

\bibitem[\protect\citeauthoryear{Abbass \& Hunjet}{Abbass \&
  Hunjet}{2021}{b}]{Abbass2020:UxV}
Abbass, H.~A. \& Hunjet, R.~A. (2021b).
\newblock {\em Shepherding UxVs for Human-Swarm Teaming: An Artificial
  Intelligence Approach to Unmanned X Vehicles}.
\newblock Springer.

\bibitem[\protect\citeauthoryear{Almeida, Costa \& Guizzardi}{Almeida
  et~al.}{2018}{}]{8423994}
Almeida, J. P.~A., Costa, P.~D. \& Guizzardi, G. (2018).
\newblock Towards an ontology of scenes and situations.
\newblock In {\em Conference on Cognitive and Computational Aspects of
  Situation Management (CogSIMA)} (pp.\ 29--35). IEEE.

\bibitem[\protect\citeauthoryear{Pilkiewicz~et al.}{Pilkiewicz~et
  al.}{2020}{}]{pilkiewicz:2020}
Pilkiewicz~et al., K. (2020).
\newblock Decoding collective communications using information theory tools.
\newblock {\em J. R. Soc. Interface}, {\em 17: 20190563}.

\bibitem[\protect\citeauthoryear{Valentini~et al.}{Valentini~et
  al.}{2019}{a}]{Valentini:2019b}
Valentini~et al., G. (2019a).
\newblock Revealing the structure of information flows discriminates similar
  animal social behaviors.
\newblock {\em bioRxiv preprint}.

\bibitem[\protect\citeauthoryear{Amornbunchornvej}{Amornbunchornvej}{2021}{}]{mFLICA:2021}
Amornbunchornvej, C. (2021).
\newblock mflica: An r package for inferring leadership of coordination from
  time series.
\newblock {\em SoftwareX}, {\em 15}, 100781.

\bibitem[\protect\citeauthoryear{Andrade, Blunsden \& Fisher}{Andrade
  et~al.}{2006}{}]{1698861}
Andrade, E., Blunsden, S. \& Fisher, R. (2006).
\newblock Modelling crowd scenes for event detection.
\newblock In {\em 18th International Conference on Pattern Recognition
  (ICPR'06)}, Volume~1 (pp.\ 175--178). ICPR.

\bibitem[\protect\citeauthoryear{Baldi \& Frasca}{Baldi \&
  Frasca}{2019}{}]{BALDI2019935}
Baldi, S. \& Frasca, P. (2019).
\newblock Adaptive synchronization of unknown heterogeneous agents: An adaptive
  virtual model reference approach.
\newblock {\em Journal of the Franklin Institute}, {\em 356(2)}, 935--955.
\newblock Special Issue on Modeling, Analysis and Control of Networked
  Autonomous Agents.

\bibitem[\protect\citeauthoryear{Basak}{Basak}{2021}{}]{Basak2021:leaderFollower}
Basak, U.~S. (2021).
\newblock {\em Study on Identification of Leader and Follower Agents and its
  Interaction Domain from Trajectories in a Collectively Moving Colony}.
\newblock PhD thesis, Hokkaido University.

\bibitem[\protect\citeauthoryear{Baumann \& B{\"u}ning}{Baumann \&
  B{\"u}ning}{2016}{}]{baumann2016learning}
Baumann, M. \& B{\"u}ning, H.~K. (2016).
\newblock {\em Learning shepherding behavior.}
\newblock PhD thesis, University of Paderborn.

\bibitem[\protect\citeauthoryear{Baxter, Hepworth, Joiner \& Abbass}{Baxter
  et~al.}{2021}{}]{Baxter:HST-GO}
Baxter, D.~P., Hepworth, A.~J., Joiner, K.~F. \& Abbass, H. (2021).
\newblock On the premise of a swarm guidance ontology for human-swarm teaming.
\newblock In {\em 65th Annual Meeting of the Human Factors and Ergonomics
  Society}. Baltimore, Maryland: Human Factors and Ergonomics Society.

\bibitem[\protect\citeauthoryear{Bossomaier, Barnett, Harr{\'e} \&
  Lizier}{Bossomaier et~al.}{2016}{}]{Bossomaier2016}
Bossomaier, T., Barnett, L., Harr{\'e}, M. \& Lizier, J.~T. (2016).
\newblock {\em An Introduction to Transfer Entropy: Information Flow in Complex
  Systems}.
\newblock Cham: Springer International Publishing.

\bibitem[\protect\citeauthoryear{Brown \& Goodrich}{Brown \&
  Goodrich}{2014}{a}]{brown2014a}
Brown, D. \& Goodrich, M. (2014a).
\newblock Limited bandwidth recognition of collective behavours in bio-inspired
  swarms.
\newblock In {\em Proceedings of the 13th International Conference on
  Autonomous Agents and Multiagent Systems (AAMAS 2014}. Paris, France: ACM.

\bibitem[\protect\citeauthoryear{Brown \& Goodrich}{Brown \&
  Goodrich}{2014}{b}]{10.5555/2615731.2615798}
Brown, D.~S. \& Goodrich, M.~A. (2014b).
\newblock Limited bandwidth recognition of collective behaviors in bio-inspired
  swarms.
\newblock In {\em Proceedings of the 2014 International Conference on
  Autonomous Agents and Multi-Agent Systems}, AAMAS '14 (p.\ 405–412).
  Richland, SC: International Foundation for Autonomous Agents and Multiagent
  Systems.

\bibitem[\protect\citeauthoryear{Butail, Mwaffo \& Porfiri}{Butail
  et~al.}{2016}{}]{PhysRevE.93.042411}
Butail, S., Mwaffo, V. \& Porfiri, M. (2016).
\newblock Model-free information-theoretic approach to infer leadership in
  pairs of zebrafish.
\newblock {\em Phys. Rev. E}, {\em 93}, 042411.

\bibitem[\protect\citeauthoryear{Chakraborty, Bhunia \& De}{Chakraborty
  et~al.}{2020}{}]{Chakraborty2020}
Chakraborty, D., Bhunia, S. \& De, R. (2020).
\newblock Survival chances of a prey swarm: how the cooperative interaction
  range affects the outcome.
\newblock {\em Scientific Reports}, {\em 10}.

\bibitem[\protect\citeauthoryear{Cofta, Ledziński, Śmigiel \&
  Gackowska}{Cofta et~al.}{2020}{}]{e22060597}
Cofta, P., Ledziński, D., Śmigiel, S. \& Gackowska, M. (2020).
\newblock Cross-entropy as a metric for the robustness of drone swarms.
\newblock {\em Entropy}, {\em 22(6)}.

\bibitem[\protect\citeauthoryear{Crosato, Jiang, Lecheval, Lizier, Wang,
  Tichit, Theraulaz \& Prokopenko}{Crosato et~al.}{2018}{}]{crosatoSwarmIntell}
Crosato, E., Jiang, L., Lecheval, V., Lizier, J.~T., Wang, X.~R., Tichit, P.,
  Theraulaz, G. \& Prokopenko, M. (2018).
\newblock Informative and misinformative interactions in a school of fish.
\newblock {\em Swarm Intell} (p.\ 283–305).

\bibitem[\protect\citeauthoryear{D, G, T-H, AO \& D}{D
  et~al.}{2020}{}]{trendafilov:2020}
D, T., G, S., T-H, H., AO, E. \& D, P. (2020).
\newblock Tilting together: An information-theoretic characterization of
  behavioral roles in rhythmic-dyadic interaction.
\newblock {\em Front. Hum. Neurosci.}, {\em 14:185}.

\bibitem[\protect\citeauthoryear{Davis}{Davis}{2000}{}]{Davis2000}
Davis, G.~B. (2000).
\newblock {\em Information Systems Conceptual Foundations: Looking Backward and
  Forward} (pp.\ 61--82).
\newblock Boston, MA: Springer US.

\bibitem[\protect\citeauthoryear{Diukman}{Diukman}{2012}{}]{Diukman:2012}
Diukman, A.~G. (2012).
\newblock Swarm observations implementing integration theory to understand an
  opponent swarm.
\newblock Master's thesis, Monterey, California. Naval Postgraduate School.

\bibitem[\protect\citeauthoryear{El-Fiqi, Campbell, Elsayed, Perry, Singh,
  Hunjet \& Abbass}{El-Fiqi et~al.}{2020}{}]{9256255}
El-Fiqi, H., Campbell, B., Elsayed, S., Perry, A., Singh, H.~K., Hunjet, R. \&
  Abbass, H.~A. (2020).
\newblock The limits of reactive shepherding approaches for swarm guidance.
\newblock {\em IEEE Access}, {\em 8}, 214658--214671.

\bibitem[\protect\citeauthoryear{Emmanuel, Maupong, Mpoeleng, Semong, Mphago \&
  Tabona}{Emmanuel et~al.}{2021}{}]{Emmanuel2021:MissingData}
Emmanuel, T., Maupong, T., Mpoeleng, D., Semong, T., Mphago, B. \& Tabona, O.
  (2021).
\newblock A survey on missing data in machine learning.
\newblock {\em J. Big Data}, {\em 8}.

\bibitem[\protect\citeauthoryear{Ferber}{Ferber}{1999}{}]{Ferber:1999:MSI:520715}
Ferber, J. (1999).
\newblock {\em Multi-Agent Systems: An Introduction to Distributed Artificial
  Intelligence\/} (1st Ed.).
\newblock Boston, MA, USA: Addison-Wesley Longman Publishing Co., Inc.

\bibitem[\protect\citeauthoryear{Fernandez-Rojas, Perry, Singh, Campbell,
  Elsayed, Hunjet \& Abbass}{Fernandez-Rojas et~al.}{2019}{}]{8658079}
Fernandez-Rojas, R., Perry, A., Singh, H., Campbell, B., Elsayed, S., Hunjet,
  R. \& Abbass, H.~A. (2019).
\newblock Contextual awareness in human-advanced-vehicle systems: A survey.
\newblock {\em IEEE Access}, {\em 7}, 33304--33328.

\bibitem[\protect\citeauthoryear{Garland, Berdahl, Sun \& Bollt}{Garland
  et~al.}{2018}{}]{garland:2018}
Garland, J., Berdahl, A.~M., Sun, J. \& Bollt, E.~M. (2018).
\newblock Anatomy of leadership in collective behaviour.
\newblock {\em Chaos}, {\em 28}.

\bibitem[\protect\citeauthoryear{Gleiss, Wilson \& Shepard}{Gleiss
  et~al.}{2011}{}]{ODBA:2011}
Gleiss, A., Wilson, R. \& Shepard, E. (2011).
\newblock Making overall dynamic body acceleration work: On the theory of
  acceleration as a proxy for energy expenditure.
\newblock {\em Methods in Ecology and Evolution} (pp.\ 23--33).

\bibitem[\protect\citeauthoryear{Gong, Kang, Walton, Kaminer \& Park}{Gong
  et~al.}{2020}{}]{doi:10.2514/1.G004115}
Gong, Q., Kang, W., Walton, C., Kaminer, I. \& Park, H. (2020).
\newblock Partial observability analysis of an adversarial swarm model.
\newblock {\em Journal of Guidance, Control, and Dynamics}, {\em 43(2)},
  250--261.

\bibitem[\protect\citeauthoryear{Haeri, Jerath \& Leachman}{Haeri
  et~al.}{2020}{}]{10.1115/1.4046580}
Haeri, H., Jerath, K. \& Leachman, J. (2020).
\newblock {Thermodynamics-Inspired Macroscopic States of Bounded Swarms}.
\newblock {\em ASME Letters in Dynamic Systems and Control}, {\em 1(1)}.
\newblock 011015.

\bibitem[\protect\citeauthoryear{Hamann, Schmickl \& Crailsheim}{Hamann
  et~al.}{2011}{}]{Hamann2012:Emergence}
Hamann, H., Schmickl, T. \& Crailsheim, K. (2011).
\newblock Explaining emergent behavior in a swarm system based on an inversion
  of the fluctuation theorem.
\newblock In {\em The 11th European Conference on Artificial Life}. ECAL.

\bibitem[\protect\citeauthoryear{Hasbach \& Bennewitz}{Hasbach \&
  Bennewitz}{2021}{}]{Hasbach2021:HSI}
Hasbach, J.~D. \& Bennewitz, M. (2021).
\newblock The design of self-organizing human–swarm intelligence.
\newblock {\em Adaptive Behavior}, {\em 30(4)}, 361--386.

\bibitem[\protect\citeauthoryear{Hauschildt \& Gerken}{Hauschildt \&
  Gerken}{2016}{}]{HAUSCHILDT201615}
Hauschildt, V. \& Gerken, M. (2016).
\newblock Effect of pasture size on behavioural synchronization and spacing in
  german blackface ewes (ovis aries).
\newblock {\em Behavioural Processes}, {\em 124}, 15--22.

\bibitem[\protect\citeauthoryear{Hepworth}{Hepworth}{2021}{}]{Hepworth2021:ARS}
Hepworth, A. (2021).
\newblock Activity recognition for shepherding.
\newblock In H.~Abbass \& R.~Hunjet (Eds.), {\em Shepherding UxVs for
  Human-Swarm Teaming}  chapter~7,  (pp.\ 131--164). Springer, Cham.

\bibitem[\protect\citeauthoryear{Hepworth, Baxter \& Abbass}{Hepworth
  et~al.}{2022}{}]{Onto4MAT:2022}
Hepworth, A.~J., Baxter, D.~P. \& Abbass, H.~A. (2022).
\newblock Onto4mat: A swarm shepherding ontology for generalized multiagent
  teaming.
\newblock {\em IEEE Access}, {\em 10}, 59843--59861.

\bibitem[\protect\citeauthoryear{{Hepworth}, {Baxter}, {Hussein}, {Yaxley},
  {Debie} \& {Abbass}}{{Hepworth} et~al.}{2021}{}]{Hepworth2021:HST3}
{Hepworth}, A.~J., {Baxter}, D.~P., {Hussein}, A., {Yaxley}, K.~J., {Debie}, E.
  \& {Abbass}, H.~A. (2021).
\newblock Human-swarm-teaming transparency and trust architecture.
\newblock {\em IEEE/CAA J. Autom. Sinica}, {\em 8(7)}, 1281–1295.

\bibitem[\protect\citeauthoryear{{Hepworth}, {Yaxley}, {Baxter}, {Joiner} \&
  {Abbass}}{{Hepworth} et~al.}{2020}{}]{Hepworth2020:Footprints}
{Hepworth}, A.~J., {Yaxley}, K.~J., {Baxter}, D.~P., {Joiner}, K.~F. \&
  {Abbass}, H. (2020).
\newblock Tracking footprints in a swarm: Information-theoretic and spatial
  centre of influence measures.
\newblock In {\em Symposium Series on Computational Intelligence (SSCI)} (pp.\
  2217--2224). IEEE.

\bibitem[\protect\citeauthoryear{Himo, Ogura \& Wakamiya}{Himo
  et~al.}{2022}{}]{Himo2022:HeterogeneousResponse}
Himo, R., Ogura, M. \& Wakamiya, N. (2022).
\newblock Iterative shepherding control for agents with heterogeneous
  responsivity.
\newblock {\em Mathematical Biosciences and Engineering}, {\em 19}, 3509--3525.

\bibitem[\protect\citeauthoryear{Hussein, Petraki, Elsawah \& Abbass}{Hussein
  et~al.}{2022}{}]{Hussein:AAMAS22}
Hussein, A., Petraki, E., Elsawah, S. \& Abbass, H. (2022).
\newblock Autonomous swarm shepherding using curriculum-based reinforcement
  learning.
\newblock In Faliszewski, P., Mascardi, V., Pelachaud, C. \& Taylor, M. (Eds.),
  {\em Proc. of the 21st International Conference on Autonomous Agents and
  Multiagent Systems (AAMAS 2022)}. Auckland, New Zealand: ACM.

\bibitem[\protect\citeauthoryear{Jankovic}{Jankovic}{2018}{}]{Jankovic2018:CFD}
Jankovic, L. (2018).
\newblock Modelling computational fluid dynamics with swarm behaviour.
\newblock In {\em 4th Building Simulation and Optimization Conference (BSO)}
  (pp.\ 112--118). IBPSA.

\bibitem[\protect\citeauthoryear{Jaén-Vargas, Leiva, Fernandes, Goncalves,
  Silva, Lopes \& Olmedo}{Jaén-Vargas
  et~al.}{2022}{}]{JaenVargas2022:WindowSizeAR}
Jaén-Vargas, M., Leiva, K.~R., Fernandes, F., Goncalves, S.~B., Silva, M.~T.,
  Lopes, D. \& Olmedo, J.~S. (2022).
\newblock Effects of sliding window variation in the performance of
  acceleration-based human activity recognition using deep learning models.
\newblock {\em PeerJ Computer Science}, {\em 8(e1052)}, 1--22.

\bibitem[\protect\citeauthoryear{Kathpalia}{Kathpalia}{2021}{}]{Kathpalia:2021}
Kathpalia, A. (2021).
\newblock {\em Theoretical and Experimental Investigations into Causality, its
  Measures and Applications}.
\newblock (ph.d.)., National Institute of Advanced Studies.

\bibitem[\protect\citeauthoryear{Kengyel, Hamann, Zahadat, Radspieler, Wotawa
  \& Schmickl}{Kengyel et~al.}{2015}{}]{Kengyel:2015}
Kengyel, D., Hamann, H., Zahadat, P., Radspieler, G., Wotawa, F. \& Schmickl,
  T. (2015).
\newblock Potential of heterogeneity in collective behaviors: A case study on
  heterogeneous swarms.
\newblock In {\em International Conference on Principles and Practice of
  Multi-Agent Systems} (pp.\ 201--217). PRIMA.

\bibitem[\protect\citeauthoryear{Kleanthous, Hussain, Khan, Sneddon,
  Al-Shamma'a \& Liatsis}{Kleanthous et~al.}{2022}{}]{KLEANTHOUS2022442}
Kleanthous, N., Hussain, A.~J., Khan, W., Sneddon, J., Al-Shamma'a, A. \&
  Liatsis, P. (2022).
\newblock A survey of machine learning approaches in animal behaviour.
\newblock {\em Neurocomputing}, {\em 491}, 442--463.

\bibitem[\protect\citeauthoryear{Lee \& Kim}{Lee \& Kim}{2017}{}]{s17122729}
Lee, W. \& Kim, D. (2017).
\newblock Autonomous shepherding behaviors of multiple target steering robots.
\newblock {\em Sensors}, {\em 17(12)}.

\bibitem[\protect\citeauthoryear{Li, Martinoli \& Abu-Mostafa}{Li
  et~al.}{2004}{}]{Li2004:CollabSwarm}
Li, L., Martinoli, A. \& Abu-Mostafa, Y.~S. (2004).
\newblock Learning and measuring specialization in collaborative swarm systems.
\newblock {\em Adaptive Behavior}, {\em 12(3-4)}, 199--212.

\bibitem[\protect\citeauthoryear{Li, Hu, Liang \& Li}{Li
  et~al.}{2012}{}]{10.1007/978-3-642-34381-0_48}
Li, M., Hu, Z., Liang, J. \& Li, S. (2012).
\newblock Shepherding behaviors with single shepherd in crowd management.
\newblock In Xiao, T., Zhang, L. \& Ma, S. (Eds.), {\em System Simulation and
  Scientific Computing} (pp.\ 415--423). Berlin, Heidelberg: Springer Berlin
  Heidelberg.

\bibitem[\protect\citeauthoryear{Liu, He, Xu, Ding \& Wang}{Liu
  et~al.}{2018}{}]{liu2018a}
Liu, Q., He, M., Xu, D., Ding, N. \& Wang, Y. (2018).
\newblock A mechanism for recognizing and suppressing the emergent behavior of
  uav swarm.
\newblock {\em Mathematical Problems in Engineering} (p.\ 6734923).

\bibitem[\protect\citeauthoryear{{Lizier}, {Prokopenko} \& {Zomaya}}{{Lizier}
  et~al.}{2008}{}]{Lizier:2008}
{Lizier}, J.~T., {Prokopenko}, M. \& {Zomaya}, A.~Y. (2008).
\newblock Local information transfer as a spatiotemporal filter for complex
  systems.
\newblock {\em Phys Rev E}, {\em 77(2)}.

\bibitem[\protect\citeauthoryear{{Long}, {Sammut}, {Sgarioto}, {Garratt} \&
  {Abbass}}{{Long} et~al.}{2020}{}]{Long2020:Comprehensive}
{Long}, N.~K., {Sammut}, K., {Sgarioto}, D., {Garratt}, M. \& {Abbass}, H.~A.
  (2020).
\newblock A comprehensive review of shepherding as a bio-inspired
  swarm-robotics guidance approach.
\newblock {\em IEEE Transactions on Emerging Topics in Computational
  Intelligence}, {\em 4(4)}, 523--537.

\bibitem[\protect\citeauthoryear{Lord, Sun, Ouellette \& Bollt}{Lord
  et~al.}{2016}{}]{7809221}
Lord, W.~M., Sun, J., Ouellette, N.~T. \& Bollt, E.~M. (2016).
\newblock Inference of causal information flow in collective animal behavior.
\newblock {\em IEEE Transactions on Molecular, Biological and Multi-Scale
  Communications}, {\em 2(1)}, 107--116.

\bibitem[\protect\citeauthoryear{Martín~López, Aguilar~de Soto, Madsen \&
  Johnson}{Martín~López et~al.}{2022}{}]{10.1111/2041-210X.13751}
Martín~López, L.~M., Aguilar~de Soto, N., Madsen, P.~T. \& Johnson, M.
  (2022).
\newblock Overall dynamic body acceleration measures activity differently on
  large versus small aquatic animals.
\newblock {\em Methods in Ecology and Evolution}, {\em 13(2)}, 447--458.

\bibitem[\protect\citeauthoryear{Matarić}{Matarić}{1995}{}]{Mataric1995:AdaptiveGroup}
Matarić, M.~J. (1995).
\newblock Designing and understanding adaptive group behavior.
\newblock {\em Adaptive Behavior}, {\em 4(1)}, 51--80.

\bibitem[\protect\citeauthoryear{Mateo, Kuan \& Bouffanais}{Mateo
  et~al.}{2017}{}]{Mateo:2017}
Mateo, D., Kuan, Y.~K. \& Bouffanais, R. (2017).
\newblock Effect of correlations in swarms on collective response.
\newblock {\em Sci Rep}, {\em 7}.

\bibitem[\protect\citeauthoryear{MathWorks}{MathWorks}{2022}{}]{Matlab:xcorr}
MathWorks (2022).
\newblock xcorr: Cross-correlation.

\bibitem[\protect\citeauthoryear{Mavridis, Tirumalai \& Baras}{Mavridis
  et~al.}{2021}{}]{SwarmInteractions:2021}
Mavridis, C., Tirumalai, A. \& Baras, J. (2021).
\newblock Learning swarm interaction dynamics from density evolution.
\newblock {\em arXiv}.

\bibitem[\protect\citeauthoryear{McGivern}{McGivern}{2020}{}]{McGivern2020:Cognition}
McGivern, P. (2020).
\newblock Active materials: minimal models of cognition?
\newblock {\em Adaptive Behavior}, {\em 28(6)}, 441--451.

\bibitem[\protect\citeauthoryear{Mert~Karakaya}{Mert~Karakaya}{2020}{}]{karakaya:2020}
Mert~Karakaya, Maurizio~Porfiri, G.~P. (2020).
\newblock Invasive alien species respond to biologically-inspired robotic
  predators.
\newblock {\em Proc. SPIE 11374, Bioinspiration, Biomimetics, and
  Bioreplication}, {\em X}.

\bibitem[\protect\citeauthoryear{Miller, Wang, Lizier, Prokopenko \&
  Rossi}{Miller et~al.}{2014}{}]{Miller2014}
Miller, J.~M., Wang, X.~R., Lizier, J.~T., Prokopenko, M. \& Rossi, L.~F.
  (2014).
\newblock {\em Measuring Information Dynamics in Swarms} (pp.\ 343--364).
\newblock Berlin, Heidelberg: Springer Berlin Heidelberg.

\bibitem[\protect\citeauthoryear{Miwa, Oishi, Nakagawa, Maeno, Anzai, Kumagai,
  Okano, Tobioka \& Hirooka}{Miwa et~al.}{2015}{}]{Miwa2015:ODBA}
Miwa, M., Oishi, K., Nakagawa, Y., Maeno, H., Anzai, H., Kumagai, H., Okano,
  K., Tobioka, H. \& Hirooka, H. (2015).
\newblock Application of overall dynamic body acceleration as a proxy for
  estimating the energy expenditure of grazing farm animals: relationship with
  heart rate.
\newblock {\em PLoS ONE}, {\em 10}.

\bibitem[\protect\citeauthoryear{Mocanu, Exarchakos \& Liotta}{Mocanu
  et~al.}{2014}{}]{6973878}
Mocanu, D.~C., Exarchakos, G. \& Liotta, A. (2014).
\newblock Node centrality awareness via swarming effects.
\newblock In {\em 2014 IEEE International Conference on Systems, Man, and
  Cybernetics (SMC)} (pp.\ 19--24). IEEE.

\bibitem[\protect\citeauthoryear{Mohamed, Elsayed, Hunjet \& Abbass}{Mohamed
  et~al.}{2021}{}]{9504706}
Mohamed, R.~E., Elsayed, S., Hunjet, R. \& Abbass, H. (2021).
\newblock A graph-based approach for shepherding swarms with limited sensing
  range.
\newblock In {\em 2021 Congress on Evolutionary Computation (CEC)} (pp.\
  2315--2322). IEEE.

\bibitem[\protect\citeauthoryear{Mould, Regens, III \& Edger}{Mould
  et~al.}{2014}{}]{MouldCounterTerrorActivityRecognition}
Mould, N., Regens, J.~L., III, C. J.~J. \& Edger, D.~N. (2014).
\newblock Video surveillance and counterterrorism: the application of
  suspicious activity recognition in visual surveillance systems to
  counterterrorism.
\newblock {\em Journal of Policing, Intelligence and Counter Terrorism}, {\em
  9(2)}, 151--175.

\bibitem[\protect\citeauthoryear{Nagaraj, Balasubramanian \& Dey}{Nagaraj
  et~al.}{2013}{}]{ETC:2013}
Nagaraj, N., Balasubramanian, K. \& Dey, S. (2013).
\newblock A new complexity measure for time series analysis and classification.
\newblock {\em The European Physical Journal Special Topics}, {\em 222}.

\bibitem[\protect\citeauthoryear{Nguyen, Garratt, Bui \& Abbass}{Nguyen
  et~al.}{2020}{}]{Hung2020:noise}
Nguyen, H.~T., Garratt, M., Bui, L.~T. \& Abbass, H. (2020).
\newblock Disturbances in influence of a shepherding agent is more impactful
  than sensorial noise during swarm guidance.
\newblock In {\em 2020 Symposium Series on Computational Intelligence}, Volume
  abs/2008.12708. IEEE.

\bibitem[\protect\citeauthoryear{Novelli \& Lizier}{Novelli \&
  Lizier}{2021}{}]{10.1162/netn_a_00178}
Novelli, L. \& Lizier, J.~T. (2021).
\newblock {Inferring network properties from time series using transfer entropy
  and mutual information: Validation of multivariate versus bivariate
  approaches}.
\newblock {\em Network Neuroscience}, {\em 5(2)}, 373--404.

\bibitem[\protect\citeauthoryear{Nowak, Porter, Blache \& Dwyer}{Nowak
  et~al.}{2008}{}]{Nowak2008SheepBehaviour}
Nowak, R., Porter, R., Blache, D. \& Dwyer, C. (2008).
\newblock {\em Behaviour and the Welfare of the Sheep} (pp.\ 81--134).
\newblock Dordrecht: Springer Netherlands.

\bibitem[\protect\citeauthoryear{Orfandis}{Orfandis}{1988}{}]{OptimumSignalProcessing:1988}
Orfandis, S.~J. (1988).
\newblock {\em Optimum Signal Processing: A n Introduction\/} (2 Ed.).
\newblock University of Michigan: Macmillan.

\bibitem[\protect\citeauthoryear{Papaspyros, Bonnet, Collignon \&
  Mondada}{Papaspyros et~al.}{2019}{}]{10.1371/journal.pone.0220559}
Papaspyros, V., Bonnet, F., Collignon, B. \& Mondada, F. (2019).
\newblock Bidirectional interactions facilitate the integration of a robot into
  a shoal of zebrafish danio rerio.
\newblock {\em PLOS ONE}, {\em 14}, 1--25.

\bibitem[\protect\citeauthoryear{Park, Gong, Kang, Walton \& Kaminer}{Park
  et~al.}{2018}{}]{8444217}
Park, H., Gong, Q., Kang, W., Walton, C. \& Kaminer, I. (2018).
\newblock Observability analysis of an adversarial swarm’s cooperation
  strategy.
\newblock In {\em 14th International Conference on Control and Automation
  (ICCA)} (pp.\ 992--997). IEEE.

\bibitem[\protect\citeauthoryear{Pernek \& Ferscha}{Pernek \&
  Ferscha}{2017}{}]{Pernek2017}
Pernek, I. \& Ferscha, A. (2017).
\newblock A survey of context recognition in surgery.
\newblock {\em Med Biol Eng Comput}, {\em 55}, 1719--1734.

\bibitem[\protect\citeauthoryear{Pikovsky, Rosenblum \& Kurths}{Pikovsky
  et~al.}{2001}{}]{pikovsky_rosenblum_kurths_2001}
Pikovsky, A., Rosenblum, M. \& Kurths, J. (2001).
\newblock {\em Synchronization: A Universal Concept in Nonlinear Sciences}.
\newblock Cambridge Nonlinear Science Series. Cambridge University Press.

\bibitem[\protect\citeauthoryear{Porfiri}{Porfiri}{2018}{}]{Porfiri2018}
Porfiri, M. (2018).
\newblock Inferring causal relationships in zebrafish-robot interactions
  through transfer entropy: a small lure to catch a big fish.
\newblock {\em Animal Behavior and Cognition}, {\em 5}.

\bibitem[\protect\citeauthoryear{Priyadarshini, Sharma, Bhatt \&
  Al-Numay}{Priyadarshini et~al.}{2022}{}]{HAR2022:CPS}
Priyadarshini, I., Sharma, R., Bhatt, D. \& Al-Numay, M. (2022).
\newblock Human activity recognition in cyber-physical systems using optimized
  machine learning techniques.
\newblock {\em Cluster Computing}.

\bibitem[\protect\citeauthoryear{Puckett, Ni \& Ouellette}{Puckett
  et~al.}{2015}{}]{PhysRevLett.114.258103}
Puckett, J.~G., Ni, R. \& Ouellette, N.~T. (2015).
\newblock Time-frequency analysis reveals pairwise interactions in insect
  swarms.
\newblock {\em Phys. Rev. Lett.}, {\em 114}, 258103.

\bibitem[\protect\citeauthoryear{Qasem, Cardew, Wilson, Griffiths, Halsey,
  Shepard, Gleiss \& Wilson}{Qasem et~al.}{2012}{}]{Qasem2012:ODBAeqn}
Qasem, L., Cardew, A., Wilson, A., Griffiths, I., Halsey, L.~G., Shepard, E.
  L.~C., Gleiss, A.~C. \& Wilson, R. (2012).
\newblock Tri-axial dynamic acceleration as a proxy for animal energy
  expenditure; should we be summing values or calculating the vector?
\newblock {\em PLoS ONE}, {\em 7}.

\bibitem[\protect\citeauthoryear{Reséndiz-Benhumea, Froese, Ramos-Fernández
  \& Smith-Aguilar}{Reséndiz-Benhumea et~al.}{2019}{}]{10.1162/isal_a_00229}
Reséndiz-Benhumea, G.~M., Froese, T., Ramos-Fernández, G. \& Smith-Aguilar,
  S.~E. (2019).
\newblock {Applying Social Network Analysis to Agent-Based Models: A Case Study
  of Task Allocation in Swarm Robotics Inspired by Ant Foraging Behavior}.
\newblock In {\em ALIFE 2019: The 2019 Conference on Artificial Life} (pp.\
  616--623). ISAL.

\bibitem[\protect\citeauthoryear{Reynolds}{Reynolds}{1987}{}]{Reynolds1987Boids}
Reynolds, C. (1987).
\newblock Flocks, herds and schools: A distributed behavioral model.
\newblock In {\em Proceedings of the 14th annual conference on computer
  graphics and interactive techniques}, Volume 21(4) of {\em Siggraph '87}
  (pp.\ 25--34). ACM.

\bibitem[\protect\citeauthoryear{Rezaei, Munoz, Jalili \& Khayyam}{Rezaei
  et~al.}{2022}{}]{Rezaei2022:VitalNode}
Rezaei, A.~A., Munoz, J., Jalili, M. \& Khayyam, H. (2022).
\newblock Vital node identification in complex networks using a machine
  learning-based approach.
\newblock {\em arXiv}.

\bibitem[\protect\citeauthoryear{Schaerf, Herbert-Read \& Ward}{Schaerf
  et~al.}{2021}{}]{Schaerf:2021}
Schaerf, T.~M., Herbert-Read, J.~E. \& Ward, A. J.~W. (2021).
\newblock A statistical method for identifying different rules of interaction
  between individuals in moving animal groups.
\newblock {\em J. R. Soc. Interface}, {\em 18}.

\bibitem[\protect\citeauthoryear{Schreiber}{Schreiber}{2000}{}]{PhysRevLett.85.461}
Schreiber, T. (2000).
\newblock Measuring information transfer.
\newblock {\em Phys. Rev. Lett.}, {\em 85}, 461--464.

\bibitem[\protect\citeauthoryear{Shang \& Bouffanais}{Shang \&
  Bouffanais}{2014}{}]{Shang2014:SwarmTopo}
Shang, Y. \& Bouffanais, R. (2014).
\newblock Influence of the number of topologically interacting neighbors on
  swarm dynamics.
\newblock {\em Scientific Reports}, {\em 4}.

\bibitem[\protect\citeauthoryear{Sheikholeslami}{Sheikholeslami}{2019}{}]{Sheikholeslami1349978}
Sheikholeslami, S. (2019).
\newblock Ablation programming for machine learning.
\newblock Master's thesis, KTH, School of Electrical Engineering and Computer
  Science (EECS).

\bibitem[\protect\citeauthoryear{Sipahi \& Morfini}{Sipahi \&
  Morfini}{2020}{}]{sipahi:2020}
Sipahi, R. \& Morfini, M. (2020).
\newblock Improving on transfer-entropy network reconstruction using
  time-delays: approach and validation.
\newblock {\em Chaos}, {\em 30 (023125)}.

\bibitem[\protect\citeauthoryear{Spinello~C \& M}{Spinello~C \&
  M}{2019}{}]{spinello:2019}
Spinello~C, Yang~Y, M.~S. \& M, P. (2019).
\newblock Zebrafish adjust their behavior in response to an interactive robotic
  predator.
\newblock {\em Front. Robot. AI}, {\em 6:38}.

\bibitem[\protect\citeauthoryear{Stoica \& Randolph}{Stoica \&
  Randolph}{2005}{}]{SpectralAnalysis:2005}
Stoica, P. \& Randolph, M. (2005).
\newblock {\em Spectral Analysis of Signals}.
\newblock Upper Saddle River, NJ: Prentice Hall.

\bibitem[\protect\citeauthoryear{Str\"{o}mbom, Mann, Wilson, Hailes, Morton,
  Sumpter \& King}{Str\"{o}mbom et~al.}{2014}{}]{Strombom:2014}
Str\"{o}mbom, D., Mann, R.~P., Wilson, A.~M., Hailes, S., Morton, A.~J.,
  Sumpter, D. J.~T. \& King, A.~J. (2014).
\newblock Solving the shepherding problem: heuristics for herding autonomous,
  interacting agents.
\newblock {\em Journal of the Royal Society Interface}, {\em 11(100)},
  20140719.

\bibitem[\protect\citeauthoryear{Surasinghe \& Bollt}{Surasinghe \&
  Bollt}{2020}{}]{e22040396}
Surasinghe, S. \& Bollt, E.~M. (2020).
\newblock On geometry of information flow for causal inference.
\newblock {\em Entropy}, {\em 22(4)}.

\bibitem[\protect\citeauthoryear{Szwaykowska, Romero \& Schwartz}{Szwaykowska
  et~al.}{2015}{}]{7063970}
Szwaykowska, K., Romero, L. M.-y.-T. \& Schwartz, I.~B. (2015).
\newblock Collective motions of heterogeneous swarms.
\newblock {\em IEEE Transactions on Automation Science and Engineering}, {\em
  12(3)}, 810--818.

\bibitem[\protect\citeauthoryear{Traboulsi \& Barbeau}{Traboulsi \&
  Barbeau}{2019}{}]{9070871}
Traboulsi, A. \& Barbeau, M. (2019).
\newblock Recognition of drone formation intentions using supervised machine
  learning.
\newblock In {\em International Conference on Computational Science and
  Computational Intelligence (CSCI)} (pp.\ 408--411). IEEE.

\bibitem[\protect\citeauthoryear{Valentini, Mizumoto, Pratt, Pavlic \&
  Walker}{Valentini et~al.}{2019}{b}]{Valentini:2019a}
Valentini, G., Mizumoto, N., Pratt, S., Pavlic, T. \& Walker, S. (2019b).
\newblock Ants acknowledge information to control its rate of transfer.
\newblock {\em bioRxiv}.

\bibitem[\protect\citeauthoryear{Wang, Miller, Lizier, Prokopenko \&
  Rossi}{Wang et~al.}{2011}{}]{Wang2011MeasuringIS}
Wang, X.~R., Miller, J.~M., Lizier, J.~T., Prokopenko, M. \& Rossi, L.~F.
  (2011).
\newblock Measuring information storage and transfer in swarms.
\newblock In {\em Eleventh European Conference on the Synthesis and Simulation
  of Living Systems} (pp.\ 838--845). Paris, France: ECAL.

\bibitem[\protect\citeauthoryear{Wang, Miller, Lizier, Prokopenko \&
  Rossi}{Wang et~al.}{2012}{}]{10.1371/journal.pone.0040084}
Wang, X.~R., Miller, J.~M., Lizier, J.~T., Prokopenko, M. \& Rossi, L.~F.
  (2012).
\newblock Quantifying and tracing information cascades in swarms.
\newblock {\em Plos One}, {\em 7(7)}, 1--7.

\bibitem[\protect\citeauthoryear{Williams}{Williams}{2007}{}]{Williams2007:working}
Williams, T. (2007).
\newblock {\em Working sheep dogs: a practical guide to breeding, training and
  handling}.
\newblock Collingwood, Victoria: Landlinks Press.

\bibitem[\protect\citeauthoryear{Wu, Su, Tang \& Tianfield}{Wu
  et~al.}{2011}{}]{Wu2011:Emergence}
Wu, Y., Su, J., Tang, H. \& Tianfield, H. (2011).
\newblock Analysis of the emergence in swarm model based on largest lyapunov
  exponent.
\newblock {\em Mathematical Problems in Engineering}, {\em 2011}.

\bibitem[\protect\citeauthoryear{Yaxley, McIntyre, Park \& Healey}{Yaxley
  et~al.}{2021}{a}]{Yaxley2021:SS}
Yaxley, K., McIntyre, N., Park, J. \& Healey, J. (2021a).
\newblock Sky shepherds: a tale of a uav and sheep.
\newblock In H.~Abbass \& R.~Hunjet (Eds.), {\em Shepherding UxVs for
  Human-Swarm Teaming}  chapter~9,  (pp.\ 189--206). Springer, Cham.

\bibitem[\protect\citeauthoryear{Yaxley, Joiner \& Abbass}{Yaxley
  et~al.}{2021}{b}]{Yaxley2020SkyShepherding}
Yaxley, K.~J., Joiner, K.~F. \& Abbass, H. (2021b).
\newblock Drone approach parameters leading to lower stress sheep flocking and
  movement: sky shepherding.
\newblock {\em Scientific reports}, {\em 11(1)}, 1--9.

\end{thebibliography}

\clearpage
\onecolumn

\begin{biog}

   \begin{table*}[ht]
        \centering
        \def\arraystretch{2.5}
        \resizebox{\textwidth}{!}{%
        \begin{tabular}{L{7em} L{35em}}
             {{\includegraphics[width=25mm,height=32mm,clip,keepaspectratio]{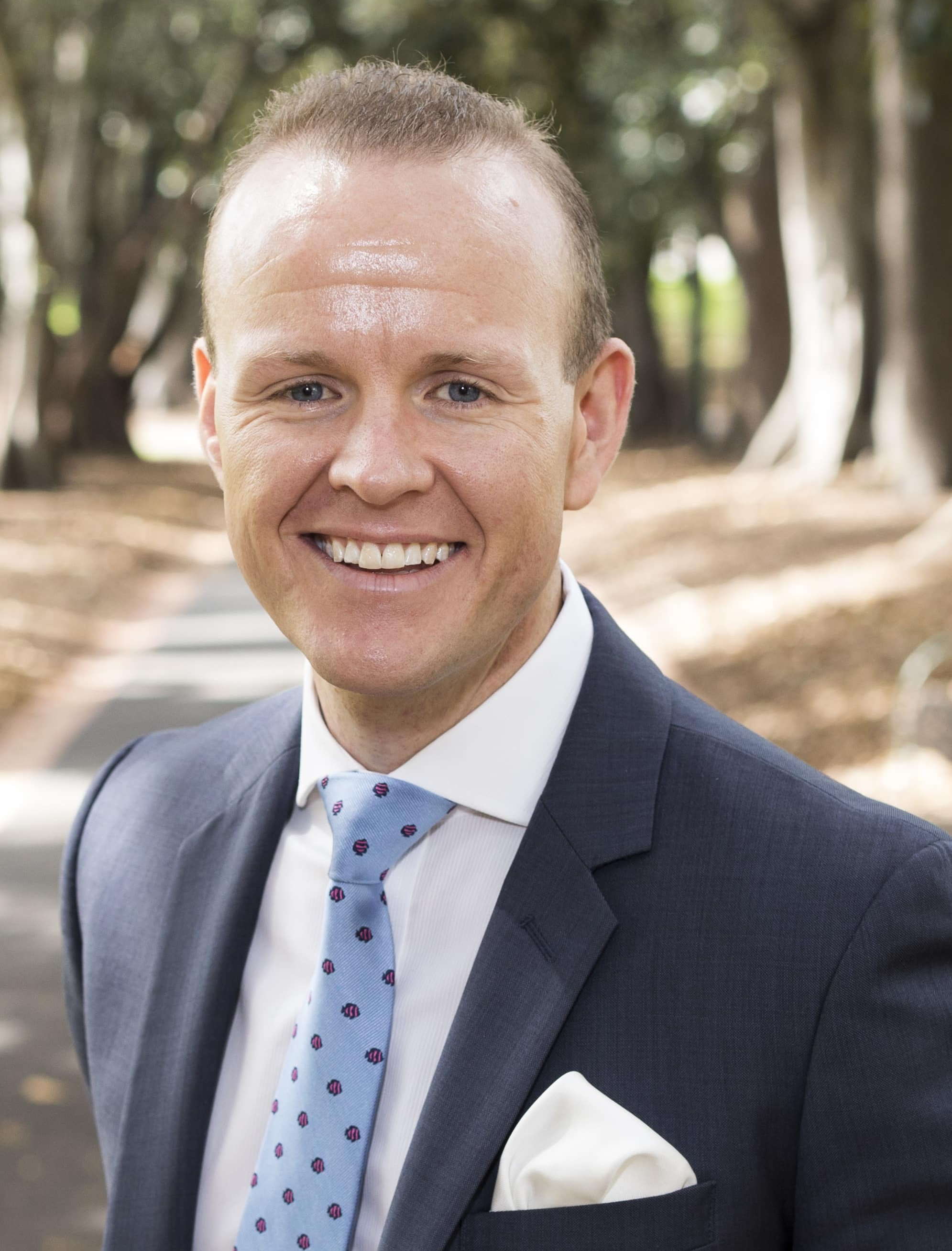}}} & {\small\textbf{Adam J. Hepworth} is a PhD candidate in Computer Science at the University of New South Wales (UNSW) Canberra, where he is the 2022 Chief of Army Scholar. He received an M.S. in Operations Research from the Naval Postgraduate School (2017), an M.Log\&SupChMngmt from the University of South Australia (2015), and a B.Sc. in Mathematics from the University of New South Wales, Canberra (2009). His current research contributes to swarm shepherding for human-swarm teaming, activity recognition and behaviour prediction, and the design of artificial intelligence systems.} \\
             
             {{\includegraphics[width=25mm,height=32mm,clip,keepaspectratio]{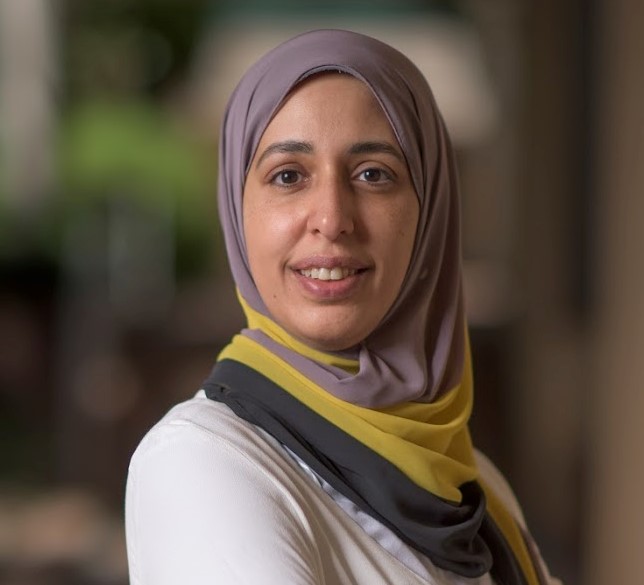}}} & {\small\textbf{Aya Hussein} is a research associate at the School of Engineering and Information Technology, University of New South Wales, Canberra. Her research focuses on combining human and machine intelligence by studying interaction schemes that allow humans to teach machines and to effectively team up with them in the field. She received her PhD in Computer Science from UNSW-Canberra in 2020 with a focus on human-swarm teaming. She got her Bachelor's and Master's degrees in Computer Engineering from Cairo University in 2011 and 2015.} \\
             
             {{\includegraphics[width=25mm,height=32mm,clip,keepaspectratio]{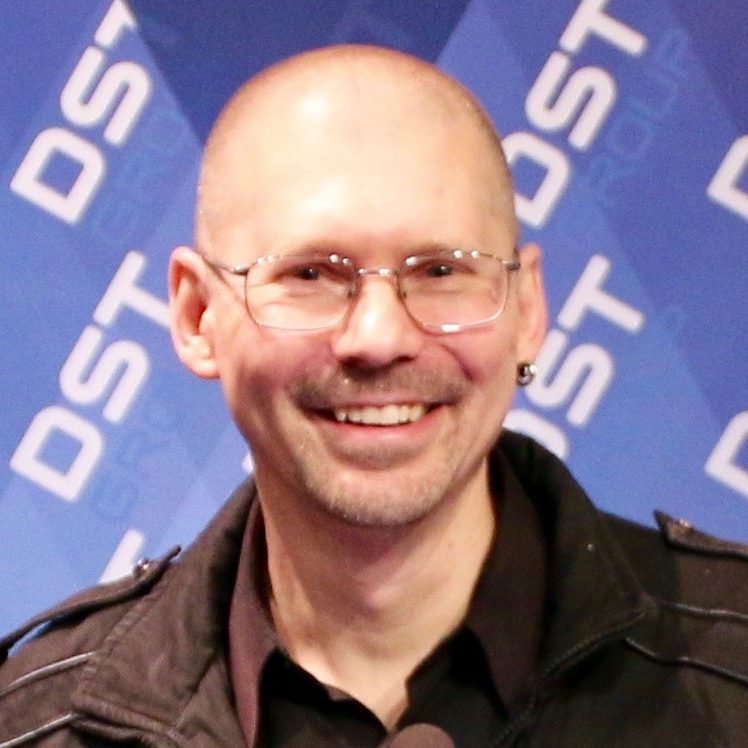}}} & {\small\textbf{Darryn J. Reid} has been with DSTO since 1995, and has worked in distributed systems, distributed databases, machine learning and artificial intelligence, interoperability, formal reasoning and logic, modelling of C2, simulation, optimisation and optimal control, electronic warfare, intelligence analysis tools, missile targeting and control, command support systems, operations research, parallel and distributed computation, hardware design, mathematical control theory, mathematical complexity, advanced web-based technologies, languages, model theory and computation, stochastic modelling, formal ontology, object-oriented and functional programming, crowd modelling and military theory. He holds the degrees of Bachelor of Science in Mathematics and Computer Science, Bachelor of Science with First Class Honours in Mathematics and Computer Science, and Doctor of Philosophy in Theoretical Computer Science from the University of Queensland. He has strong research interests in pure and applied mathematics, theoretical and applied computer science, philosophy, military theory and economics. In other words, he knows just enough to realise how ignorant he is.} \\
             
             {{\includegraphics[width=25mm,height=32mm,clip,keepaspectratio]{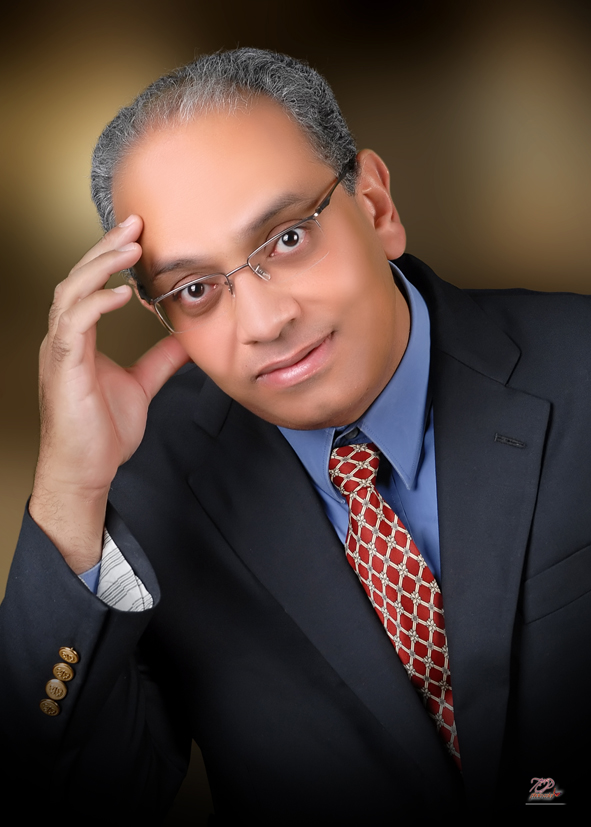}}} & {\small\textbf{Hussein A. Abbass} is a full professor with the School of Engineering and Information Technology, University of New South Wales, Canberra (UNSW Canberra) Campus. He is a Fellow of IEEE, a Fellow of the Australian Computer Society, a Fellow of the UK Operational Research Society, a Fellow of the Australian Institute of Managers and Leaders, and a Graduate Member of the Australian Institute of Company Directors (GAICD). He was the National President (2016-2019) for the Australian Society for Operations Research (ASOR), the Vice-President for Technical Activities (2016-2019) for the IEEE Computational Intelligence Society, and an ExCom and AdCom member (2016-2019) of the IEEE Computational Intelligence Society. Prof. Abbass is a Distinguished Lecturer for the IEEE Computational Intelligence Society and the Founding Editor-in-Chief of the IEEE Transactions on Artificial Intelligence. He is an Associate Editor of several IEEE journals including IEEE Transactions on Cybernetics and IEEE Transactions on Evolutionary Computation, and a Senior Associate Editor for ACM Computing Surveys. After 10 years in industry and academia, he joined UNSW Canberra in 2000 and has been a full professor since 2007. His current research focuses on AI-enabled swarm systems, shepherding-based swarm guidance, human-AI teaming, trusted human-swarm teaming, and machine education.} \\
        \end{tabular}
        }
    \end{table*}
\end{biog}

\clearpage
    \begin{sm}
        \section{Swarm Literature Summary Methods, Techniques and Measures}
        
            \bottomcaption{Summary of methods, techniques, and measures identified in this study, organised by primary academic field (column~3). The three primary swarm lenses introduced in the literature review summarise the use of each MTM for swarm analyses (columns 4-6).}
            \label{table:AnalysisLiterature}
            \begin{xtabular*}{\textwidth}{L{6em} C{10em} L{7em} C{7em} C{7em} C{8em}}
                \toprule
                 \textbf{Method, Technique or Measure (MTM)} & \textbf{Source} & \textbf{Primary Field} &  \textbf{Leadership, Coordination and Influence} & \textbf{Swarm Dynamics and Emergent Behaviour} & \textbf{Agents and Individual Characterisation} \\ \midrule
                \rule{0pt}{4ex}Synchronicity                   & \rule{0pt}{4ex}Hepworth et al.,~\citeyear{Hepworth2020:Footprints}                                                                                                   & \rule{0pt}{4ex}Information Theory    &            & \rule{0pt}{4ex}\checkmark & \rule{0pt}{4ex}\checkmark \\
                \rule{0pt}{4ex}Conditional TE                  & \rule{0pt}{4ex}Bossomaier et al.,~\citeyear{Bossomaier2016}                                                                                                          & \rule{0pt}{4ex}Information Theory    &            & \rule{0pt}{4ex}\checkmark & \rule{0pt}{4ex}\checkmark \\
                \rule{0pt}{4ex}Local TE                        & \rule{0pt}{4ex}Crosato et al.,~\citeyear{crosatoSwarmIntell}, Bossomaier et al.,~\citeyear{Bossomaier2016}                                                           & \rule{0pt}{4ex}Information Theory    &            & \rule{0pt}{4ex}\checkmark & \rule{0pt}{4ex}\checkmark \\
                \rule{0pt}{4ex}Information Storage             & \rule{0pt}{4ex}Wang et al.,~\citeyear{10.1371/journal.pone.0040084}, Bossomaier et al.,~\citeyear{Bossomaier2016}, Wang et al.,~\citeyear{Wang2011MeasuringIS}       & \rule{0pt}{4ex}Information Theory    &            & \rule{0pt}{4ex}\checkmark & \rule{0pt}{4ex}\checkmark \\
                \rule{0pt}{4ex}Global TE                       & \rule{0pt}{4ex}Bossomaier et al.,~\citeyear{Bossomaier2016}                                                                                                          & \rule{0pt}{4ex}Information Theory    & \rule{0pt}{4ex}\checkmark & \rule{0pt}{4ex}\checkmark &            \\
                \rule{0pt}{4ex}Shannon Entropy                 & \rule{0pt}{4ex}Hamann et al.,~\citeyear{Hamann2012:Emergence}, Bossomaier et al.,~\citeyear{Bossomaier2016}                                                          & \rule{0pt}{4ex}Information Theory    &            & \rule{0pt}{4ex}\checkmark &            \\
                \rule{0pt}{4ex}Cross Entropy                   & \rule{0pt}{4ex}Bossomaier et al.,~\citeyear{Bossomaier2016}, Cofta et al.,~\citeyear{e22060597}                                                                      & \rule{0pt}{4ex}Information Theory    &            & \rule{0pt}{4ex}\checkmark &            \\
                \rule{0pt}{4ex}Causation Entropy               & \rule{0pt}{4ex}Pilkiewicz et al.,~\citeyear{pilkiewicz:2020}, Lord et al.,~\citeyear{7809221}                                                                        & \rule{0pt}{4ex}Information Theory    & \rule{0pt}{4ex}\checkmark &            &            \\
                \rule{0pt}{4ex}Time Delayed TE                 & \rule{0pt}{4ex}Sipahi \& Morfini~\citeyear{sipahi:2020}                                                                                                              & \rule{0pt}{4ex}Information Theory    & \rule{0pt}{4ex}\checkmark &            &            \\
                \rule{0pt}{4ex}Effort-to-Compress              & \rule{0pt}{4ex}Kathpalia~\citeyear{Kathpalia:2021}, Nagaraj et al.,~\citeyear{ETC:2013}                                                                              & \rule{0pt}{4ex}Information Theory    & \rule{0pt}{4ex}\checkmark &            &            \\
                \rule{0pt}{4ex}Information Flow                & \rule{0pt}{4ex}Bossomaier et al.,~\citeyear{Bossomaier2016}, Wang et al.,~\citeyear{10.1371/journal.pone.0040084}, Li et al.,~\citeyear{Li2004:CollabSwarm}          & \rule{0pt}{4ex}Information Theory    & \rule{0pt}{4ex}\checkmark &            &            \\
                \rule{0pt}{4ex}Polarisation                    & \rule{0pt}{4ex}Crosato et al.,~\citeyear{crosatoSwarmIntell}, Brown \& Goodrich~\citeyear{10.5555/2615731.2615798}                                                   & \rule{0pt}{4ex}Geometric \& Spatial  & \rule{0pt}{4ex}\checkmark &            & \rule{0pt}{4ex}\checkmark \\
                \rule{0pt}{4ex}Geotaxi                         & \rule{0pt}{4ex}Spinello C \& M~\citeyear{spinello:2019}                                                                                                              & \rule{0pt}{4ex}Geometric \& Spatial  & \rule{0pt}{4ex}\checkmark &            &            \\
                \rule{0pt}{4ex}Situation Awareness             & \rule{0pt}{4ex}Hepworth et al.,~\citeyear{Hepworth2020:Footprints}                                                                                                   & \rule{0pt}{4ex}Geometric \& Spatial  &            & \rule{0pt}{4ex}\checkmark & \rule{0pt}{4ex}\checkmark \\
                \rule{0pt}{4ex}Predation Risk                  & \rule{0pt}{4ex}Hepworth et al.,~\citeyear{Hepworth2020:Footprints}                                                                                                   & \rule{0pt}{4ex}Geometric \& Spatial  &            & \rule{0pt}{4ex}\checkmark & \rule{0pt}{4ex}\checkmark \\
                \rule{0pt}{4ex}Spatial Distance                & \rule{0pt}{4ex}Str\" {o}mbom et al.,~\citeyear{Strombom:2014}, Valentini et al.,~\citeyear{Valentini:2019a}                                                          & \rule{0pt}{4ex}Geometric \& Spatial  &            &            & \rule{0pt}{4ex}\checkmark \\
                \rule{0pt}{4ex}Escape Trajectory               & \rule{0pt}{4ex}Chakraborty et al.,~\citeyear{Chakraborty2020}                                                                                                        & \rule{0pt}{4ex}Geometric \& Spatial  &            &            & \rule{0pt}{4ex}\checkmark \\
                \rule{0pt}{4ex}Speed                           & \rule{0pt}{4ex}Schaerf et al.,~\citeyear{Schaerf:2021}, Abbass and Hunjet~\citeyear{Abbass2020:UxV}, Traboulsi \& Barbeau~\citeyear{9070871}                         & \rule{0pt}{4ex}Geometric \& Spatial  &            & \rule{0pt}{4ex}\checkmark & \rule{0pt}{4ex}\checkmark \\
                \rule{0pt}{4ex}Heading                         & \rule{0pt}{4ex}Schaerf et al.,~\citeyear{Schaerf:2021}, Abbass and Hunjet~\citeyear{Abbass2020:UxV}                                                                  & \rule{0pt}{4ex}Geometric \& Spatial  &            &            & \rule{0pt}{4ex}\checkmark \\
                \rule{0pt}{4ex}Acceleration                    & \rule{0pt}{4ex}Schaerf et al.,~\citeyear{Schaerf:2021}, Abbass and Hunjet~\citeyear{Abbass2020:UxV}                                                                  & \rule{0pt}{4ex}Geometric \& Spatial  &            &            & \rule{0pt}{4ex}\checkmark \\
                \rule{0pt}{4ex}Angular Velocity                & \rule{0pt}{4ex}Hepworth~\citeyear{Hepworth2021:ARS}, Brown \& Goodrich~\citeyear{10.5555/2615731.2615798}                                                            & \rule{0pt}{4ex}Geometric \& Spatial  &            &            & \rule{0pt}{4ex}\checkmark \\
                \rule{0pt}{4ex}Dynamic Body Acceleration       & \rule{0pt}{4ex}Gleiss et al.,~\citeyear{ODBA:2011}, Martín López et al.,~\citeyear{10.1111/2041-210X.13751}                                                          & \rule{0pt}{4ex}Geometric \& Spatial  &            &            & \rule{0pt}{4ex}\checkmark \\
                \rule{0pt}{4ex}Topological Analysis            & \rule{0pt}{4ex}Papaspyros et al.,\citeyear{10.1371/journal.pone.0220559}                                                                                             & \rule{0pt}{4ex}Geometric \& Spatial  & \rule{0pt}{4ex}\checkmark &            &            \\
                \rule{0pt}{4ex}Geometric Information Flow      & \rule{0pt}{4ex}Surasinghe \& Bollt~\citeyear{e22040396}                                                                                                              & \rule{0pt}{4ex}Geometric \& Spatial  & \rule{0pt}{4ex}\checkmark &            &            \\
                \rule{0pt}{4ex}Spatial Alignment               & \rule{0pt}{4ex}Reynolds~\citeyear{Reynolds1987Boids}                                                                                                                 & \rule{0pt}{4ex}Geometric \& Spatial  &            & \rule{0pt}{4ex}\checkmark &            \\
                \rule{0pt}{4ex}Spatial Cohesion                & \rule{0pt}{4ex}Reynolds~\citeyear{Reynolds1987Boids}                                                                                                                 & \rule{0pt}{4ex}Geometric \& Spatial  &            & \rule{0pt}{4ex}\checkmark &            \\
                \rule{0pt}{4ex}Spatial Separation              & \rule{0pt}{4ex}Reynolds~\citeyear{Reynolds1987Boids}                                                                                                                 & \rule{0pt}{4ex}Geometric \& Spatial  &            & \rule{0pt}{4ex}\checkmark &            \\
                \rule{0pt}{4ex}Centre of Mass                  & \rule{0pt}{4ex}Str\" {o}mbom et al.,~\citeyear{Strombom:2014}, Valentini et al.,~\citeyear{Valentini:2019a}                                                          & \rule{0pt}{4ex}Geometric \& Spatial  &            & \rule{0pt}{4ex}\checkmark &            \\
                \rule{0pt}{4ex}Dynamic Time Warping            & \rule{0pt}{4ex}Amornbunchornvej~\citeyear{mFLICA:2021}                                                                                                               & \rule{0pt}{4ex}Time Series Analysis  & \rule{0pt}{4ex}\checkmark & \rule{0pt}{4ex}\checkmark &            \\
                \rule{0pt}{4ex}Granger Causality               & \rule{0pt}{4ex}Lord et al.,~\citeyear{7809221}                                                                                                                       & \rule{0pt}{4ex}Time Series Analysis  & \rule{0pt}{4ex}\checkmark & \rule{0pt}{4ex}\checkmark &            \\
                \rule{0pt}{4ex}Lyapunov Exponent               & \rule{0pt}{4ex}Wu et al.,~\citeyear{Wu2011:Emergence}, Baldi \& Frasca~\citeyear{BALDI2019935}                                                                       & \rule{0pt}{4ex}Time Series Analysis  &            & \rule{0pt}{4ex}\checkmark &            \\
                \rule{0pt}{4ex}Frequency Analysis              & \rule{0pt}{4ex}Puckett et al.,~\citeyear{PhysRevLett.114.258103}                                                                                                     & \rule{0pt}{4ex}Time Series Analysis  &            & \rule{0pt}{4ex}\checkmark &            \\
                \rule{0pt}{4ex}Spectral Analysis               & \rule{0pt}{4ex}Andrade et al.,\citeyear{1698861}                                                                                                                     & \rule{0pt}{4ex}Time Series Analysis  &            & \rule{0pt}{4ex}\checkmark &            \\
                \rule{0pt}{4ex}Correlation Function            & \rule{0pt}{4ex}Mateo et al.,~\citeyear{Mateo:2017}                                                                                                                   & \rule{0pt}{4ex}Time Series Analysis  & \rule{0pt}{4ex}\checkmark & \rule{0pt}{4ex}\checkmark &            \\
                \rule{0pt}{4ex}Thermo \& Fluid Dynamics        & \rule{0pt}{4ex}Haeri et al.,\citeyear{10.1115/1.4046580}, Jankovic~\citeyear{Jankovic2018:CFD}, Mavridis et al.,~\citeyear{SwarmInteractions:2021}                   & \rule{0pt}{4ex}Physics               &            & \rule{0pt}{4ex}\checkmark &            \\
                \rule{0pt}{4ex}Fluctuation Theorem             & \rule{0pt}{4ex}Hamann et al.,~\citeyear{Hamann2012:Emergence}                                                                                                        & \rule{0pt}{4ex}Physics               &            & \rule{0pt}{4ex}\checkmark &            \\
                \rule{0pt}{4ex}Density and Pressure            & \rule{0pt}{4ex}Andrade et al.,\citeyear{1698861}                                                                                                                     & \rule{0pt}{4ex}Physics               &            & \rule{0pt}{4ex}\checkmark &            \\
                \rule{0pt}{4ex}Social Network Analysis         & \rule{0pt}{4ex}Res\'{e}ndiz-Benhumea et al.,~\citeyear{10.1162/isal_a_00229}                                                                                         & \rule{0pt}{4ex}Graph Theory          & \rule{0pt}{4ex}\checkmark &            &            \\
                \rule{0pt}{4ex}Nodal Analysis                  & \rule{0pt}{4ex}Mocanu et al.,\citeyear{6973878}, Shang \& Bouffanais~\citeyear{Shang2014:SwarmTopo}                                                                  & \rule{0pt}{4ex}Graph Theory          & \rule{0pt}{4ex}\checkmark &            &            \\
                \bottomrule
            \end{xtabular*}
            
        \newpage
        \section{Information Marker Equations \& Experimental Design Marker Set}
    
        We use the term \textit{segment} throughout this appendix to reference the period of observation for a marker. A segment period consists of $k$ observations over the window $\Delta_k = t_1\rightarrow t_k$, with each marker output summarised over $\Delta_k$. The sequence of segments $\Delta_{k_1}\prec \Delta_{k_2}~\prec~\dots~\prec~\Delta_{ k_K}$ contain a consistent overlap of $\alpha$, given as $\Delta_k^\alpha$. The following common notations, vide Abbass and Hunjet~\citeyear{Abbass2020:UxV}, are throughout
        \begin{itemize}
            \item $\pi_i$, swarm agent $i$.
            \item $\beta$, swarm control agent.
            \item $S^t$, speed at time $t$.
            \item $P^t$, position of an agent at time $t$.
            \item $P^T$, the position of an agent at an end time $T$.
        \end{itemize}
    
        Table~\ref{table:MarkerLiteratures} contains the set of markers ($\mathcal{M}$) used throughout this study that are identified numerically as M$i$. There are three primitive information elements in our work being speed (M1, M3\textemdash M4), heading (M5\textemdash M6) and distance (M2, M17\textemdash M20), which are calculated by standard methods. In addition, there are derivations of these calculations for the segment of observation, including mean and variance. These primitive markers are used both in the discrimination of agent and swarm types and inputs to higher-order markers.
    
        \subsection{Situation Awareness (M7\textemdash M8)}
    
        Our first higher-order marker is \textit{Situation Awareness} (SA), calculated as defined in Hepworth et al.,~\citeyear{Hepworth2020:Footprints}. There are two variations for SA: the mean and variance of each agent $\in\Delta_k$. The SA is formulated to capture the perspective of each swarm agent relative to the swarm control agent. SA is maximised \enquote{when there exists an unobstructed line-of-sight}~(pg.4) between a swarm agent and the swarm control agent and is \enquote{minimum at the furthest point of the convex hull from}~(pg.4) from the swarm control agent with the maximum number of line-of-sight obstructions. The SA is given as
        \begin{equation}\label{eq:SA}
            SA^t_{\pi_i} = \dfrac{1}{\dfrac{d_{\pi_{i}\rightarrow\beta}^2}{d_{\pi_{i}\rightarrow\Gamma_{\Pi}} * \enskip d_{\Pi\rightarrow\beta}} * \Theta+1},
        \end{equation}
        which is calculated with a measure of distance from a swarm agent to the swarm control agent, the number of swarm control agents impeding the line-of-sight, and distance to the swarm Global Centre of Mass ($\Gamma_\Pi$). The number of line-of-sight impediments is given as $\Theta$, distance between agents denoted by $d$ for $\pi_i \rightarrow \beta \ \forall \ \pi_i \in \Pi$.
    
        \subsection{Predation Risk (M9\textemdash M10)}
    
        The next higher-order marker is Predation Risk (PR), calculated as defined in Hepworth et al.,~\citeyear{Hepworth2020:Footprints}. There are two PR variations, the mean and variance of each agent $\in\Delta_k$. The PR is formulated to capture the perspective of each swarm control agent, relative to the \enquote{likelihood of an agent encountering a predator and the potential to safety, should this predator (perceived or real) attack the same agent}~(pg.4). The PR may be characterised as capturing the centre-seeking behaviour of a swarm agent, given as
        \begin{equation}
            PR^t_{\pi_i} = \dfrac{1}{O_b}*\dfrac{N}{\Omega_{\pi\pi}+1},
        \end{equation}
        where the number of bins ($B$) is determined by the ceiling-integer square root of the number of swarm agents ($B=\lceil \sqrt{N} \rceil$), $O_b$ is the bin-order, and $N$ is the cardinality of $\Pi$. Each $O_b$ is uniformly distributed between the agent closest to the swarm control agent to the furthest agent, with $O_1$ representing the closest, $O_B$ representing the furthest, and the highest PR observed in $O_1$.
    
        \subsection{Dynamic Body Acceleration (M11\textemdash M13)}
    
        Our higher-order marker Dynamic Body Acceleration (DBA) is calculated as defined in~\shortcite{ODBA:2011,10.1111/2041-210X.13751,Miwa2015:ODBA,Qasem2012:ODBAeqn}, with three variations being the cumulative DBA (Overall DBA, ODBA), mean and variance of each agent. The DBA is the tri-axial acceleration ($a$) of the agent, given as
        \begin{equation}
            DBA_{\pi_i}^t = \sqrt{a_x^2 + a_y^2 + a_z^2},
        \end{equation}
        where $a_x,a_y,a_z$ are the vector-component accelerations of an agent. For our purposes in simulation, we reduce this to two dimensions, omitting the vertical dimension ($z$). ODBA is the cumulative $DBA$, given as
        \begin{equation}
            ODBA_{\pi_i}^k = \sum_{t=1}^{\vert k\vert}{\vert a_x^t \vert + \vert a_y^t \vert + \vert a_z^t \vert}.
        \end{equation}
        ODBA is the \enquote{sum of the DA magnitude over a reference interval used as an activity and energy proxy}~(pg.10)~\shortcite{10.1111/2041-210X.13751}.
    
        \subsection{Rate of Change (M14)}
    
        The higher-order marker rate of change is calculated as defined in~\shortcite{Hepworth2021:ARS}, with one marker for the angular rate of change being the velocity. This marker \enquote{calculates the rate of change for the direction of the angle of \dots motion}~(pg.157). An underlying assumption of this calculation is that there exists a smoothness throughout the change, given as
        \begin{equation}
            \delta = \dfrac{\text{atan2}\left( \vert\vert P^{t+1} - P^{t} \vert\vert \right) - \text{atan2}\left( \vert\vert P^{t} - P^{t-1} \vert\vert \right)}{\Delta{t}},
        \end{equation}
        where the present coordinate of an agent is denoted by $P^{t}$, the previous position of an agent is denoted by $P^{t-1}$ and the future position of an agent is denoted by $P^{t+1}$.
    
        \subsection{Cross Correlation (M15\textemdash M16)}
    
        The use of Cross-Correlation is as described in~\shortcite{Mateo:2017} and defined in~\shortcite{SpectralAnalysis:2005,OptimumSignalProcessing:1988}. There are two cross-correlation markers, the mean and variance for a given period of observation, $k$. The cross-correlation computes the \enquote{similarity between a vector $x$ and shifted (lagged) copies of a vector $y$ as a function of the lag}~\shortcite{Matlab:xcorr} and is computed via standard means.
    
        \subsection{Transfer Entropy (M23, M27, M29\textemdash M32)}
    
        Our higher-order marker for Transfer Entropy (TE) is calculated as defined in~\shortcite{crosatoSwarmIntell,Bossomaier2016}. One marker for TE is the Local TE (Net). The TE is a \enquote{a non-parametric approach that provides a measure of the asymmetric, directed transfer of information between two stochastic processes}~(pg.3)~\shortcite{Hepworth2020:Footprints}. Schreiber~\citeyear{PhysRevLett.85.461} first defined TE as
        \begin{equation}\label{eqn:TE}
            T_{J \rightarrow I}(k,l) = \sum_{i,j} p\left(i_{t+1}, i_t^{(k)}, j_t^{(l)}\right) \cdot log \left(\dfrac{p\left(i_{t+1}| i_t^{(k)}, j_t^{(l)}\right)}{p\left(i_{t+1}|i_t^{(k)}\right)}\right),
        \end{equation}
        \begin{equation}\label{eq:localTransferEntropy}
            te_{J\rightarrow I} = t(i,j,n+1,l) = \lim_{k \rightarrow \infty} \log \dfrac{p (x_{i, n+1} \vert x_{i,n}^{(k)}, x_{i-j,n}^{(l)})}{p(x_{i, n+1} \vert x_{i,n}^{(k)})}.
        \end{equation}
        In our system we assume as first-order Markov process, with the embedding dimension ($k$), embedding delay ($\tau$) and lag ($l$) equal, such that $k=\tau=l=1$ ~\shortcite{Lizier:2008,Bossomaier2016}. With this assumption, Equation~\ref{eq:localTransferEntropy} is implemented as
        \begin{equation}\label{eq:lteSimple}
            te_{J\rightarrow I} = t(i,j,n) = \log \dfrac{p (x_{i, n} \vert x_{i,n-1}, x_{i-j,n-1})}{p(x_{i, n} \vert x_{i,n-1})},
        \end{equation}
        where the NetTE as defined in Porfiri~\citeyear{Porfiri2018} is given as
        \begin{equation}\label{eqn:NetTE}
            T^{\text{net}}_{J \rightarrow I} = \text{NetTE}_{J \rightarrow I} = \text{te}_{J \rightarrow I} - \text{te}_{I \rightarrow J}.
        \end{equation}
        The net is widely used to infer the dynamics of swarms~\shortcite{trendafilov:2020,Porfiri2018,PhysRevE.93.042411}, where non-zero NetTE provides insight as to the asymmetry of interactions between two agents. The net value may be interpreted for three cases.
        \begin{itemize}
            \item $T^{\text{net}}_{J \rightarrow I} > 0$ infers that $J$ is informative, or influences $I$.
            \item $T^{\text{net}}_{J \rightarrow I} < 0$ infers that $J$ is misinformative, or influence by $I$.
            \item $T^{\text{net}}_{J \rightarrow I} = 0$ infers that there is no detected coupling between $J$ and $I$, or that the interaction between these agents is symmetric.
        \end{itemize}
        The Total TE (TotTE) is a measure of the \enquote{magnitude of total influence for a pairwise interaction}~(pg.3)~\shortcite{Hepworth2020:Footprints}, differing from the NetTE as it does not consider the directionality of interaction between agents. We define TotTE here as
        \begin{equation}\label{eqn:TotTE}
            T^{\text{tot}}_{J \rightarrow I} = \text{TotTE}_{J \rightarrow I} = \text{te}_{J \rightarrow I} + \text{te}_{I \rightarrow J},
        \end{equation}
        intuitively capturing the intensity of interactions between agents.
    
        \subsection{Synchronicity (M21\textemdash M22)}
    
        Our equation for Synchronicity is defined in~\shortcite{Hepworth2020:Footprints}. There are two markers for synchronicity, the mean and variance. Synchronicity is defined as the \enquote{alignment in time and space of action resulting from a significant influence}~(pg.3)~\shortcite{Hepworth2020:Footprints}, based on the work of Pikovsky et al.,~\citeyear{pikovsky_rosenblum_kurths_2001}. The Synchronicity is calculated based on TE, combining the NetTE and TotTE defined in Equations~\ref{eqn:NetTE} and Equation~\ref{eqn:TotTE}, given as
        \begin{equation}\label{eq:synchronicity}
            S_{J\rightarrow I}^t = \text{sgn} \left( T^{\text{net}}_{J \rightarrow I} \right) * \vert T^{\text{tot}}_{J \rightarrow I} \vert,
        \end{equation}
        where the direction NetTE$_{J\rightarrow I}$ agent interaction is returned by $sign(.)$. As with TE, we highlight three cases of interaction between a source agent $J$ and target agent $I$
        \begin{itemize}
            \item $S_{J\rightarrow I} > 0$. $J$ is informative, or influences, $I$.
            \item $S_{J\rightarrow I} < 0$. $J$ is misinformative or is influenced by $I$.
            \item $S_{J\rightarrow I} = 0$. $J$ does not inform, or does not influence, $I$.
        \end{itemize}
        Where $S_{J\rightarrow I} >> 0$ or $S_{J\rightarrow I} << 0$, Hepworth et al.,~\citeyear{Hepworth2020:Footprints} suggest that this represents the \enquote{intensity of the relationship between $J$ and $I$}~(pg.4).
    \end{sm}
\end{document}